\documentclass{article} % For LaTeX2e
\usepackage{iclr2026_conference,times}
\pdfoutput=1 % 
\usepackage{amsmath,amsfonts,bm}

\def\eqref#1{equation~\ref{#1}}
\def\1{\bm{1}}

\DeclareMathAlphabet{\mathsfit}{\encodingdefault}{\sfdefault}{m}{sl}
\SetMathAlphabet{\mathsfit}{bold}{\encodingdefault}{\sfdefault}{bx}{n}

\usepackage{hyperref}
\usepackage{url}

\usepackage{times}  
\usepackage{helvet}  
\usepackage{courier}  
\usepackage{graphicx} 
\usepackage{natbib}   
\usepackage{caption}
\usepackage{listings}
\usepackage{tcolorbox}
\usepackage{algorithm}
\usepackage{algorithmic}
\usepackage{amsfonts}
\usepackage{xcolor}
\usepackage{graphicx}
\usepackage{subcaption}
\usepackage{threeparttable}
\usepackage{booktabs}    % For \toprule, \midrule, \bottomrule, \cmidrule
\usepackage{multirow}    % For \multirow
\usepackage{amsmath}     % For \text
\usepackage{xspace}  
\usepackage{array}
\newcolumntype{C}[1]{>{\centering\arraybackslash}m{#1}}
\usepackage{adjustbox}
\usepackage{colortbl}
\definecolor{sem}{HTML}{77BEF0}
\definecolor{myblue}{RGB}{6,119,215}
\usepackage{subcaption}
\usepackage{placeins}
\title{\raggedright {Follow-Your-Preference: Towards Preference-Aligned Image Inpainting}}

\author{Yutao Shen$^{1*} $  Junkun Yuan$^{2*\dag}$ Toru Aonishi$^{1}$ Hideki Nakayama$^{1}$ Yue Ma$^{3\dag}$ \\
$^{1} $The University of Tokyo $^{2} $Zhejiang University $^{3} $Tsinghua University \\
}

\iclrfinalcopy % Uncomment for camera-ready version, but NOT for submission.
\begin{document}

\maketitle
\let\thefootnote\relax\footnotetext{$*$ Equal contribution.}
\let\thefootnote\relax\footnotetext{$\dag$ Corresponding author.}
\begin{figure*}[h]
\centering
\includegraphics[width=\linewidth, trim=0mm 0mm 0mm 0mm, clip]{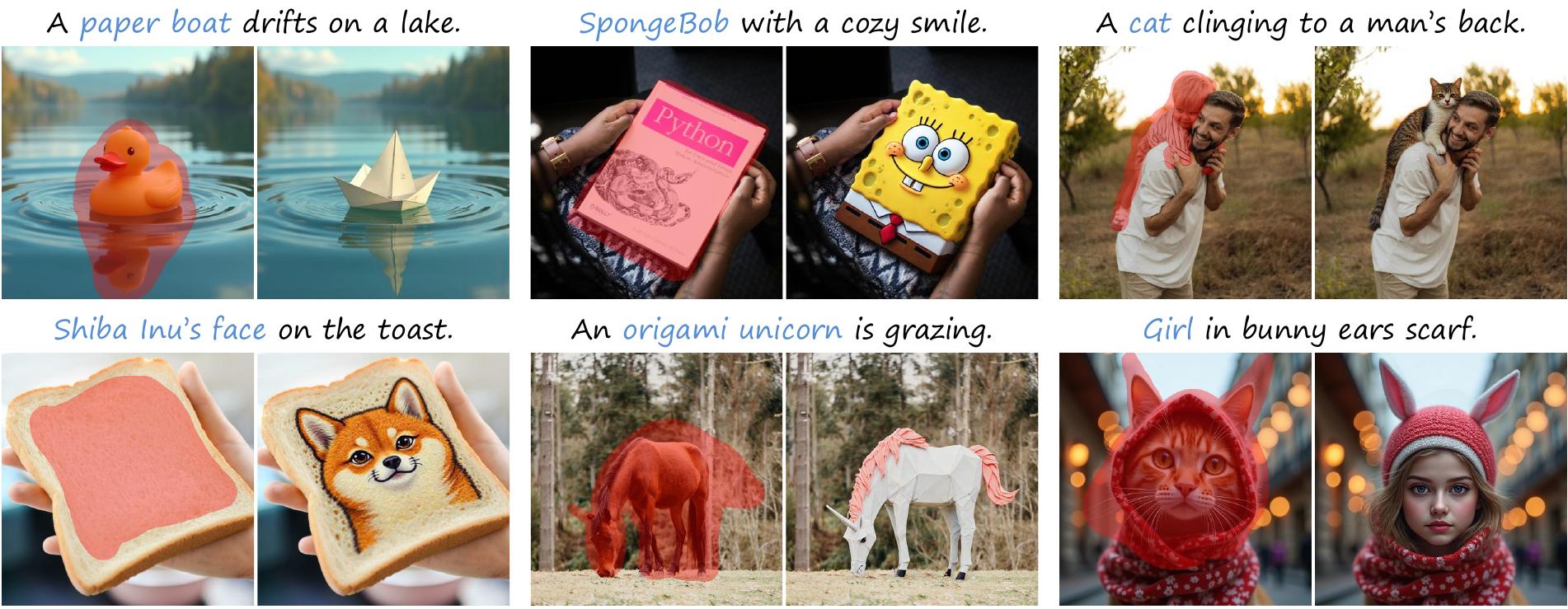}
\caption{\textbf{Results of our model.} The inpainting outputs generated by our model are visually coherent, semantically aligned with text prompts, and consistent with human aesthetic preferences.}
\end{figure*}
\begin{abstract}
This paper investigates image inpainting with preference alignment. Instead of introducing a novel method, we go back to basics and revisit fundamental problems in achieving such alignment. We leverage the prominent direct preference optimization approach for alignment training and employ public reward models to construct preference training datasets. Experiments are conducted across nine reward models, two benchmarks, and two baseline models with varying structures and generative algorithms. Our key findings are as follows: (1) Most reward models deliver valid reward scores for constructing preference data, even if some of them are not reliable evaluators. (2) Preference data demonstrates robust trends in both candidate scaling and sample scaling across models and benchmarks. (3) Observable biases in reward models, particularly in brightness, composition, and color scheme, render them susceptible to cause reward hacking. (4) A simple ensemble of these models yields robust and generalizable results by mitigating such biases. Built upon these observations, our alignment models significantly outperform prior models across standard metrics, GPT-4 assessments, and human evaluations, without any changes to model structures or the use of new datasets. We hope our work can set a simple yet solid baseline, pushing this promising frontier. Our code is open-sourced at: \url{https://github.com/shenytzzz/Follow-Your-Preference}.
\end{abstract}

\section{Introduction}

Image inpainting~\citep{inpaint} aims to fill in user-specified regions of an image in a visually coherent and realistic manner. It holds great value in applications such as photo restoration~\citep{restoration}, content creation~\citep{powerpaint}, and image editing~\citep{magicbrush}. With the unprecedented success of diffusion models~\citep{DDPM} and flow-based models~\citep{FlowMatching}, image inpainting has become a prominent research focus in recent years.

Aligning human preferences in visual generation has emerged as a focal point of research efforts~\citep{Diffusion-DPO, dancegrpo, DDPO, dpok}. While great progress has been made in image inpainting~\citep{PreAliSurvey, BrushNet, powerpaint, HD-Painter}, 
research on aligning inpainting results with human preferences remains limited.

This paper explores image inpainting with preference alignment. Given the limited work on this task, our goal is not to present a novel method, but rather to rethink foundational questions.
To this end, we adopt the prominent Direct Preference Optimization (DPO)~\citep{DPO, Diffusion-DPO, flow-dpo} to conduct studies due to its simplicity and efficiency. Instead of relying on costly and non-scalable human annotations, we employ public, off-the-shelf reward models for constructing preference training datasets. Our study focuses on several key questions: (1) How \textit{effective} are these reward models in scoring and constructing high-quality preference data? (2) How \textit{scalable} is preference data with respect to the candidate quantity and the sample quantity? (3) How does \textit{reward hacking}~\citep{hack} occur, and what method can be used to mitigate it?

To answer these questions, we conduct experiments across \textit{nine widely used reward models} (e.g., HPSv2~\citep{hpsv2}, PickScore~\citep{pickscore}), \textit{two representative evaluation benchmarks} (BrushBench~\citep{BrushNet}, EditBench~\citep{EditBench}), and \textit{two baseline inpainting models} (BrushNet~\citep{BrushNet}, FLUX.1 Fill~\citep{flux1filldev}) with diverse architectures (U-Net~\citep{UNet}, Transformer-based~\citep{Transformer}) and generative algorithms (diffusion~\citep{DDPM}, flow-based~\citep{FlowMatching}). Our findings reveal that: \textbf{(1)} Most reward models provide \textit{valid} reward signals for constructing effective preference training data, despite some being unreliable as evaluators and exhibiting shared biases. \textbf{(2)} Preference data shows \textit{consistent trends} in both candidate scaling and sample scaling across baseline models and benchmarks. However, biases in certain reward models (e.g., HPSv2) can lead to reward hacking, undermining scaling effectiveness. \textbf{(3)} We identify explicit biases in reward models---particularly in \textit{brightness}, \textit{composition}, and \textit{color scheme}---making them vulnerable to reward hacking. For example, HPSv2 tends to favor images with bright lighting, complex composition with rich details, and vivid colors; PickScore shows the opposite tendency. We also find that BrushNet generates vibrant images, making PickScore suitable for it; while FLUX.1 Fill produces plain images, aligning well with the property of HPSv2. \textbf{(4)} A simple ensemble of these reward models exhibits strong versatility across models, producing balanced and aesthetically pleasing inpainting results.

Building on these observations, we propose simple yet effective models via reward ensemble. Without modifying model architectures or introducing new datasets, our models substantially outperform state-of-the-art models---across standard metrics, GPT-4 assessments, and human evaluations. Visualizations show that our models generate more coherent and visually appealing results than competitors. We hope our work can establish a simple yet strong baseline to advance this research field.

\section{Related Work}
\label{sec:related-work}

\textbf{Image inpainting}~\citep{inpaint} aims to fill in missing or damaged regions of an image.
Recently, great progress~\citep{HD-Painter, ASUKA, follow-your-canvas, follow-your-emoji, emoji2, csac, ceg, dsbf} has been achieved by employing diffusion models~\citep{DDPM, DDIM} and flow-based models~\citep{FlowMatching, hap, Wan}. 
Some previous works make the first attempt~\citep{BLD, sd, CNI} to achieve it, and others~\citep{BrushNet, powerpaint, StructureMatters} later advance and refine it. For example, BrushNet~\citep{BrushNet} introduces a dual-branch diffusion model that decouples masked image feature extraction from generation. 
FLUX.1 Fill~\citep{flux1filldev} employs rectified flow transformer for image inpainting.

\textbf{Image generation with preference alignment} seeks to align synthesized images with human preferences~\citep{PreAliSurvey}. Some previous works~\citep{DDPO, dpok} employ reinforcement learning~\citep{rl}, while recent approaches explore direct preference optimization~\citep{DPO}. For instance, Diffusion-DPO~\citep{Diffusion-DPO} optimizes pairwise feedback via a diffusion-aware extension of DPO. PrefPaint~\citep{Prefpaint} aligns image inpainting results with human preferences by using a reward model trained on human-annotated data.

\section{Preliminaries}

\subsection{Diffusion Models and Flow-Based Models}
Diffusion models~\citep{thermodynamics, DDIM}, such as \textbf{DDPM}~\citep{DDPM}, are a class of generative models that learn to reverse a gradual noise corruption process. DDPM assumes a forward process that gradually applies noise to real data. At timestep $t$, the real data $x_0$ is destroyed to $x_t$: $q(x_t|x_0)=\mathcal{N}(x_t;\sqrt{\bar{\alpha}_t}x_0,(1-\bar{\alpha}_t)\mathbf{I})$, where $\bar{\alpha}_t$ is noise scheduling hyper-parameters. It has a reparameterization formula: $x_t=\sqrt{\bar{\alpha}_t}x_0 + \sqrt{1-\bar{\alpha}_t}\epsilon$, where noise $\epsilon\sim\mathcal{N}(0, \mathbf{I})$. DDPM learns a reverse process using a denoising model $\epsilon_{\theta}$ with parameters $\theta$, inverting the forward process: $p_{\theta}(x_{t-1}|x_t)=\mathcal{N}(\mu_{\theta}(x_t),\Sigma_{\theta}(x_t))$. The denoising model $\epsilon_{\theta}$ can be trained by minimizing:
\begin{equation}
    \mathcal{L}_{\mathrm{DDPM}}=\mathbb{E}_{t, x_0,\epsilon}\left[|| \epsilon - \epsilon_{\theta}(\sqrt{\bar{\alpha}_t}x_0 + \sqrt{1-\bar{\alpha}_t}\epsilon, t)||^2\right].
\label{eq:ddpm}
\end{equation}
Flow-based models~\citep{SD3} are generative models that learn to model data distributions using invertible transformations. Recently, \textbf{Flow Matching}~\citep{FlowMatching} has emerged as a prominent approach for visual generation~\citep{Sit, Seedream}. It usually learns a continuous-time flow that transforms a simple prior distribution into the data distribution by solving an ODE. The process, with an optimal-transport path, employs a linear interpolation scheme: $x_t=(1-t)x_0+t\epsilon$. A denoising model $v_{\theta}$ is trained to predict the velocity field by minimizing:
\begin{equation}
\mathcal{L}_{\mathrm{FlowMatching}}=\mathbb{E}_{t,x_0,\epsilon}\left[|| v_{\theta}((1-t)x_0+t\epsilon, t)-(\epsilon - x_0) ||^2\right].
\label{eq:fm}
\end{equation}
\textbf{U-Net}~\citep{UNet} is used as the basic model structure by many previous denoising models~\citep{DDPM, DDIM}. U-Net is a symmetric encoder-decoder architecture that captures multi-scale features through progressive downsampling and upsampling. \textbf{Transformers}~\citep{Transformer}, employed in recent works~\citep{Seedance, Seedream, Wan, HunyuanVideo}, process all data elements in parallel using attention, facilitating training scalibility.

To improve the reliability and generalization of conclusions drawn in our studies, we conduct investigations using two different \textbf{baseline models}---\textbf{BrushNet}~\citep{BrushNet} and \textbf{FLUX.1 Fill}~\citep{flux1filldev}, introduced in~\autoref{sec:related-work}. BrushNet is built on a U-Net-like architecture and trained with the DDPM loss, while FLUX leverages transformers and learns via Flow Matching.

\subsection{Preference Alignment}
The standard pipeline for training large-scale models typically involves pre-training, supervised fine-tuning, and preference alignment. Preference alignment refines model outputs to better match human values. Reinforcement Learning from Human Feedback (RLHF)~\citep{RLHF} is a popular alignment approach. It utilizes human preferences on model outputs to train a \textit{separate reward model}, which subsequently provides rewards for alignment via reinforcement learning algorithms such as PPO~\citep{PPO} and GRPO~\citep{GRPO}. In comparison, \textbf{Direct Preference Optimization (DPO)}~\citep{DPO}, which performs direct supervised learning, offers higher training efficiency. It constructs a preference dataset that comprises \textit{preferred samples} and \textit{dispreferred samples}. DPO learns human preferences implicitly contained within the data by maximizing:
\begin{equation}
    \mathbb{E}_{x,y^w,y^l}[\log\sigma(\beta\log\frac{\pi_{\theta}(y^w|x)}{\pi_{\mathrm{ref}}(y^w|x)}-\beta\log\frac{\pi_{\theta}(y^l|x)}{\pi_{\mathrm{ref}}(y^l|x)})],
\label{eq:dpo}
\end{equation}
where $\sigma$ is the sigmoid function; $\pi_{\theta}$ and $\pi_{\mathrm{ref}}$ are the \textit{policy} and the \textit{reference policy} respectively. In image generation, given a text prompt $x$, $y^w$ and $y^l$ denote the generated preferred image and dispreferred image, respectively. The hyper-parameter $\beta$ controls the strength of regularization: a large value of $\beta$ increases regularization pressure, dampening preference learning. In visual generation, \autoref{eq:dpo} can be derived to yield a simplified loss~\citep{Diffusion-DPO, flow-dpo}:
\begin{equation}
    \mathcal{L}_{\mathrm{DPO}}=-\mathbb{E}[\log\sigma(-\beta((\mathcal{L}_{\theta}^{w}-\mathcal{L}_{\mathrm{ref}}^{w})-(\mathcal{L}_{\theta}^{l}-\mathcal{L}_{\mathrm{ref}}^{l})))],
\label{eq:dpo-loss}
\end{equation}
where $\mathcal{L}_{\theta}^{w}$ and $\mathcal{L}_{\theta}^{l}$ denote the loss (\autoref{eq:ddpm} or \autoref{eq:fm}) applied to the policy on preferred samples and dispreferred samples, respectively; similarly, $\mathcal{L}_{\mathrm{ref}}^{w}$ and $\mathcal{L}_{\mathrm{ref}}^{l}$ denote the loss applied to the reference policy. This loss function aligns the distribution of generated samples with the preferred data distribution and diverges from the dispreferred distribution. Due to the simplicity, efficiency, and stability of DPO, \textit{this paper will explore preference alignment for image inpainting by optimizing~\autoref{eq:dpo-loss} on different preference datasets that are constructed for investigation.}

\subsection{Reward Models}
Reward models play an important role in preference alignment: they provide real-time rewards in RLHF~\citep{PPO}, and offer scores for constructing offline preference data in DPO~\citep{unifiedreward,capo}. However, prior works~\citep{SimpleAR, dancegrpo} directly employ off-the-shelf reward models for visual preference alignment without sufficient evaluations. In this paper, we evaluate the effectiveness of these reward models in constructing preference data via extensive studies. Specifically, we examine the following \textbf{public reward models}:
(1) \textbf{CLIPScore}~\citep{clipscore} measures semantic alignment between images and text prompts by calculating cosine similarities of their CLIP embeddings~\citep{CLIP}.
(2) \textbf{Aesthetic}~\citep{aesthetic} predicts human aesthetic preferences on top of the CLIP embeddings.
(3) \textbf{ImageReward (ImageR)}~\citep{imagereward} is trained by fine-tuning BLIP~\citep{BLIP} on 137K preference samples. 
(4) \textbf{PickScore}~\citep{pickscore} is a CLIP-based image scoring model, trained on over 500K synthesized image samples with users' preference choices.
(5) \textbf{HPSv2}~\citep{hpsv2} is also a CLIP-based model that evaluates both image quality and text-image alignment by learning from 798K human preferences on 433K sample pairs.
(6) \textbf{VQAScore}~\citep{vqascore} provides a semantic alignment score by computing the probability of a VQA model answering ``yes'' to each question: ``Does this figure show \{text\}?''.
(7) \textbf{UnifiedReward (UnifiedR)}~\citep{unifiedreward} is a unified model that assesses both visual generation and understanding.
(8) \textbf{Perception Encoder (Perception)}~\citep{perception} is trained by contrastive visual-language pre-training, producing semantically aligned multimodal embeddings.
(9) \textbf{HPSv3}~\citep{hpsv3} is trained on 1.5M annotated sample pairs using Qwen2VL-7B~\citep{qwen2-vl}.

To assess their efficacy, these models are employed to assign reward scores to candidate samples which are generated by the baseline models with different random seeds. The resulting highest- and lowest-scoring samples from each text prompt are subsequently utilized as the preferred and dispreferred samples for DPO training. Based on the evaluation of training results, the top-performing ones are designated as the most effective reward models to provide accurate rewards, and vice versa.

\section{How Effective are Reward Models?}
\label{sec:effective}

\begin{table*}[t]
\centering
\caption{Comparisons of reward models using \textbf{BrushNet} on BrushBench and EditBench.}
\resizebox{1.\linewidth}{!}{
\begin{threeparttable}  
\renewcommand{\arraystretch}{1.8}
\setlength{\tabcolsep}{-0.5pt}
{
     \begin{tabular}{lC{1.2cm}C{1.2cm}C{1.2cm}C{1.2cm}C{1.2cm}C{1.2cm}C{1.2cm}C{1.2cm}C{1.2cm}C{1.2cm}C{1.2cm}C{1.2cm}C{1.2cm}C{1.2cm}C{1.2cm}C{1.2cm}C{1.2cm}C{1.2cm}C{1.2cm}C{1.2cm}}
          \toprule
          \multirow{2}{*}{reward model}    & \multicolumn{2}{c}{{CLIPScore}}   & \multicolumn{2}{c}{{Aesthetic}}  & \multicolumn{2}{c}{{ImageR}} & \multicolumn{2}{c}{{PickScore}}  & \multicolumn{2}{c}{{HPSv2}}   & \multicolumn{2}{c}{{VQAScore}} & \multicolumn{2}{c}{{UnifiedR}} & \multicolumn{2}{c}{{Perception}} & \multicolumn{2}{c}{{HPSv3}} & \multicolumn{2}{c}{\cellcolor{sem!15}{GPT-4}}     \\ 
                                             \cmidrule(lr){2-3} \cmidrule(lr){4-5} \cmidrule(lr){6-7} \cmidrule(lr){8-9} \cmidrule(lr){10-11} \cmidrule(lr){12-13} \cmidrule(lr){14-15} \cmidrule(lr){16-17} \cmidrule(lr){18-19} \cmidrule(lr){20-21}
                                        & Brush. & Edit. & Brush. & Edit. & Brush. & Edit. & Brush. & Edit. & Brush. & Edit. & Brush. & Edit. & Brush. & Edit. & Brush. & Edit. & Brush. & Edit. & \cellcolor{sem!15}Brush. & \cellcolor{sem!15}Edit. \\
          \midrule
  \color{gray}Baseline &     \color{gray}26.415 &     \color{gray}27.337 &   \color{gray}6.425 &   \color{gray}5.392 &     \color{gray}12.717 &     \color{gray}-1.296 &     \color{gray}22.133 &   \color{gray}20.616 &     \color{gray}27.509 &     \color{gray}23.076 &             \color{gray}{9.060} &                 \color{gray}6.770 &   \color{gray}3.303 &   \color{gray}2.100 &     \color{gray}26.290 &     \color{gray}26.410 &     \color{gray}5.749 &   \color{gray}0.403 &     \cellcolor{sem!15}\color{gray}79.391 &     \cellcolor{sem!15}\color{gray}57.046 \\
    \color{gray}Random &     \color{gray}26.441 &     \color{gray}27.631 &   \color{gray}6.424 &   \color{gray}5.392 &     \color{gray}12.685 &     \color{gray}-1.136 &     \color{gray}22.130 &   \color{gray}20.642 &     \color{gray}27.501 &     \color{gray}23.067 &               \color{gray}9.050 &        \color{gray}\textbf{6.917} &   \color{gray}3.302 &   \color{gray}2.110 &     \color{gray}26.292 &     \color{gray}26.422 &     \color{gray}5.738 &   \color{gray}0.425 &     \cellcolor{sem!15}\color{gray}79.177 &     \cellcolor{sem!15}\color{gray}56.753 \\
             CLIPScore &                 26.461 &        \textbf{27.710} &               6.430 &               5.393 &                 12.782 &                 -0.720 &                 22.146 &               20.680 &                 27.582 &                 23.316 &                         {9.062} &   \underline{\color{myblue}6.894} &               3.341 &               2.124 &                 26.324 &     \underline{26.548} &                 5.777 &               0.640 &                 \cellcolor{sem!15}79.661 &                 \cellcolor{sem!15}57.539 \\
             Aesthetic &                 26.465 &   \color{myblue}27.355 &   \underline{6.477} &      \textbf{5.520} &     \underline{12.994} &                 -0.877 &                 22.221 &               20.689 &                 27.594 &                 23.166 &                         {9.065} &               \color{myblue}6.828 &               3.343 &               2.140 &                 26.293 &   \color{myblue}26.342 &                 5.922 &               0.597 &                 \cellcolor{sem!15}81.603 &                 \cellcolor{sem!15}58.603 \\
                ImageR &                 26.471 &   \color{myblue}27.539 &               6.462 &               5.434 &                 12.891 &               {-0.377} &                 22.153 &               20.701 &                 27.672 &               {23.467} &             \color{myblue}9.036 &               \color{myblue}6.761 &               3.334 &               2.144 &                 26.305 &                 26.501 &                 5.913 &               0.782 &                 \cellcolor{sem!15}80.341 &                 \cellcolor{sem!15}57.806 \\
             PickScore &   \color{myblue}26.397 &   \color{myblue}27.199 &               6.454 &               5.454 &                 12.893 &   \color{myblue}-1.364 &        \textbf{22.254} &   \underline{20.732} &   \color{myblue}27.322 &   \color{myblue}22.933 &                         {9.062} &               \color{myblue}6.873 &   \underline{3.353} &      \textbf{2.178} &   \color{myblue}26.273 &                 26.427 &                 5.750 &               0.469 &        \cellcolor{sem!15}\textbf{82.726} &        \cellcolor{sem!15}\textbf{59.550} \\
                 HPSv2 &     \underline{26.481} &     \underline{27.677} &               6.476 &             {5.495} &                 12.890 &         \textbf{0.128} &                 22.137 &             {20.725} &        \textbf{27.818} &        \textbf{23.742} &                  \textbf{9.073} &               \color{myblue}6.818 &               3.332 &               2.155 &        \textbf{26.361} &        \textbf{26.678} &                 5.979 &      \textbf{1.061} &                 \cellcolor{sem!15}79.914 &                 \cellcolor{sem!15}57.658 \\
              VQAScore &                 26.442 &   \color{myblue}27.524 &               6.429 &               5.407 &   \color{myblue}12.658 &                 -0.800 &   \color{myblue}22.126 &               20.667 &                 27.527 &                 23.234 &             \color{myblue}9.038 &               \color{myblue}6.879 &               3.311 &               2.139 &                 26.326 &   \color{myblue}26.406 &   \color{myblue}5.723 &               0.555 &   \color{myblue}\cellcolor{sem!15}78.877 &   \color{myblue}\cellcolor{sem!15}56.975 \\
              UnifiedR &   \color{myblue}26.428 &   \color{myblue}27.505 &               6.433 &               5.402 &                 12.764 &                 -0.812 &                 22.157 &               20.675 &                 27.562 &                 23.204 &                         {9.061} &               \color{myblue}6.857 &               3.329 &               2.155 &                 26.320 &                 26.436 &                 5.800 &               0.540 &                 \cellcolor{sem!15}80.333 &                 \cellcolor{sem!15}57.185 \\
            Perception &                 26.448 &   \color{myblue}27.484 &               6.428 &               5.393 &                 12.789 &                 -0.973 &                 22.177 &               20.660 &                 27.519 &                 23.111 &               \underline{9.069} &   \underline{\color{myblue}6.894} &               3.327 &             {2.160} &                 26.310 &                 26.515 &                 5.764 &               0.433 &                 \cellcolor{sem!15}80.277 &                 \cellcolor{sem!15}57.254 \\
                 HPSv3 &                 26.461 &   \color{myblue}27.547 &               6.464 &               5.448 &                 12.922 &     \underline{-0.146} &                 22.176 &               20.713 &                 27.758 &                 23.491 &                         {9.065} &               \color{myblue}6.850 &               3.344 &               2.158 &                 26.317 &                 26.535 &     \underline{6.014} &               0.863 &                 \cellcolor{sem!15}80.623 &                 \cellcolor{sem!15}57.485 \\
              Ensemble &        \textbf{26.535} &   \color{myblue}27.398 &      \textbf{6.485} &   \underline{5.497} &        \textbf{13.037} &                 -0.352 &     \underline{22.229} &      \textbf{20.735} &     \underline{27.797} &     \underline{23.522} &             \color{myblue}9.053 &             {\color{myblue}6.892} &      \textbf{3.365} &   \underline{2.176} &     \underline{26.338} &     \underline{26.603} &        \textbf{6.074} &   \underline{1.015} &     \cellcolor{sem!15}\underline{82.172} &     \cellcolor{sem!15}\underline{58.986} \\
          \bottomrule
     \end{tabular}
     \begin{tablenotes}
     \item \textbf{Bold} values denote the best results. \underline{Underlined} values denote the second-best results. Values in {\color{myblue}blue} denote the results below the baseline or random chance.
     \end{tablenotes}
}
\end{threeparttable}
}
\label{tab:brushnet_metrics_8cols_twobench}
\vspace{-4mm}
\end{table*}

\begin{table*}[t]
\centering
\caption{Comparisons of reward models using \textbf{FLUX.1 Fill} on BrushBench and EditBench.}
\resizebox{1.\linewidth}{!}{
\begin{threeparttable}
\renewcommand{\arraystretch}{1.8}
\setlength{\tabcolsep}{-0.5pt}
{
     \begin{tabular}{lC{1.2cm}C{1.2cm}C{1.2cm}C{1.2cm}C{1.2cm}C{1.2cm}C{1.2cm}C{1.2cm}C{1.2cm}C{1.2cm}C{1.2cm}C{1.2cm}C{1.2cm}C{1.2cm}C{1.2cm}C{1.2cm}C{1.2cm}C{1.2cm}C{1.2cm}C{1.2cm}}
          \toprule
          \multirow{2}{*}{reward model}    & \multicolumn{2}{c}{{CLIPScore}}   & \multicolumn{2}{c}{{Aesthetic}}  & \multicolumn{2}{c}{{ImageR}} & \multicolumn{2}{c}{{PickScore}}  & \multicolumn{2}{c}{{HPSv2}}   & \multicolumn{2}{c}{{VQAScore}} & \multicolumn{2}{c}{{UnifiedR}} & \multicolumn{2}{c}{{Perception}} & \multicolumn{2}{c}{{HPSv3}} & \multicolumn{2}{c}{\cellcolor{sem!15}{GPT-4}}     \\ 
                                             \cmidrule(lr){2-3} \cmidrule(lr){4-5} \cmidrule(lr){6-7} \cmidrule(lr){8-9} \cmidrule(lr){10-11} \cmidrule(lr){12-13} \cmidrule(lr){14-15} \cmidrule(lr){16-17} \cmidrule(lr){18-19} \cmidrule(lr){20-21}
                                        & Brush. & Edit. & Brush. & Edit. & Brush. & Edit. & Brush. & Edit. & Brush. & Edit. & Brush. & Edit. & Brush. & Edit. & Brush. & Edit. & Brush. & Edit. & \cellcolor{sem!15}Brush. & \cellcolor{sem!15}Edit. \\
          \midrule
  \color{gray}Baseline &     \color{gray}26.244 &     \color{gray}27.103 &   \color{gray}6.429 &   \color{gray}5.458 &     \color{gray}12.760 &     \color{gray}4.910 &   \color{gray}22.327 &     \color{gray}21.211 &   \color{gray}27.476 &   \color{gray}24.076 &                 \color{gray}9.081 &       \color{gray}8.012 &     \color{gray}3.360 &   \color{gray}2.485 &   \color{gray}25.945 &     \color{gray}26.834 &     \color{gray}6.055 &     \color{gray}2.470 &     \cellcolor{sem!15}\color{gray}83.935 &     \cellcolor{sem!15}\color{gray}66.979 \\
    \color{gray}Random &     \color{gray}26.239 &     \color{gray}27.078 &   \color{gray}6.431 &   \color{gray}5.459 &     \color{gray}12.772 &     \color{gray}4.955 &   \color{gray}22.328 &     \color{gray}21.211 &   \color{gray}27.475 &   \color{gray}24.100 &                 \color{gray}9.077 &     \color{gray}{8.030} &     \color{gray}3.356 &   \color{gray}2.491 &   \color{gray}25.944 &     \color{gray}26.838 &     \color{gray}6.056 &     \color{gray}2.490 &     \cellcolor{sem!15}\color{gray}83.517 &     \cellcolor{sem!15}\color{gray}66.942 \\
             CLIPScore &   \color{myblue}26.233 &   \color{myblue}27.072 &               6.432 &               5.477 &                 12.791 &                 4.997 &               22.329 &                 21.215 &               27.487 &               24.121 &             {\color{myblue}9.071} &          \textbf{8.071} &                 3.361 &               2.512 &               25.948 &                 26.859 &                 6.056 &                 2.499 &                 \cellcolor{sem!15}83.942 &                 \cellcolor{sem!15}66.997 \\
             Aesthetic &     \underline{26.250} &     \underline{27.200} &               6.432 &   \underline{5.478} &                 12.823 &     \underline{5.175} &               22.337 &                 21.219 &               27.520 &               24.142 &             \color{myblue}{9.075} &    \color{myblue} 8.001 &                 3.363 &               2.507 &               25.954 &                 26.878 &                 6.075 &     \underline{2.577} &                 \cellcolor{sem!15}83.950 &                 \cellcolor{sem!15}67.906 \\
                ImageR &        \textbf{26.251} &                 27.121 &               6.434 &      \textbf{5.481} &                 12.823 &                 5.001 &               22.336 &                 21.211 &               27.518 &               24.143 &             \color{myblue}{9.080} &     \color{myblue}7.977 &                 3.362 &   \underline{2.536} &               25.946 &                 26.846 &                 6.078 &                 2.550 &                 \cellcolor{sem!15}84.176 &                 \cellcolor{sem!15}67.785 \\
             PickScore &   \color{myblue}26.236 &               {27.195} &               6.436 &               5.476 &               {12.879} &                 5.134 &               22.341 &               {21.223} &               27.530 &             {24.154} &             \color{myblue}{9.076} &    \color{myblue} 8.003 &        \textbf{3.383} &               2.514 &               25.955 &     \underline{26.900} &                 6.105 &                 2.548 &                 \cellcolor{sem!15}84.188 &                 \cellcolor{sem!15}67.100 \\
                 HPSv2 &                 26.246 &                 27.160 &   \underline{6.441} &               5.475 &        \textbf{12.904} &               {5.145} &      \textbf{22.356} &        \textbf{21.232} &      \textbf{27.605} &      \textbf{24.202} &                    \textbf{9.085} &   \color{myblue}{8.028} &                 3.363 &      \textbf{2.553} &      \textbf{25.963} &               {26.895} &        \textbf{6.181} &        \textbf{2.605} &        \cellcolor{sem!15}\textbf{84.699} &        \cellcolor{sem!15}\textbf{68.186} \\
              VQAScore &   \color{myblue}26.243 &   \color{myblue}27.101 &               6.432 &               5.466 &                 12.781 &   \color{myblue}4.926 &               22.329 &                 21.215 &               27.486 &               24.104 &             {\color{myblue}9.075} &     \color{myblue}8.020 &   \color{myblue}3.353 &               2.501 &               25.950 &                 26.861 &   \color{myblue}6.046 &   \color{myblue}2.473 &   \cellcolor{sem!15}\color{myblue}83.854 &   \cellcolor{sem!15}\color{myblue}66.793 \\
              UnifiedR &   \color{myblue}26.235 &                 27.133 &               6.434 &               5.473 &   \color{myblue}12.769 &                 5.034 &               22.331 &                 21.212 &               27.485 &               24.120 &             \color{myblue}{9.076} &     \color{myblue}8.014 &               {3.366} &               2.517 &               25.954 &                 26.853 &                 6.057 &                 2.524 &                 \cellcolor{sem!15}83.950 &                 \cellcolor{sem!15}67.372 \\
            Perception &        \textbf{26.251} &   \color{myblue}27.088 &               6.433 &               5.466 &   \color{myblue}12.752 &                 4.975 &               22.331 &   \color{myblue}21.209 &               27.479 &               24.109 &                           {9.082} &     \color{myblue}8.023 &   \color{myblue}3.356 &               2.514 &               25.951 &   \color{myblue}26.818 &                 6.064 &                 2.504 &                 \cellcolor{sem!15}84.022 &                 \cellcolor{sem!15}68.024 \\
                 HPSv3 &   \color{myblue}26.238 &        \textbf{27.223} &   \underline{6.441} &               5.465 &                 12.855 &                 5.092 &               22.340 &     \underline{21.226} &               27.534 &   \underline{24.155} &                 \underline{9.083} &     \color{myblue}8.000 &     \underline{3.378} &               2.497 &   \underline{25.957} &                 26.859 &                 6.106 &                 2.568 &                 \cellcolor{sem!15}84.615 &     \cellcolor{sem!15}\underline{68.107} \\
              Ensemble &   \color{myblue}26.239 &                 27.158 &      \textbf{6.442} &               5.472 &     \underline{12.884} &        \textbf{5.239} &   \underline{22.346} &                 21.215 &   \underline{27.560} &               24.146 &                           {9.082} &       \underline{8.036} &                 3.367 &               2.507 &      \textbf{25.963} &        \textbf{26.905} &     \underline{6.151} &     \underline{2.577} &     \cellcolor{sem!15}\underline{84.628} &                 \cellcolor{sem!15}67.549 \\
          \bottomrule
     \end{tabular}
     \begin{tablenotes}
     \item \textbf{Bold} values denote the best results. \underline{Underlined} values denote the second-best results. Values in {\color{myblue}blue} denote the results below the baseline or random chance.
     \end{tablenotes}
}
\end{threeparttable}
}
\label{tab:flux_metrics_8col_twobench}
\vspace{-4mm}
\end{table*}

\begin{figure*}[t]
\begin{subfigure}[t]{\linewidth}
    \centering
    \includegraphics[width=0.195\linewidth, trim=5mm 5mm 5mm 5mm, clip]{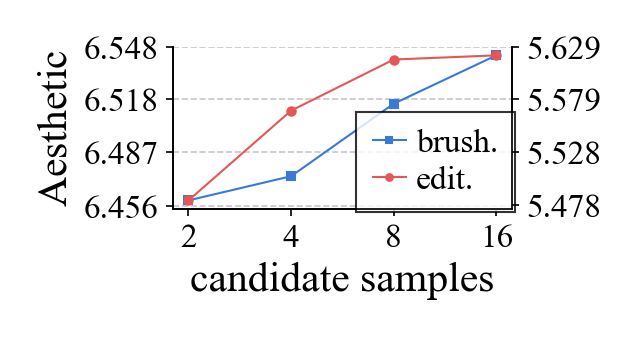}
    \includegraphics[width=0.195\linewidth, trim=5mm 5mm 5mm 5mm, clip]{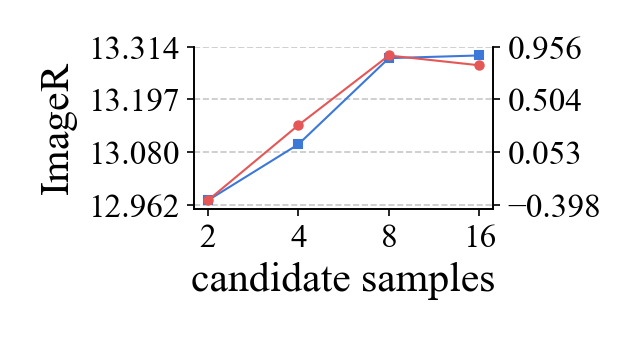}
    \includegraphics[width=0.195\linewidth, trim=5mm 5mm 5mm 5mm, clip]{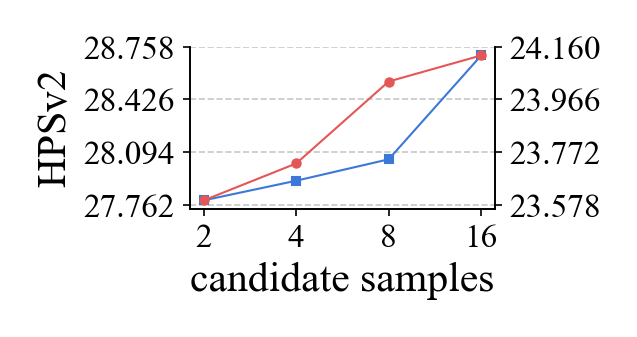}
    \includegraphics[width=0.195\linewidth, trim=5mm 5mm 5mm 5mm, clip]{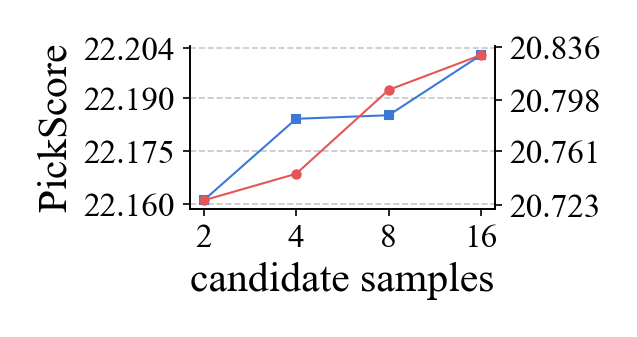}
    \includegraphics[width=0.195\linewidth, trim=5mm 5mm 5mm 5mm, clip]{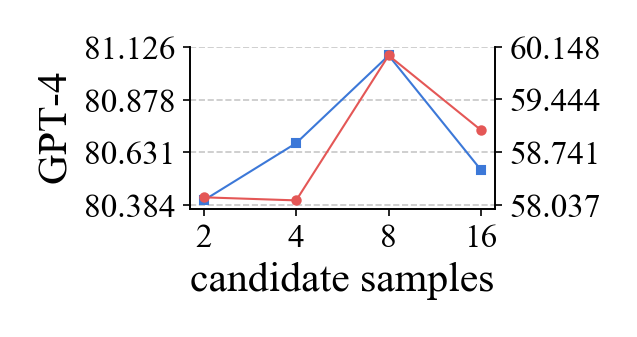}
    \includegraphics[width=0.195\linewidth, trim=5mm 5mm 5mm 5mm, clip]{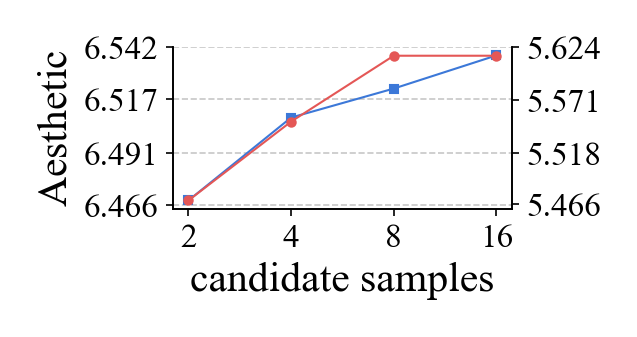}
    \includegraphics[width=0.195\linewidth, trim=5mm 5mm 5mm 5mm, clip]{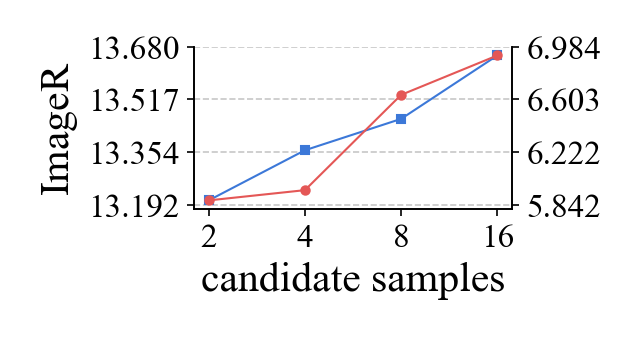}
    \includegraphics[width=0.195\linewidth, trim=5mm 5mm 5mm 5mm, clip]{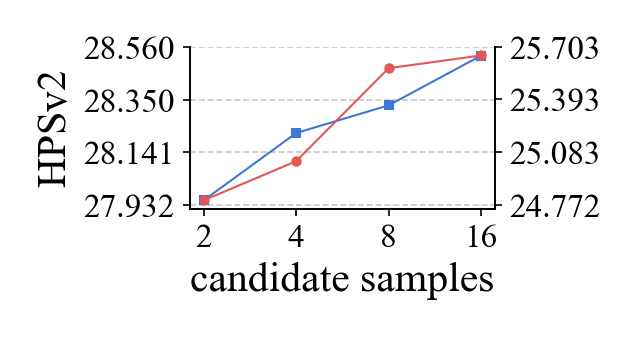}
    \includegraphics[width=0.195\linewidth, trim=5mm 5mm 5mm 5mm, clip]{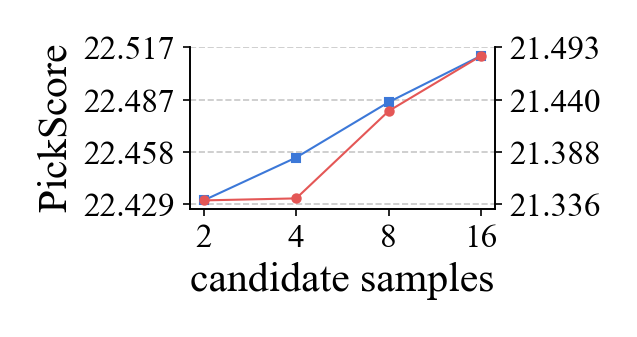}
    \includegraphics[width=0.195\linewidth, trim=5mm 5mm 5mm 5mm, clip]{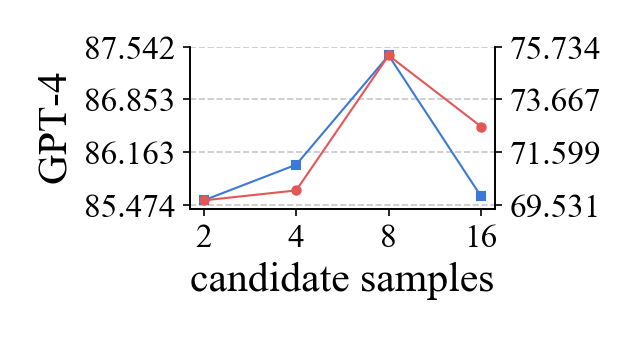}
    \vspace{-6mm}
    \caption{Candidate scaling using HPSv2.}
  \end{subfigure}
  \begin{subfigure}[t]{\linewidth}
    \centering
    % \vspace{2mm}
    \includegraphics[width=0.195\linewidth, trim=5mm 5mm 5mm 5mm, clip]{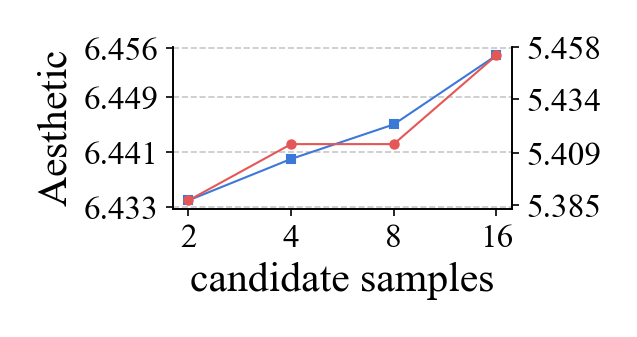}
    \includegraphics[width=0.195\linewidth, trim=5mm 5mm 5mm 5mm, clip]{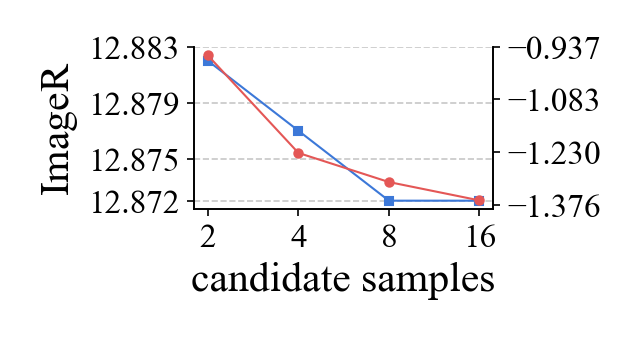}
    \includegraphics[width=0.195\linewidth, trim=5mm 5mm 5mm 5mm, clip]{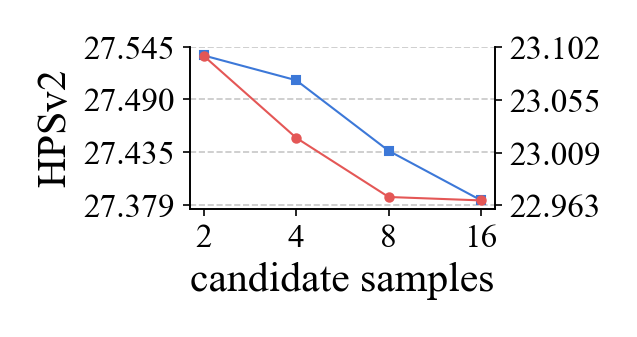}
    \includegraphics[width=0.195\linewidth, trim=5mm 5mm 5mm 5mm, clip]{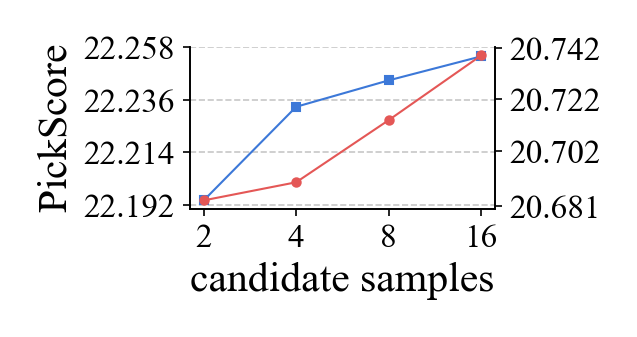}
    \includegraphics[width=0.195\linewidth, trim=5mm 5mm 5mm 5mm, clip]{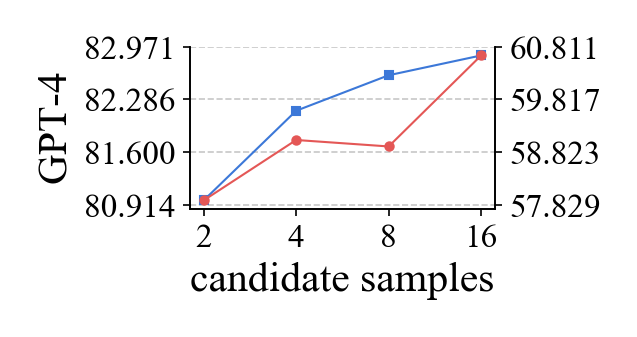}
    \includegraphics[width=0.195\linewidth, trim=5mm 5mm 5mm 5mm, clip]{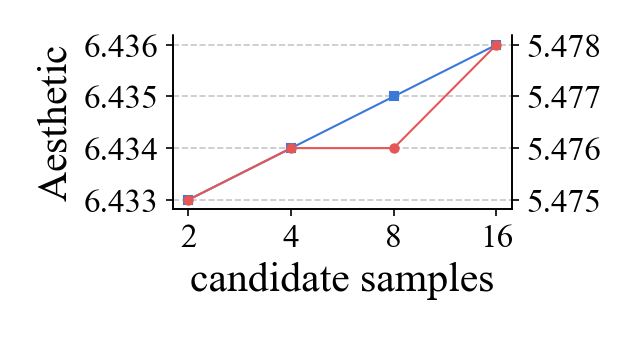}
    \includegraphics[width=0.195\linewidth, trim=5mm 5mm 5mm 5mm, clip]{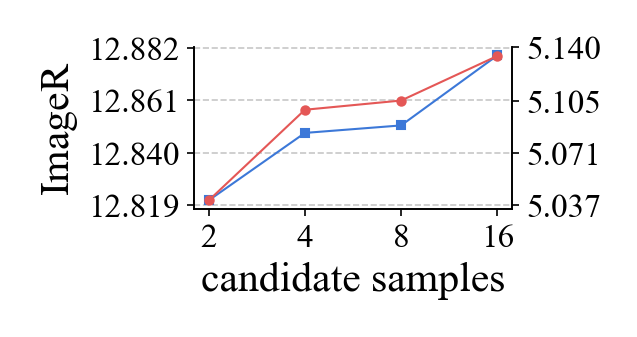}
    \includegraphics[width=0.195\linewidth, trim=5mm 5mm 5mm 5mm, clip]{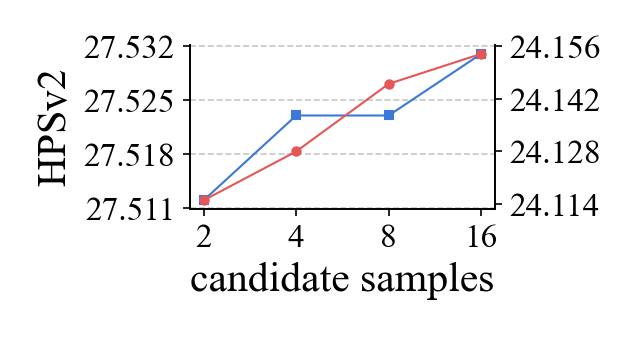}
    \includegraphics[width=0.195\linewidth, trim=5mm 5mm 5mm 5mm, clip]{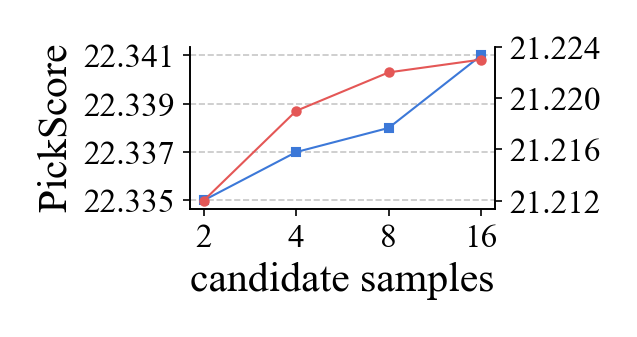}
    \includegraphics[width=0.195\linewidth, trim=5mm 5mm 5mm 5mm, clip]{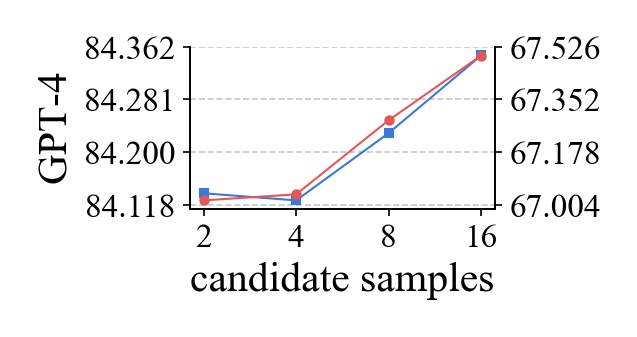}
    \vspace{-6mm}
    \caption{Candidate scaling using PickScore.}
\end{subfigure}
\begin{subfigure}[t]{\linewidth}
    \centering
    % \vspace{2mm}
    \includegraphics[width=0.195\linewidth, trim=5mm 5mm 5mm 5mm, clip]{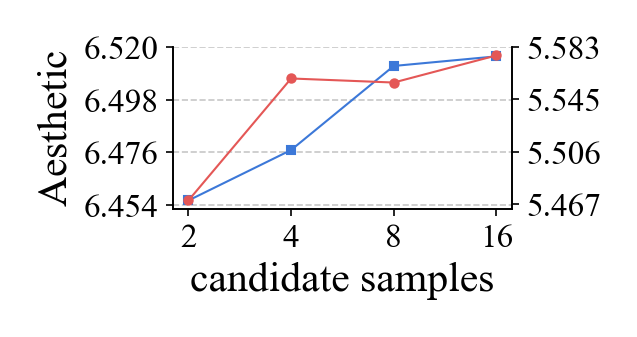}
    \includegraphics[width=0.195\linewidth, trim=5mm 5mm 5mm 5mm, clip]{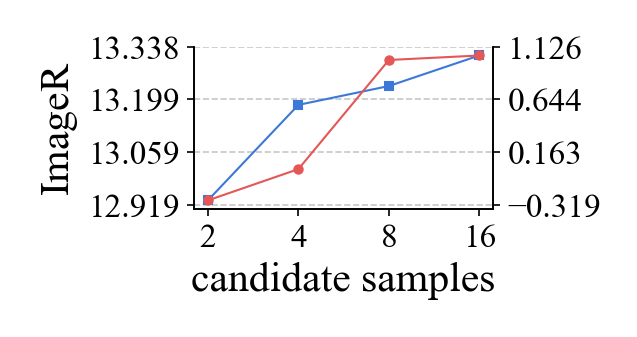}
    \includegraphics[width=0.195\linewidth, trim=5mm 5mm 5mm 5mm, clip]{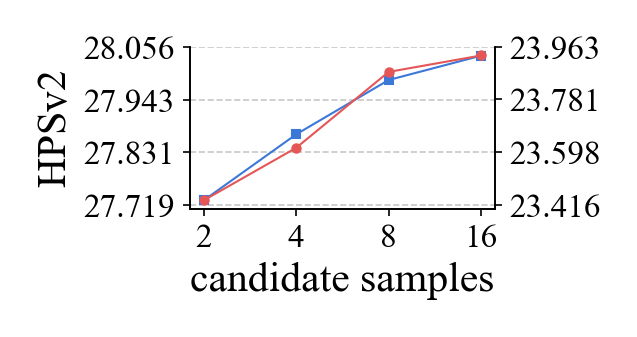}
    \includegraphics[width=0.195\linewidth, trim=5mm 5mm 5mm 5mm, clip]{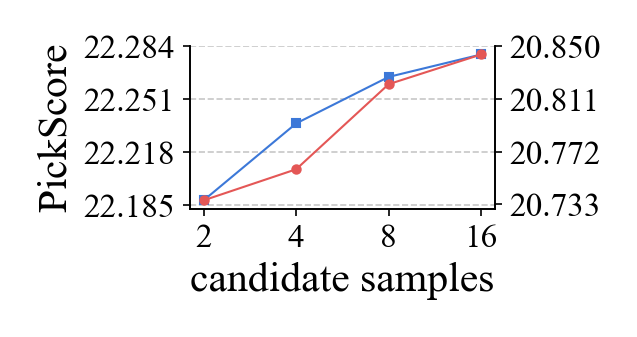}
    \includegraphics[width=0.195\linewidth, trim=5mm 5mm 5mm 5mm, clip]{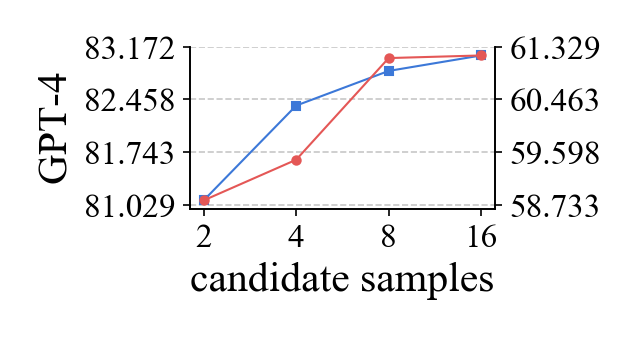}
    \includegraphics[width=0.195\linewidth, trim=5mm 5mm 5mm 5mm, clip]{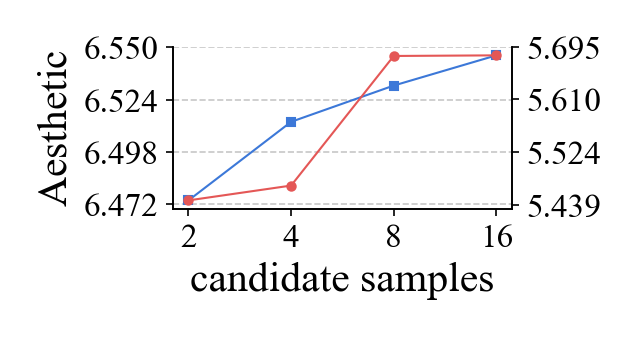}
    \includegraphics[width=0.195\linewidth, trim=5mm 5mm 5mm 5mm, clip]{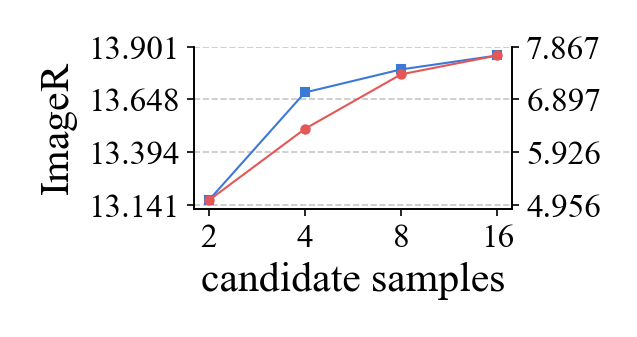}
    \includegraphics[width=0.195\linewidth, trim=5mm 5mm 5mm 5mm, clip]{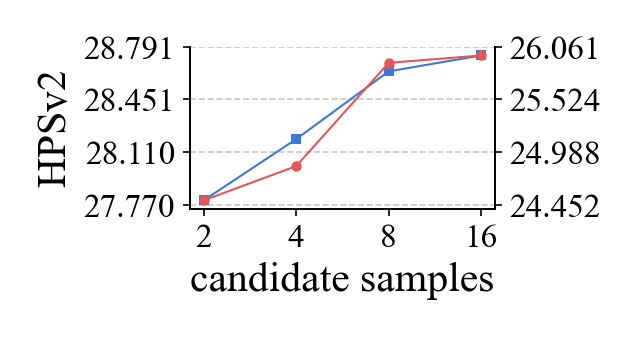}
    \includegraphics[width=0.195\linewidth, trim=5mm 5mm 5mm 5mm, clip]{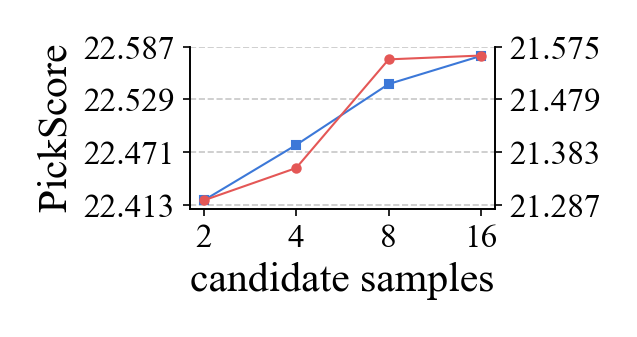}
    \includegraphics[width=0.195\linewidth, trim=5mm 5mm 5mm 5mm, clip]{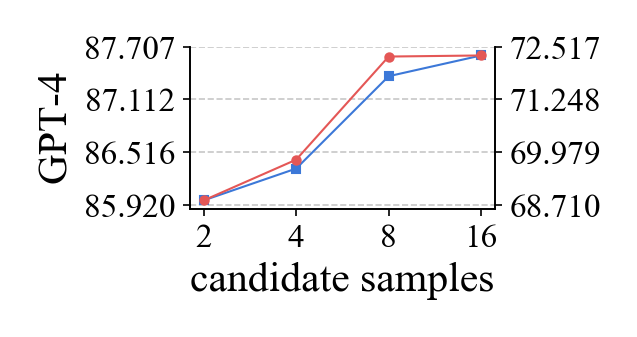}
    \vspace{-6mm}
    \caption{Candidate scaling using Ensemble.}
\end{subfigure}
  \begin{subfigure}[t]{\linewidth}
    \centering
    % \vspace{2mm}
    \includegraphics[width=0.195\linewidth, trim=5mm 5mm 5mm 5mm, clip]{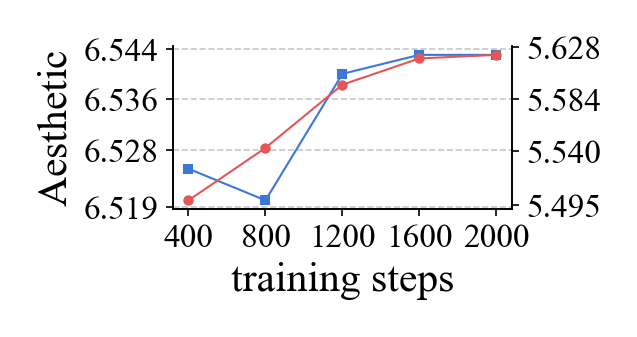}
    \includegraphics[width=0.195\linewidth, trim=5mm 5mm 5mm 5mm, clip]{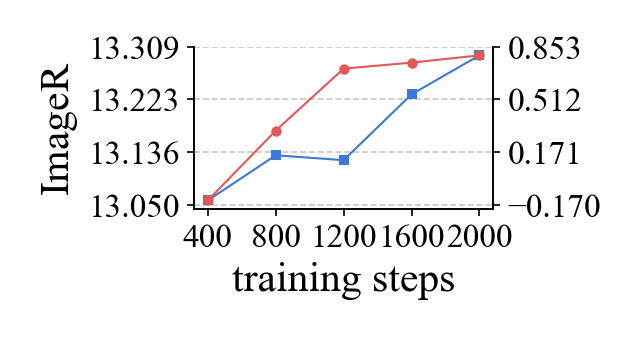}
    \includegraphics[width=0.195\linewidth, trim=5mm 5mm 5mm 5mm, clip]{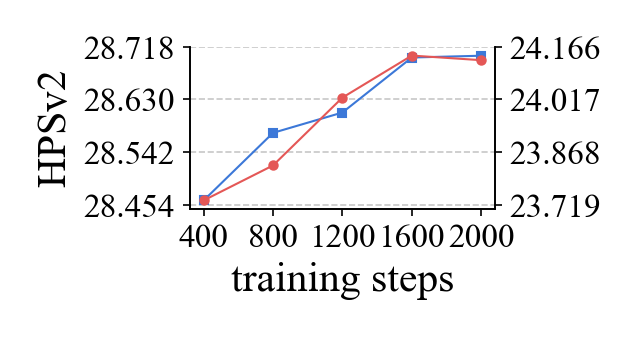}
    \includegraphics[width=0.195\linewidth, trim=5mm 5mm 5mm 5mm, clip]{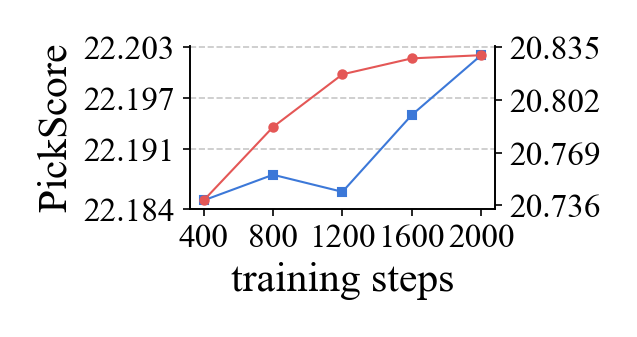}
    \includegraphics[width=0.195\linewidth, trim=5mm 5mm 5mm 5mm, clip]{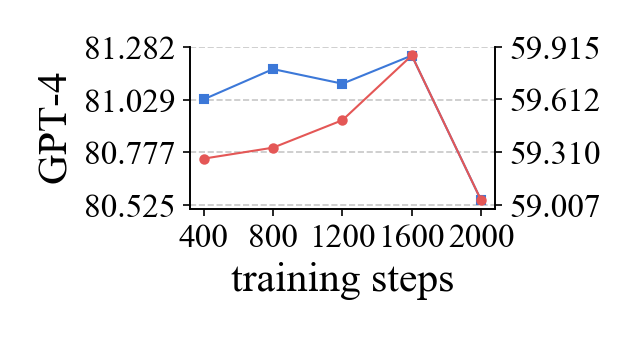}
    \includegraphics[width=0.195\linewidth, trim=5mm 5mm 5mm 5mm, clip]{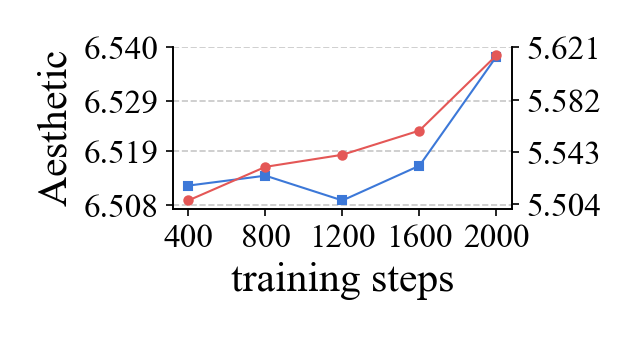}
    \includegraphics[width=0.195\linewidth, trim=5mm 5mm 5mm 5mm, clip]{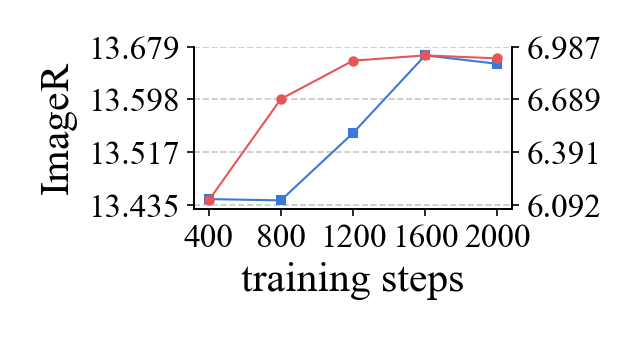}
    \includegraphics[width=0.195\linewidth, trim=5mm 5mm 5mm 5mm, clip]{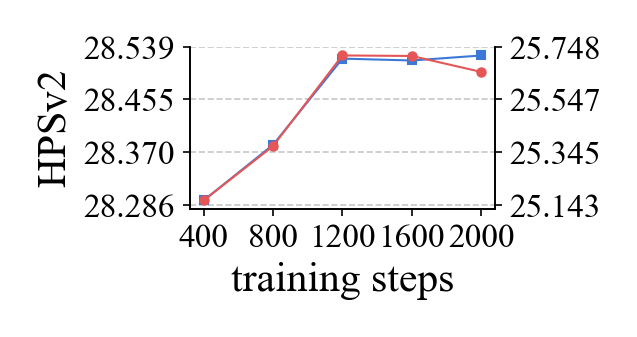}
    \includegraphics[width=0.195\linewidth, trim=5mm 5mm 5mm 5mm, clip]{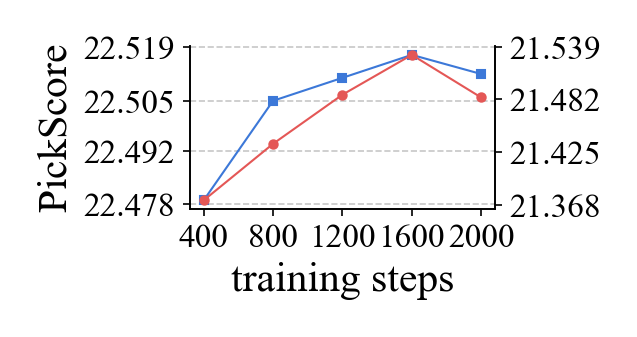}
    \includegraphics[width=0.195\linewidth, trim=5mm 5mm 5mm 5mm, clip]{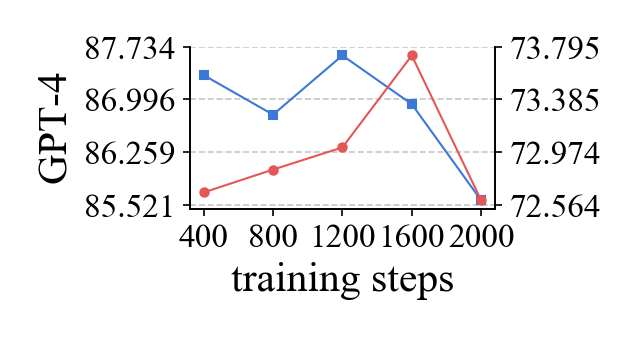}
    \vspace{-6mm}
    \caption{Sample scaling using HPSv2.}
  \end{subfigure}
  \begin{subfigure}[t]{\linewidth}
    \centering
    % \vspace{2mm}
    \includegraphics[width=0.195\linewidth, trim=5mm 5mm 5mm 5mm, clip]{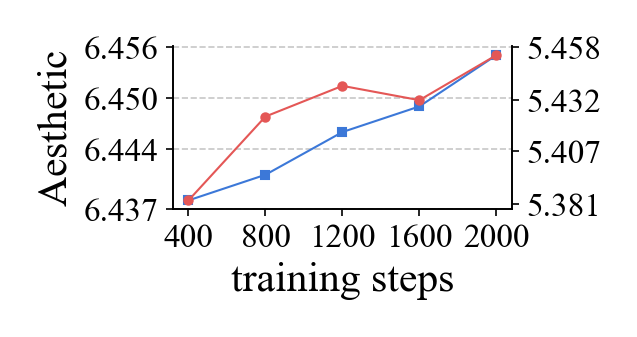}
    \includegraphics[width=0.195\linewidth, trim=5mm 5mm 5mm 5mm, clip]{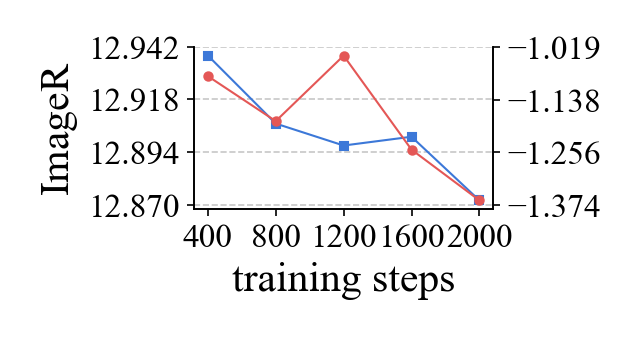}
    \includegraphics[width=0.195\linewidth, trim=5mm 5mm 5mm 5mm, clip]{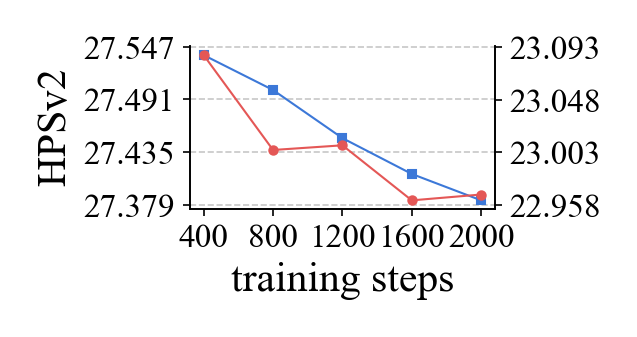}
    \includegraphics[width=0.195\linewidth, trim=5mm 5mm 5mm 5mm, clip]{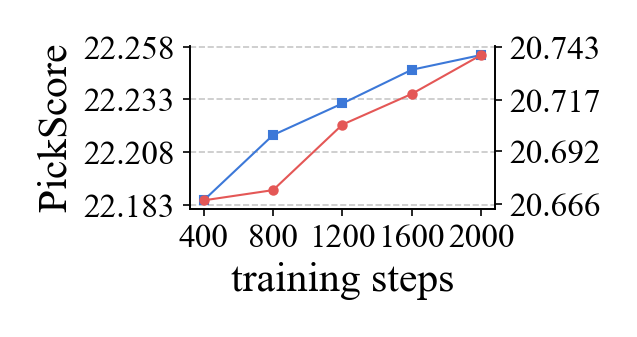}
    \includegraphics[width=0.195\linewidth, trim=5mm 5mm 5mm 5mm, clip]{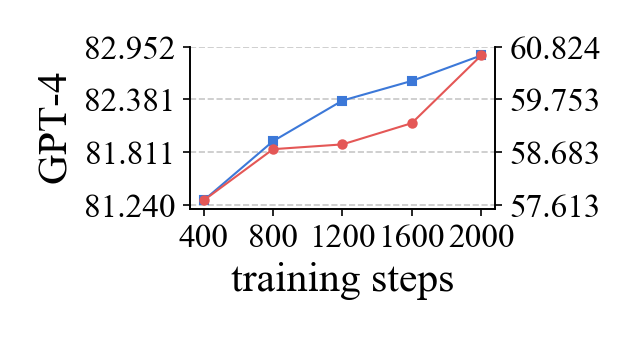}
    \includegraphics[width=0.195\linewidth, trim=5mm 5mm 5mm 5mm, clip]{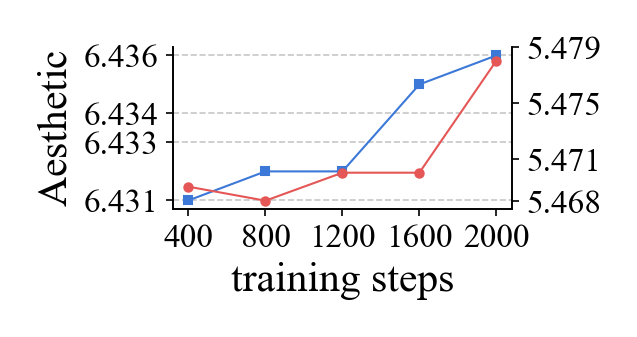}
    \includegraphics[width=0.195\linewidth, trim=5mm 5mm 5mm 5mm, clip]{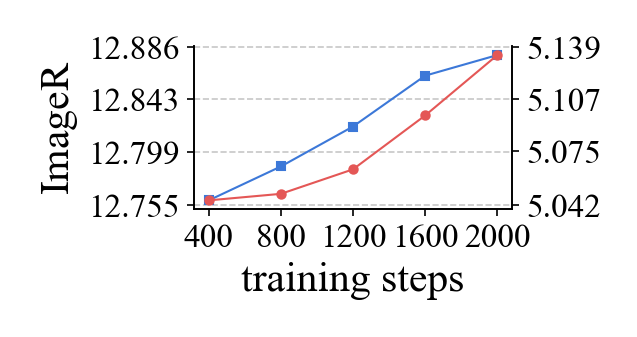}
    \includegraphics[width=0.195\linewidth, trim=5mm 5mm 5mm 5mm, clip]{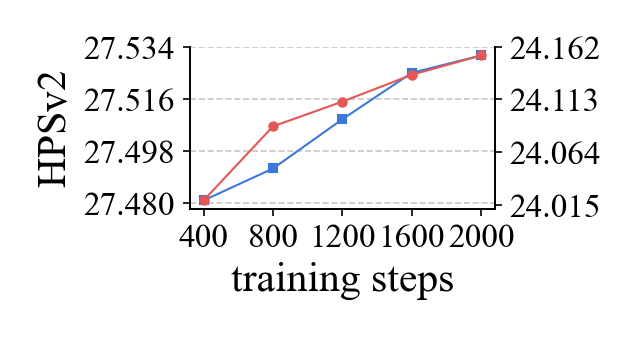}
    \includegraphics[width=0.195\linewidth, trim=5mm 5mm 5mm 5mm, clip]{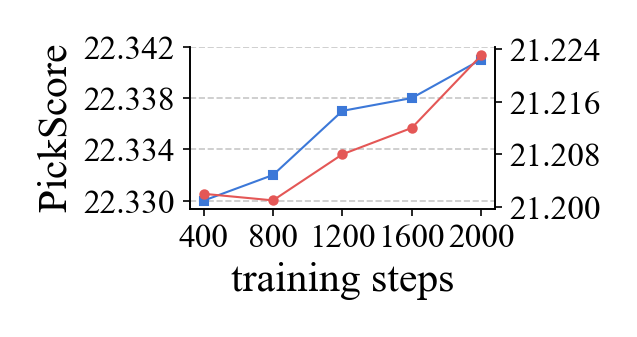}
    \includegraphics[width=0.195\linewidth, trim=5mm 5mm 5mm 5mm, clip]{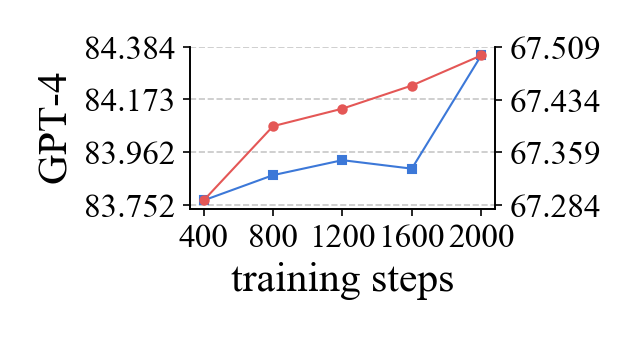}
    \vspace{-6mm}
    \caption{Sample scaling using PickScore.}
\end{subfigure}
\begin{subfigure}[t]{\linewidth}
    \centering
    % \vspace{2mm}
    \includegraphics[width=0.195\linewidth, trim=5mm 5mm 5mm 5mm, clip]{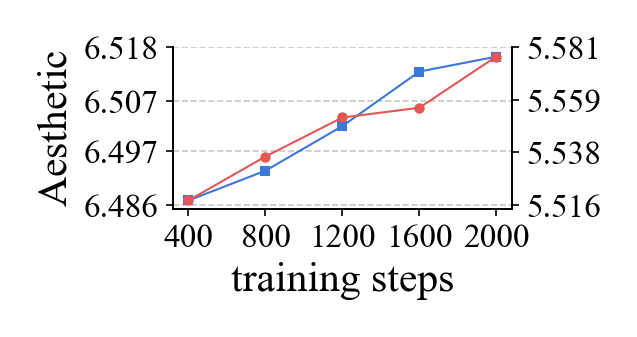}
    \includegraphics[width=0.195\linewidth, trim=5mm 5mm 5mm 5mm, clip]{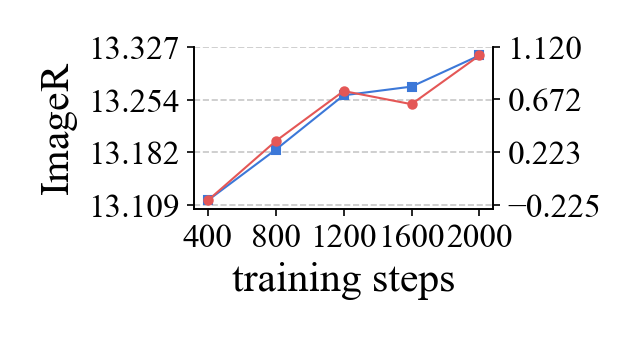}
    \includegraphics[width=0.195\linewidth, trim=5mm 5mm 5mm 5mm, clip]{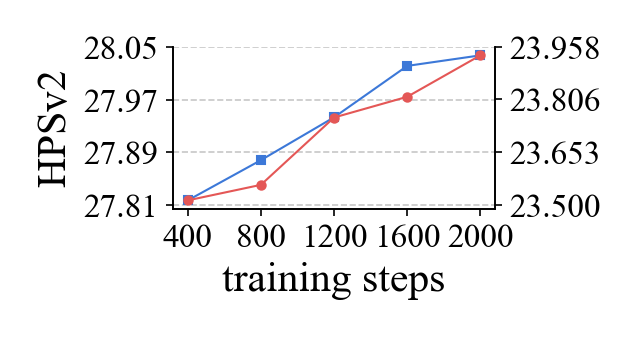}
    \includegraphics[width=0.195\linewidth, trim=5mm 5mm 5mm 5mm, clip]{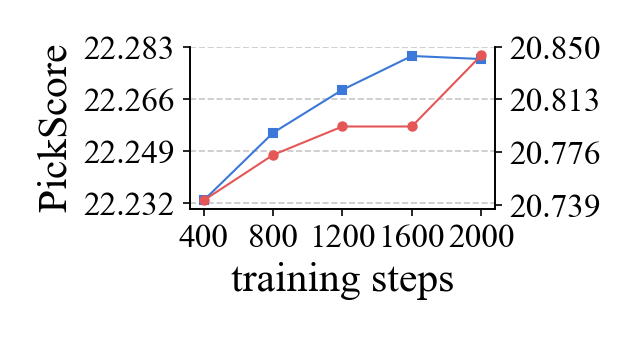}
    \includegraphics[width=0.195\linewidth, trim=5mm 5mm 5mm 5mm, clip]{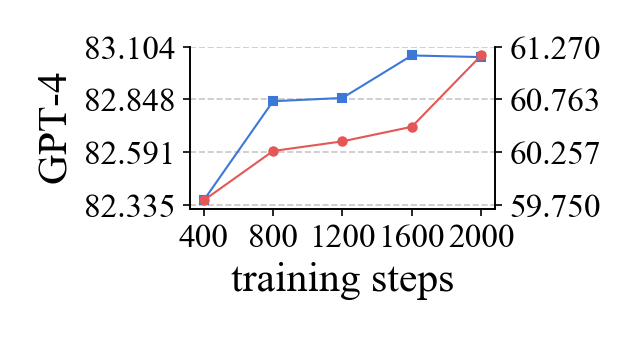}
    \includegraphics[width=0.195\linewidth, trim=5mm 5mm 5mm 5mm, clip]{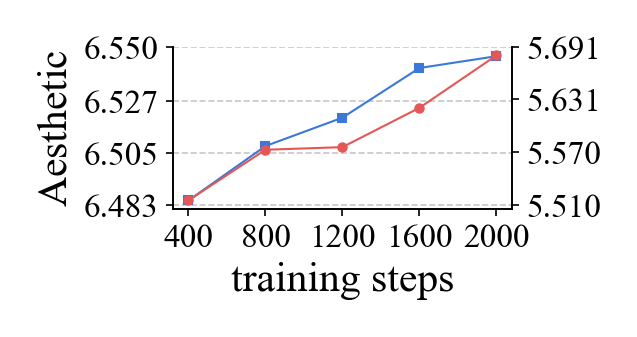}
    \includegraphics[width=0.195\linewidth, trim=5mm 5mm 5mm 5mm, clip]{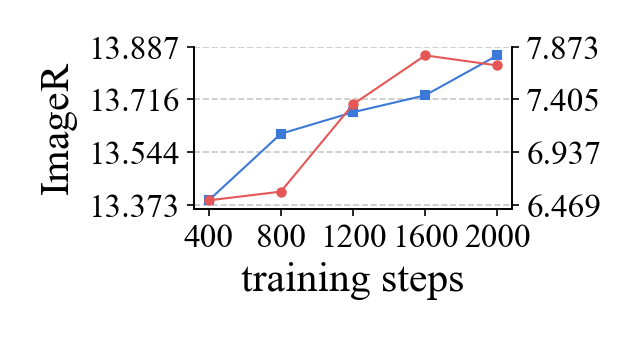}
    \includegraphics[width=0.195\linewidth, trim=5mm 5mm 5mm 5mm, clip]{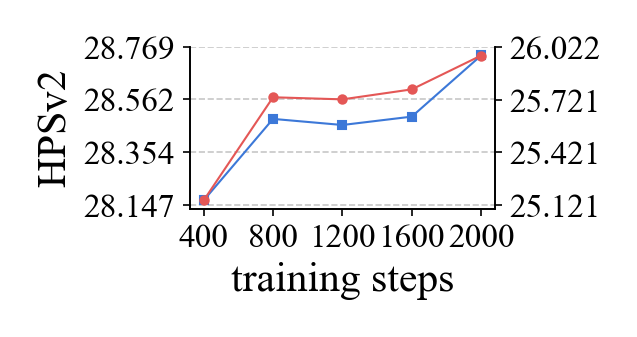}
    \includegraphics[width=0.195\linewidth, trim=5mm 5mm 5mm 5mm, clip]{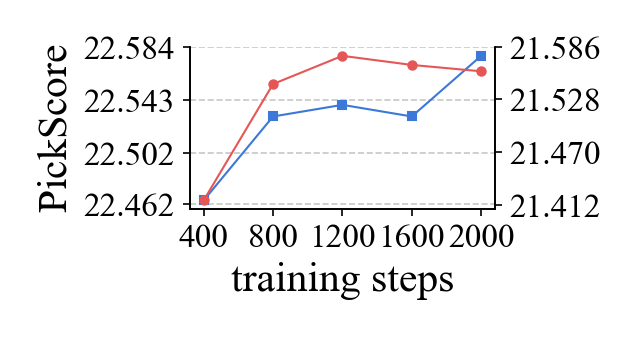}
    \includegraphics[width=0.195\linewidth, trim=5mm 5mm 5mm 5mm, clip]{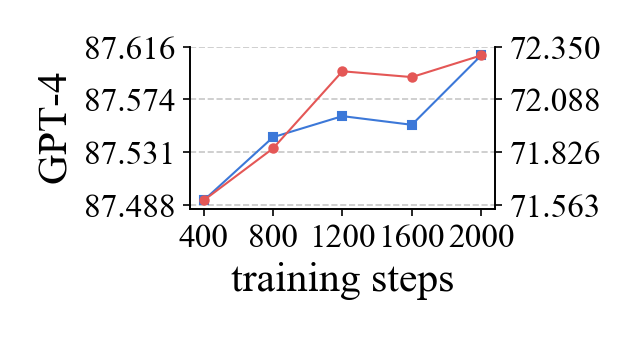}
    \vspace{-6mm}
    \caption{Sample scaling using Ensemble.}
\end{subfigure}
\vspace{-4mm}
\caption{\textbf{Candidate scaling} (a-c) and \textbf{sample scaling} (d-f) using HPSv2, PickScore, and Ensemble. We employ Aesthetic, ImageR, HPSv2, PickScore, and GPT-4 for evaluation. The first and second row of each sub-figure is based on BrushNet and FLUX.1 Fill, respectively. We use training steps to indicate the consumed samples to align the scaling across models (their batch-sizes are different).}
\label{fig:scaling}
\vspace{-4mm}
\end{figure*}

The ability of reward models to accurately predict human preferences is critical to the performance of preference alignment algorithms. To evaluate this capability, we apply DPO on the preference data constructed by the reward models and evaluate the model'performance after training. Specifically, based on the popular dataset of BrushData~\citep{BrushNet}, we generate 16 \textbf{candidate} inpainting results with varied random seeds for each prompt and the corresponding masked image. The candidates are scored by the reward models, and the highest-scoring (preferred) and lowest-scoring (dispreferred) samples form preference pairs for DPO training. Following ~\citep{Inference-Time-Scaling, unifiedreward}, the reward models are employed to serve two purposes: (1) \textit{providing scores to construct training data}, and (2) \textit{evaluating performance after training}. 
All experiments adhere to the same training configurations by default (e.g., learning rate is 1e-7, $\beta$ is 2000, 2000 training steps, etc.), with the only variation being the reward model used to score and construct the training data. 
Note that we may encounter an \textbf{oracle reward model}~\citep{Inference-Time-Scaling}, where the same model is used both for data construction and performance evaluation within one experiment. We assess results on two benchmarks, i.e., BrushBench~\citep{BrushNet} and EditBench~\citep{EditBench}.

Here, we introduce a new reward model---\textbf{Ensemble} that constructs samples based on the average ranking of all the reward models. We make \textbf{GPT-4}~\citep{GPT-4} serve as a ``fair'' evaluator by assessing aesthetic quality, structural coherence, and semantic alignment of the results (see details in Appendix). We report two other results. \textbf{Baseline:} The model's performance prior to DPO training. \textbf{Random:} It involves training with randomly sampled preferred and dispreferred pairs. Results are reported in Table~\ref{tab:brushnet_metrics_8cols_twobench} and Table~\ref{tab:flux_metrics_8col_twobench}. We have the following observations and conclusions.

\textbf{Some reward models are not reliable evaluators.}
It is believed that an accurate and robust reward model should assign high evaluation scores to models trained on its own preference dataset (i.e., the oracle reward model setting). Surprisingly, we find that CLIPScore, VQAScore, and Perception fail to meet this requirement---in~\autoref{tab:flux_metrics_8col_twobench}, their scores can be even lower than baseline or random results. 
% Worse still, VQAScore assigns nearly identical scores to most of the results on BrushBench, indicating a severe lack of discriminative capability. 
We hypothesize that the fail of CLIPScore and Perception stems from their large-scale yet potentially coarse contrastive pre-training; the fail of VQAScore likely arises from its simplistic, VQA-like evaluation approach. In light of this, \textit{we exclude these models from subsequent analyses}.

\textbf{Most reward models provide valid reward scores.} 
Most reward models are capable of offering valid reward scores for preference data construction, as they outperform both the baseline and random selection across most evaluation results---especially GPT-4. Even though CLIPScore and Perception are observed to be less effective at accurately evaluating on small-scale benchmarks, they remain viable when their reward scores are incorporated into larger-scale preference training datasets. In this context, we continue to attribute VQAScore’s limitations to its simple scoring methodology.

\textbf{Reward models may share common biases.}
We find that the model trained on HPSv2-constructed data outperforms most competitors when evaluated using public reward models. Specifically, when trained using BrushNet, it ranks first or second in 4 out of 12 evaluations; when trained using FLUX.1 Fill, it ranks first or second in 9 out of 12 evaluations. This pattern aligns with GPT-4’s results when using FLUX.1 Fill but diverges when using BrushNet---under the latter condition, the model is largely outperformed by PickScore. We posit that HPSv2 and many other models may share some common biases, which can potentially lead to reward hacking~\citep{hack}. 

\textbf{Ensemble is an accurate and robust reward model.}
It shows that Ensemble ranks first or second in 11 out of 12 public model evaluations when using BrushNet, and 7 out of 12 when using FLUX.1 Fill. Besides, Ensemble ranks first or second in 3 out of 4 GPT-4's evaluations across both baseline models, demonstrating its robustness in constructing effective preference data. We hypothesize that its versatility arises from the bias of reward models being weakened in Ensemble.

\textbf{Discussion.}
Part of the above analysis is based on an untested assumption---GPT-4 is an ideal evaluator. We will examine its validity as well as reward hacking in~\autoref{hacking} and the Appendix.

\section{How Scalable are Preference Data?}
\label{sec:scalable}

\begin{figure*}[t]
%brushnet
\begin{subfigure}[t]{\linewidth}
    \centering
    \includegraphics[width=0.11\linewidth, trim=0mm 0mm 0mm 0mm, clip]{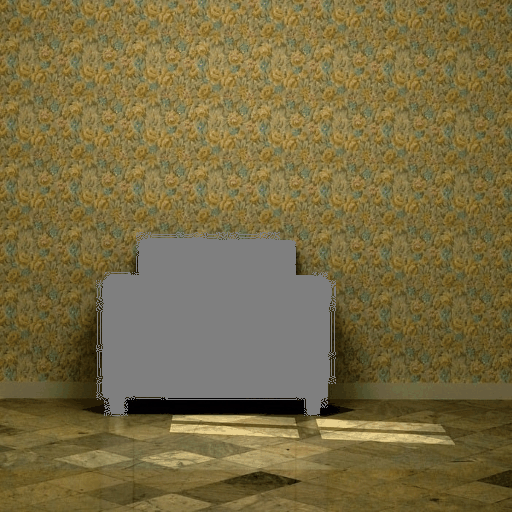}
    \includegraphics[width=0.11\linewidth, trim=0mm 0mm 0mm 0mm, clip]{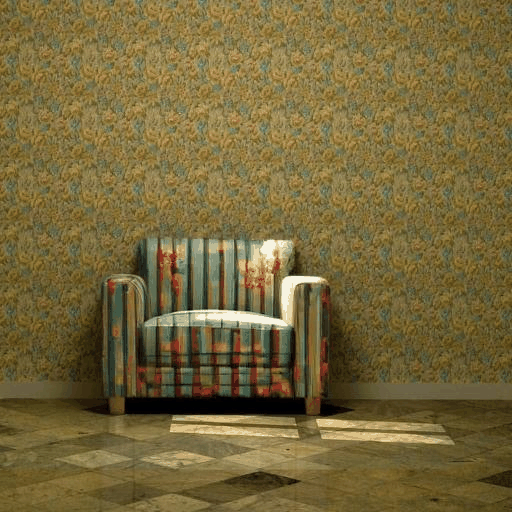}
    \includegraphics[width=0.11\linewidth, trim=0mm 0mm 0mm 0mm, clip]{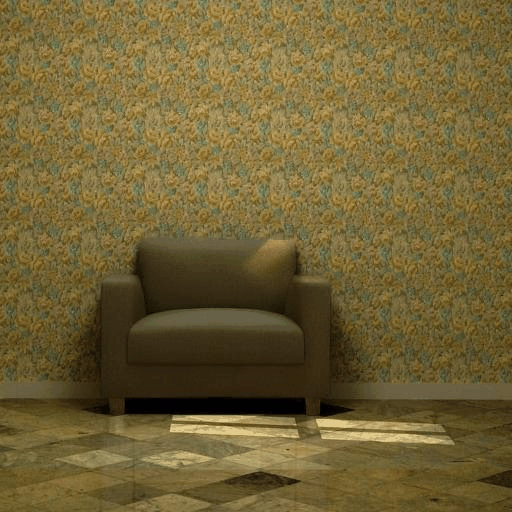}
    \includegraphics[width=0.11\linewidth, trim=0mm 0mm 0mm 0mm, clip]{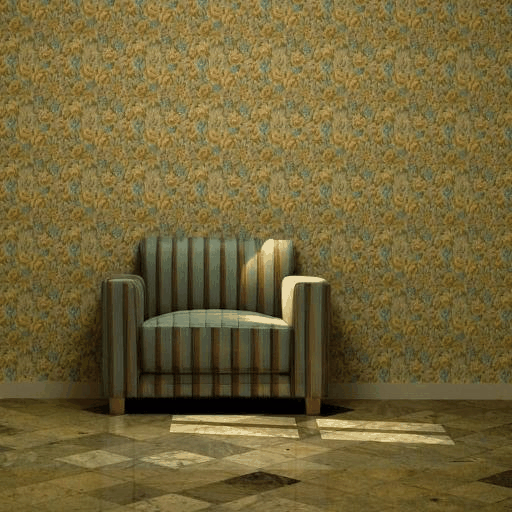}
    \includegraphics[width=0.11\linewidth, trim=0mm 0mm 0mm 0mm, clip]{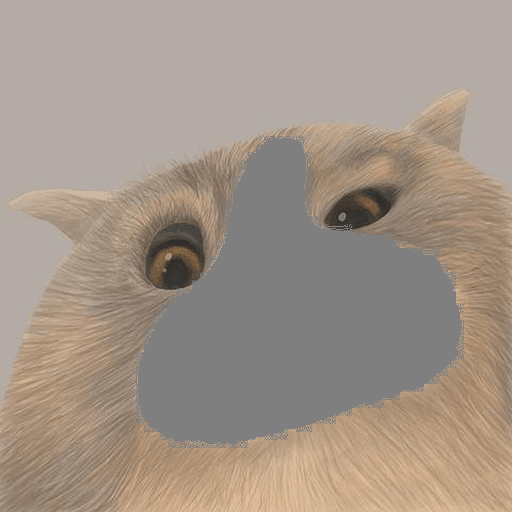}
    \includegraphics[width=0.11\linewidth, trim=0mm 0mm 0mm 0mm, clip]{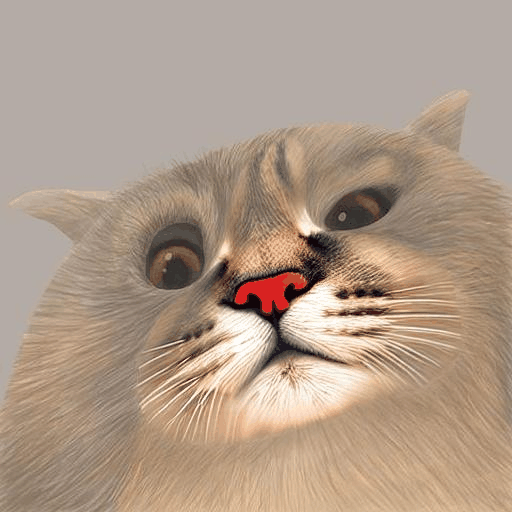}
    \includegraphics[width=0.11\linewidth, trim=0mm 0mm 0mm 0mm, clip]{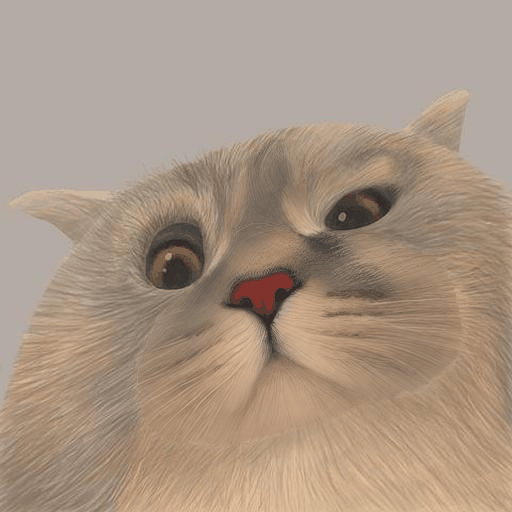}
    \includegraphics[width=0.11\linewidth, trim=0mm 0mm 0mm 0mm, clip]{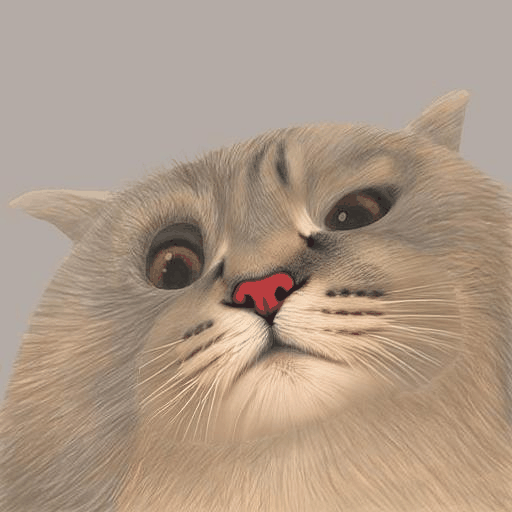}
    \includegraphics[width=0.11\linewidth, trim=0mm 0mm 0mm 0mm, clip]{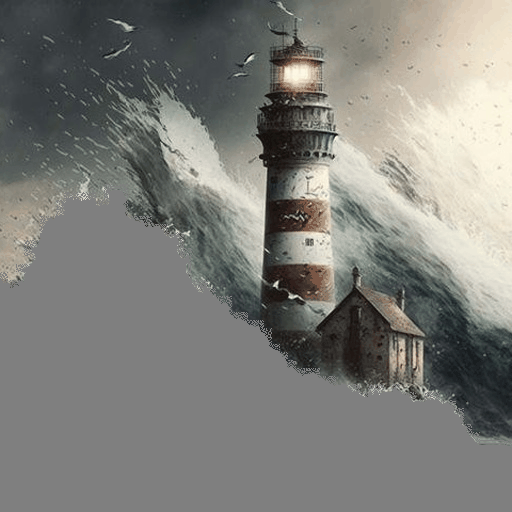}
    \includegraphics[width=0.11\linewidth, trim=0mm 0mm 0mm 0mm, clip]{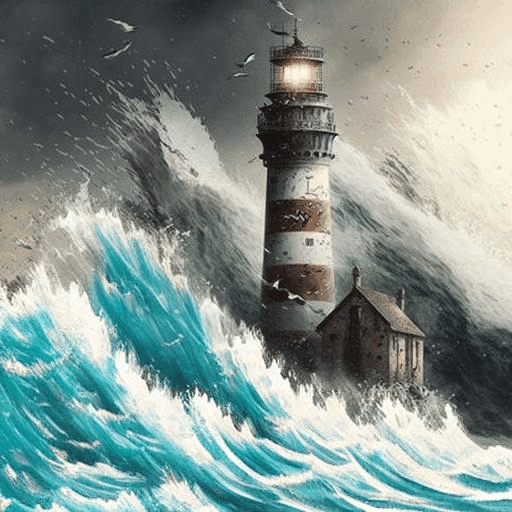}
    \includegraphics[width=0.11\linewidth, trim=0mm 0mm 0mm 0mm, clip]{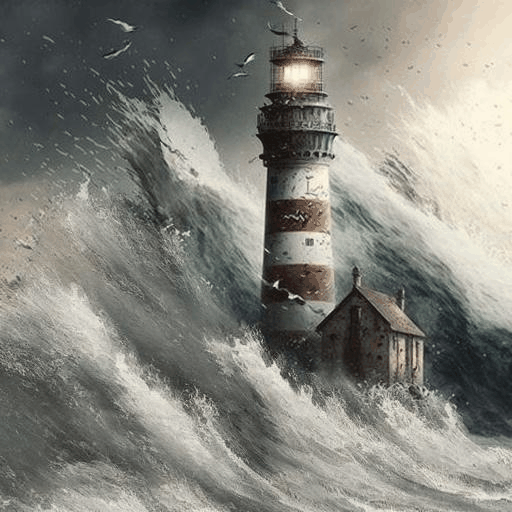}
    \includegraphics[width=0.11\linewidth, trim=0mm 0mm 0mm 0mm, clip]{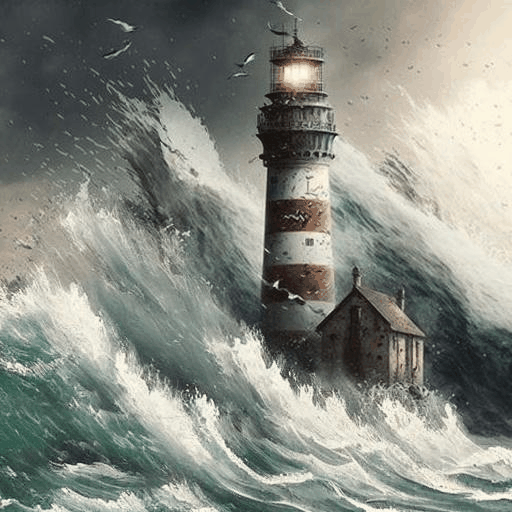}
    \includegraphics[width=0.11\linewidth, trim=0mm 0mm 0mm 0mm, clip]{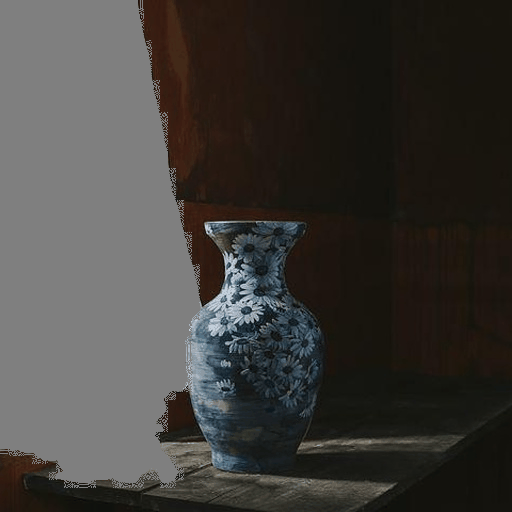}
    \includegraphics[width=0.11\linewidth, trim=0mm 0mm 0mm 0mm, clip]{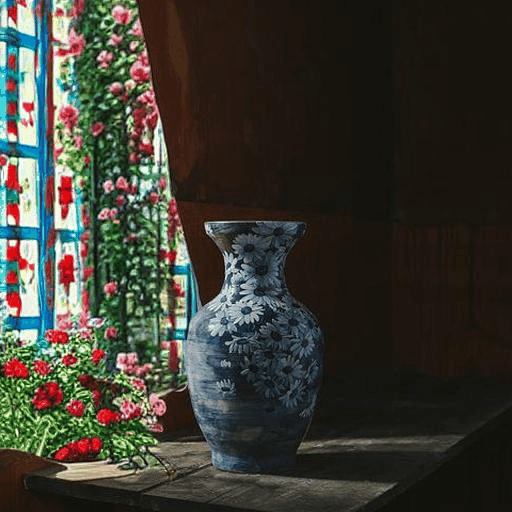}
    \includegraphics[width=0.11\linewidth, trim=0mm 0mm 0mm 0mm, clip]{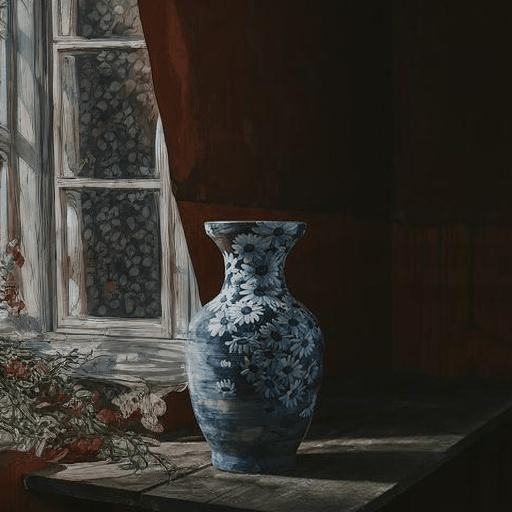}
    \includegraphics[width=0.11\linewidth, trim=0mm 0mm 0mm 0mm, clip]{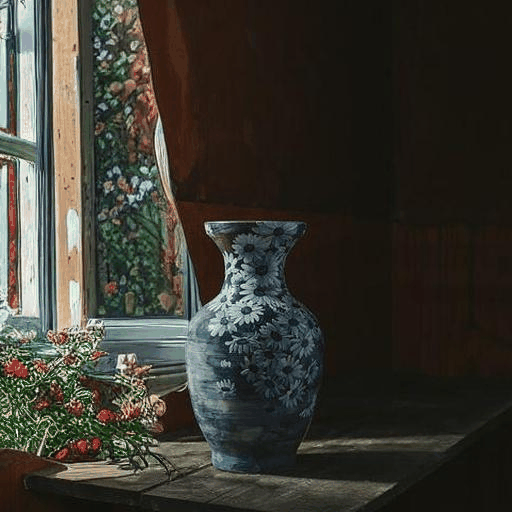}
    \caption{Examples from models trained using \textbf{BrushNet}.}
\end{subfigure}
% flux
\begin{subfigure}[t]{\linewidth}
    \centering
    \includegraphics[width=0.11\linewidth, trim=0mm 0mm 0mm 0mm, clip]{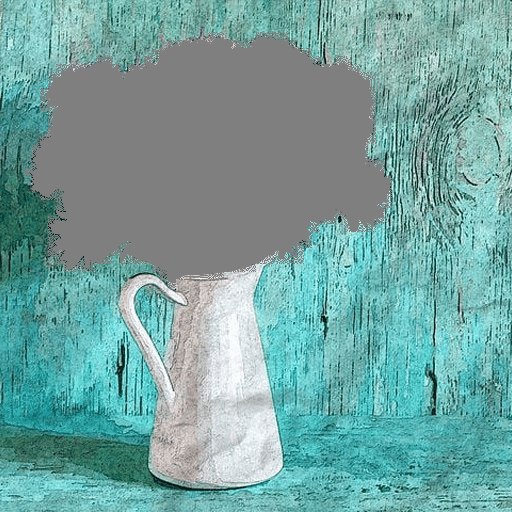}
    \includegraphics[width=0.11\linewidth, trim=0mm 0mm 0mm 0mm, clip]{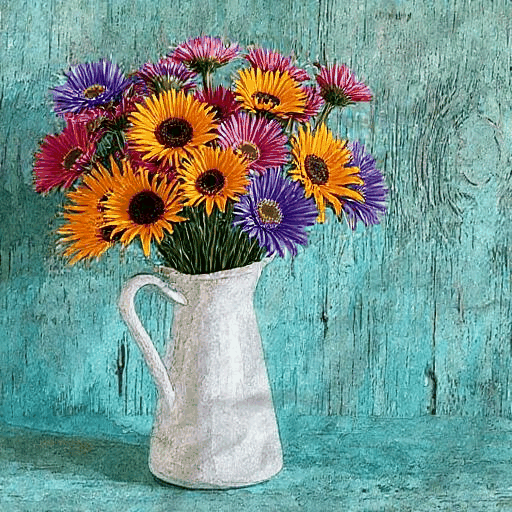}
    \includegraphics[width=0.11\linewidth, trim=0mm 0mm 0mm 0mm, clip]{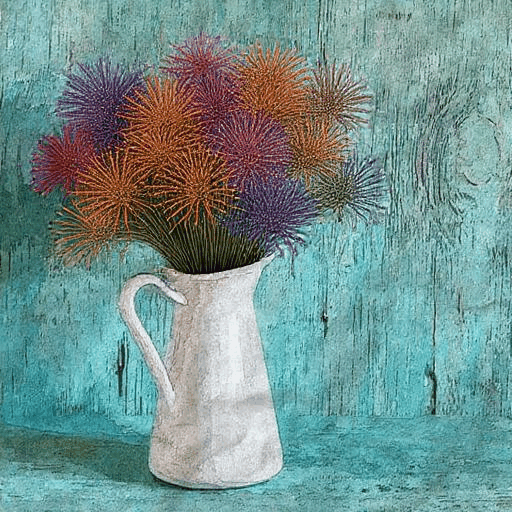}
    \includegraphics[width=0.11\linewidth, trim=0mm 0mm 0mm 0mm, clip]{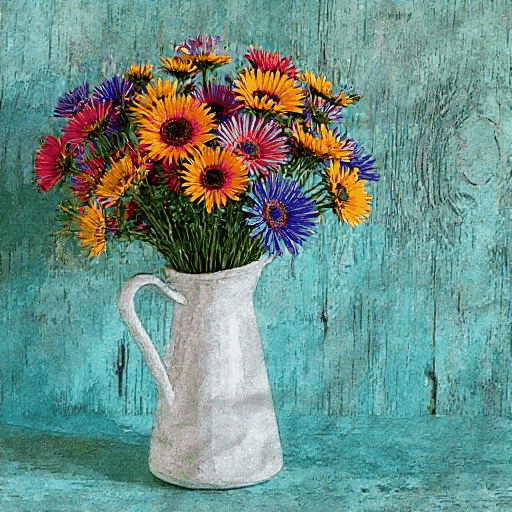}
    \includegraphics[width=0.11\linewidth, trim=0mm 0mm 0mm 0mm, clip]{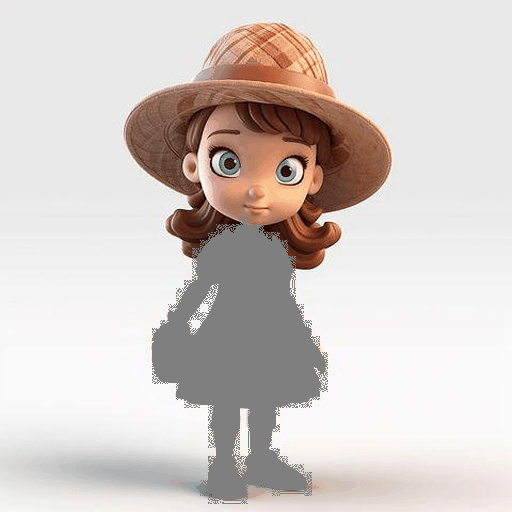}
    \includegraphics[width=0.11\linewidth, trim=0mm 0mm 0mm 0mm, clip]{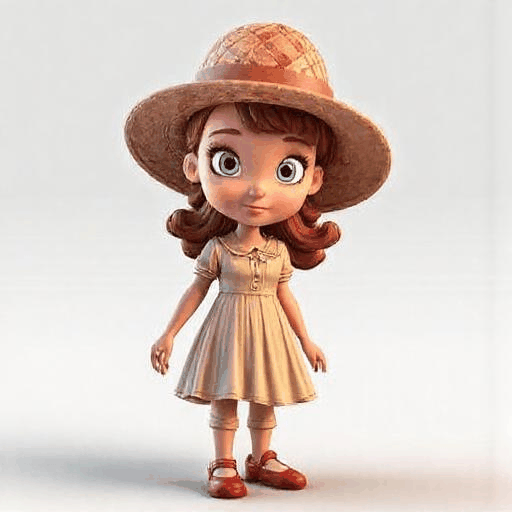}
    \includegraphics[width=0.11\linewidth, trim=0mm 0mm 0mm 0mm, clip]{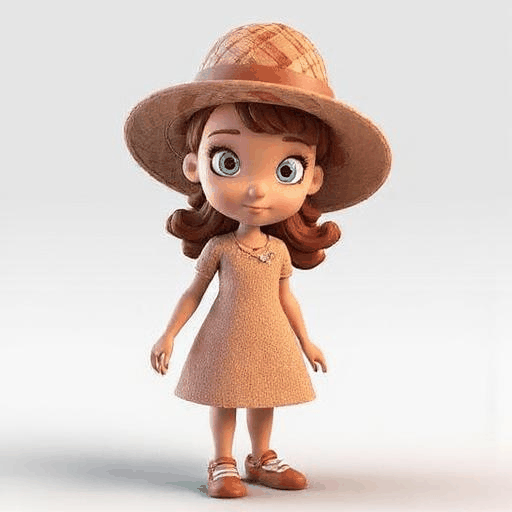}
    \includegraphics[width=0.11\linewidth, trim=0mm 0mm 0mm 0mm, clip]{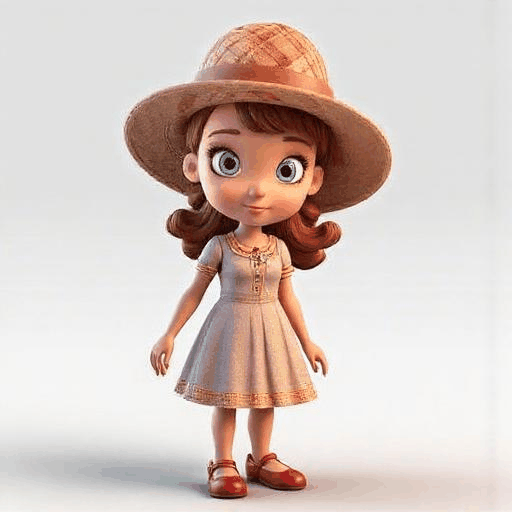}
    \includegraphics[width=0.11\linewidth, trim=0mm 0mm 0mm 0mm, clip]{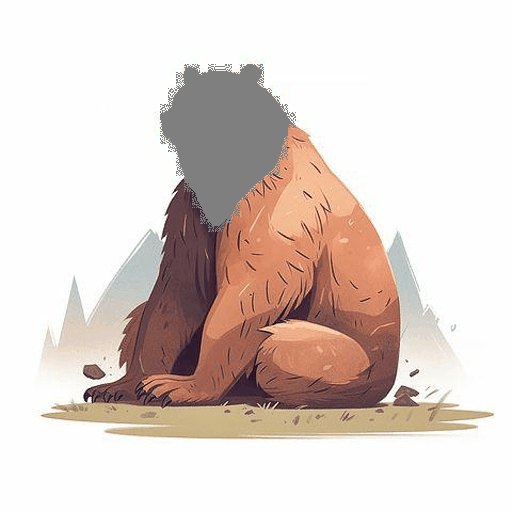}
    \includegraphics[width=0.11\linewidth, trim=0mm 0mm 0mm 0mm, clip]{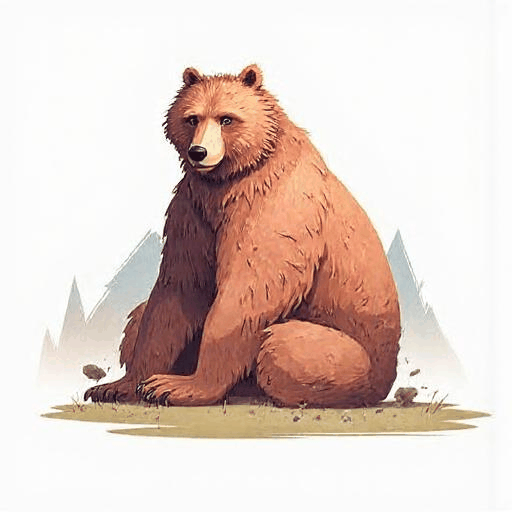}
    \includegraphics[width=0.11\linewidth, trim=0mm 0mm 0mm 0mm, clip]{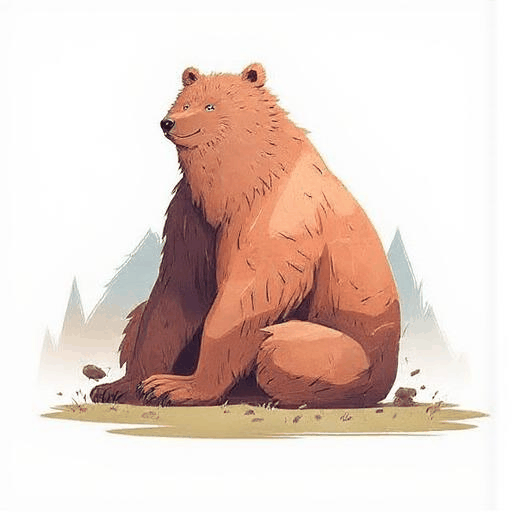}
    \includegraphics[width=0.11\linewidth, trim=0mm 0mm 0mm 0mm, clip]{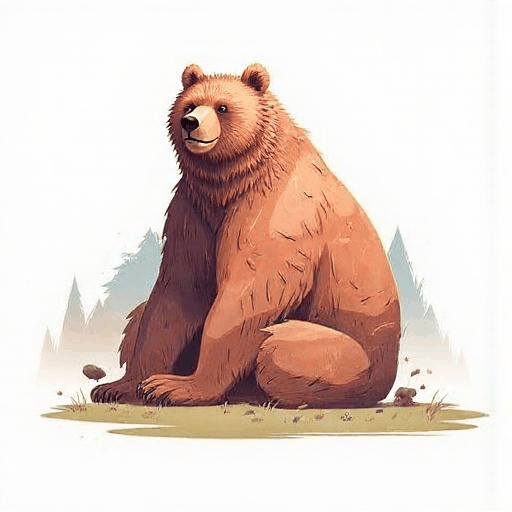}
    \includegraphics[width=0.11\linewidth, trim=0mm 0mm 0mm 0mm, clip]{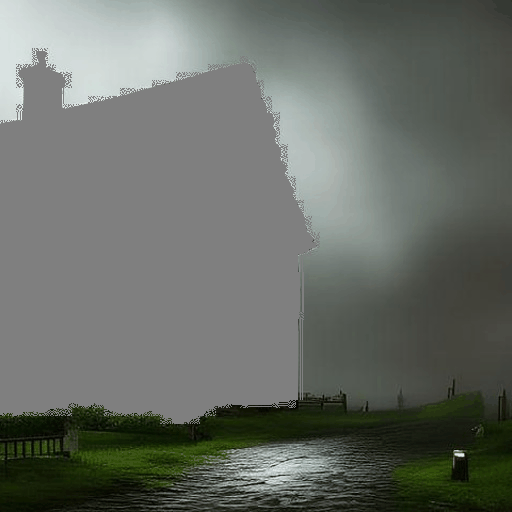}
    \includegraphics[width=0.11\linewidth, trim=0mm 0mm 0mm 0mm, clip]{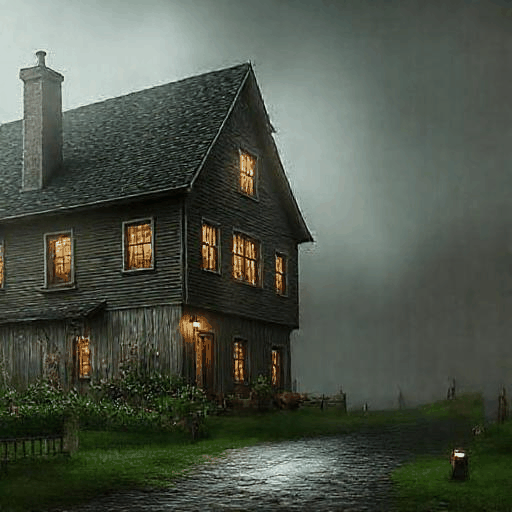}
    \includegraphics[width=0.11\linewidth, trim=0mm 0mm 0mm 0mm, clip]{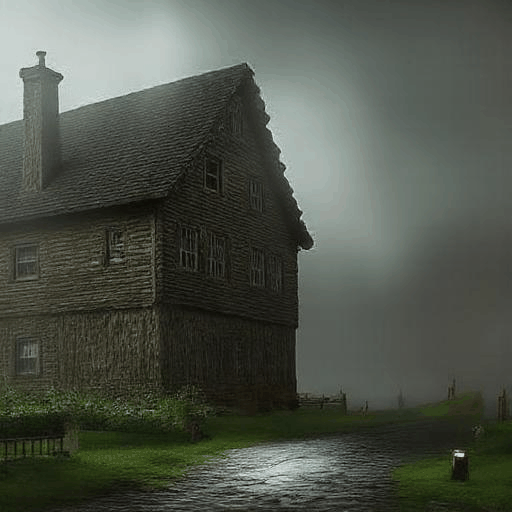}
    \includegraphics[width=0.11\linewidth, trim=0mm 0mm 0mm 0mm, clip]{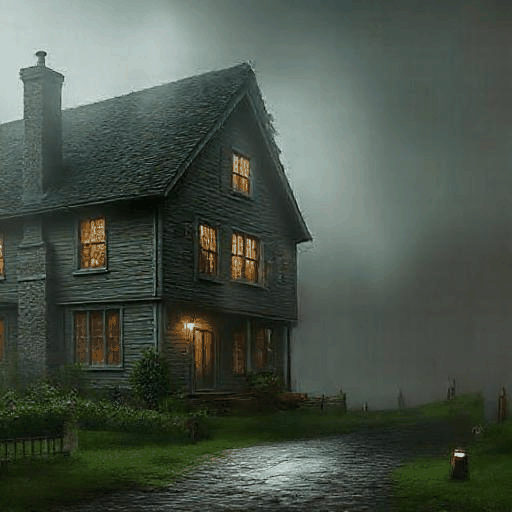}
    \caption{Examples from models trained using \textbf{FLUX.1 Fill}.}
    \vspace{-3mm}
\end{subfigure}
\caption{\textbf{Bias studies.} In each sub-figure, the four images (from left to right) display: the \textit{masked image}, followed by inpainting results from models trained using \textit{HPSv2}, \textit{PickScore}, and \textit{Ensemble}. We omit text prompts for brevity. Zoom in to see details. Find more examples in the Appendix.}
\label{fig:bias}
\vspace{-4mm}
\end{figure*}

\begin{figure*}[t]
\begin{subfigure}[t]{\linewidth}
\centering
	\includegraphics[width=0.1\linewidth, trim=0mm 0mm 0mm 0mm, clip]{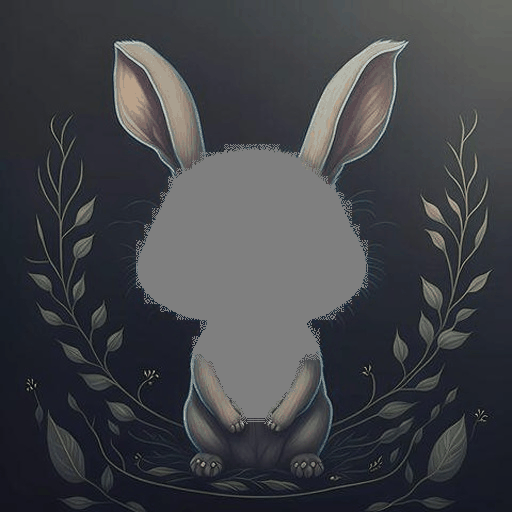}
	\includegraphics[width=0.1\linewidth, trim=0mm 0mm 0mm 0mm, clip]{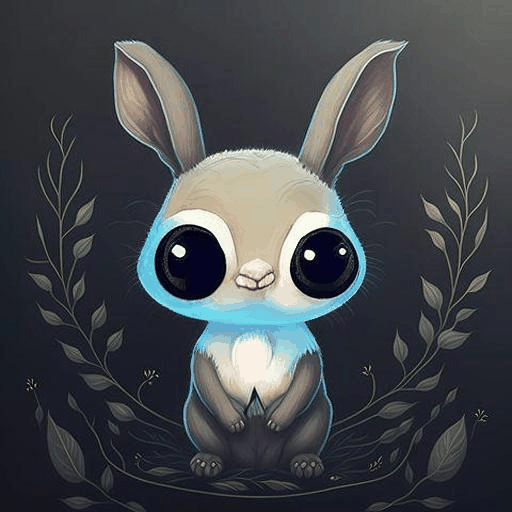}
	\includegraphics[width=0.1\linewidth, trim=0mm 0mm 0mm 0mm, clip]{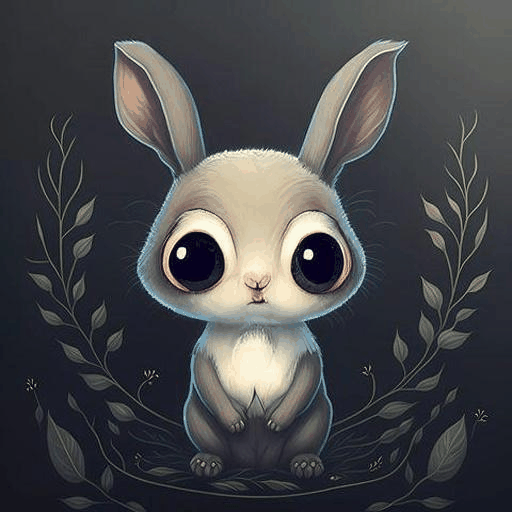}
	\includegraphics[width=0.1\linewidth, trim=0mm 0mm 0mm 0mm, clip]{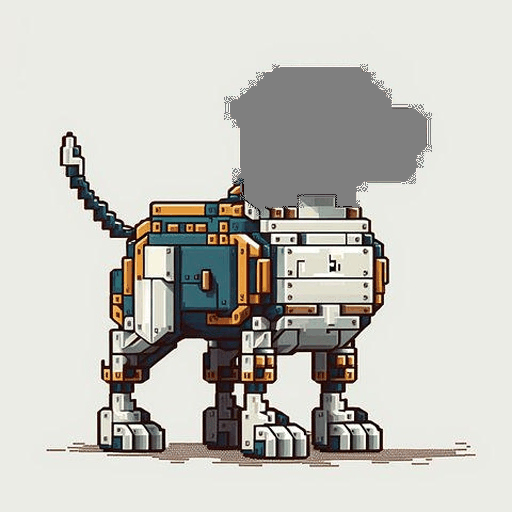}
	\includegraphics[width=0.1\linewidth, trim=0mm 0mm 0mm 0mm, clip]{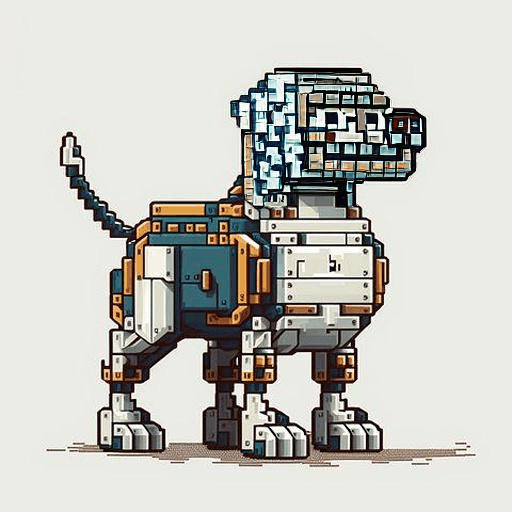}
	\includegraphics[width=0.1\linewidth, trim=0mm 0mm 0mm 0mm, clip]{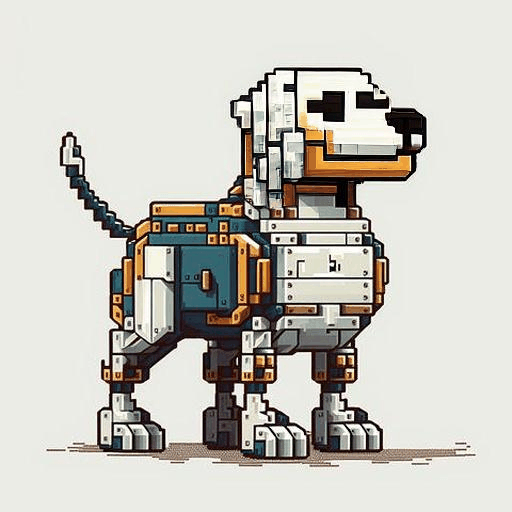}
	\includegraphics[width=0.1\linewidth, trim=0mm 0mm 0mm 0mm, clip]{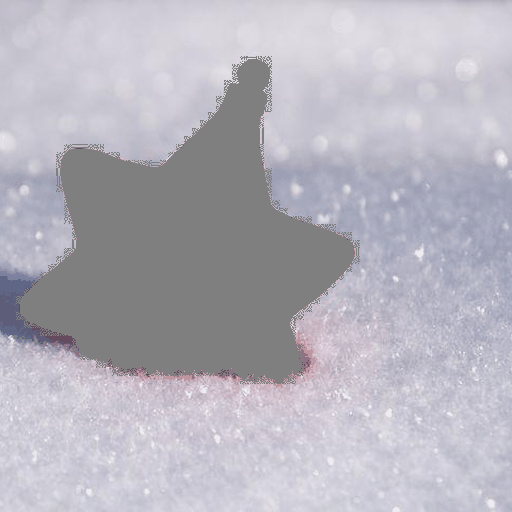}
	\includegraphics[width=0.1\linewidth, trim=0mm 0mm 0mm 0mm, clip]{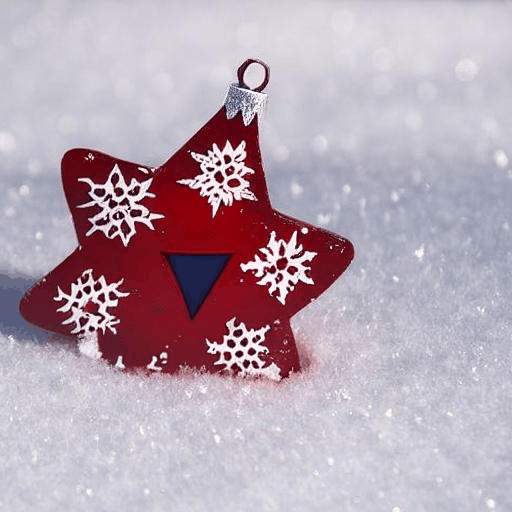}
	\includegraphics[width=0.1\linewidth, trim=0mm 0mm 0mm 0mm, clip]{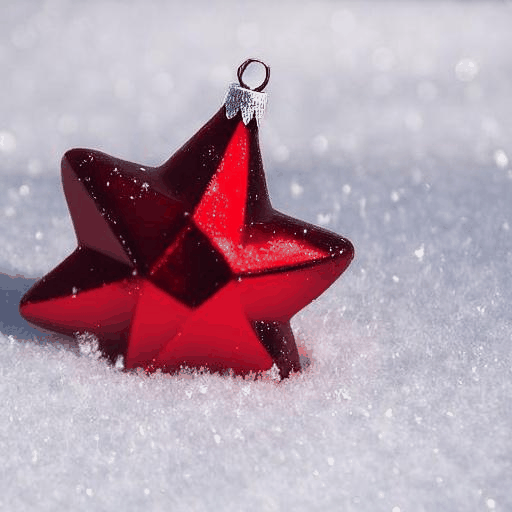}
\caption{Examples from models trained using \textbf{BrushNet}.}
\end{subfigure}

\begin{subfigure}[t]{\linewidth}
\centering
	\includegraphics[width=0.1\linewidth, trim=0mm 0mm 0mm 0mm, clip]{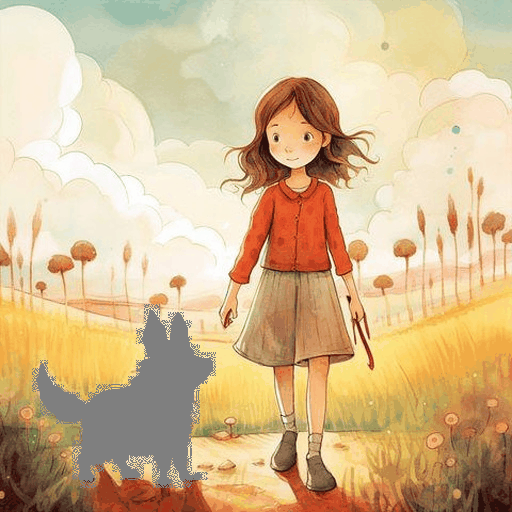}
	\includegraphics[width=0.1\linewidth, trim=0mm 0mm 0mm 0mm, clip]{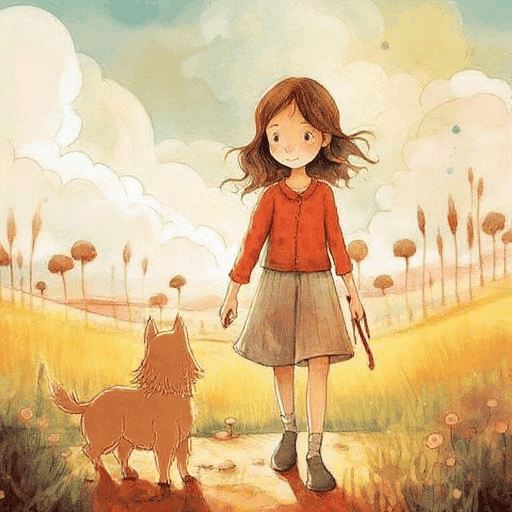}
	\includegraphics[width=0.1\linewidth, trim=0mm 0mm 0mm 0mm, clip]{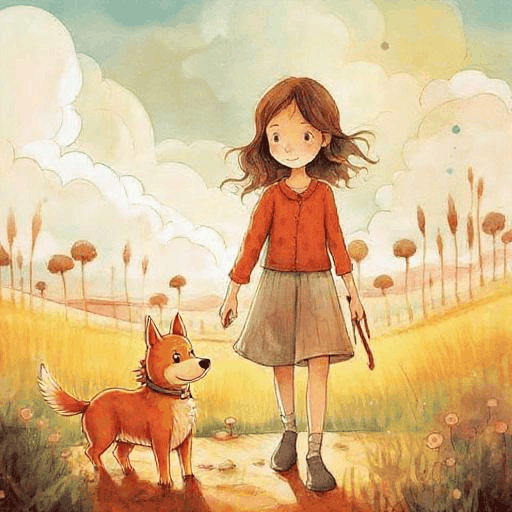}
	\includegraphics[width=0.1\linewidth, trim=0mm 0mm 0mm 0mm, clip]{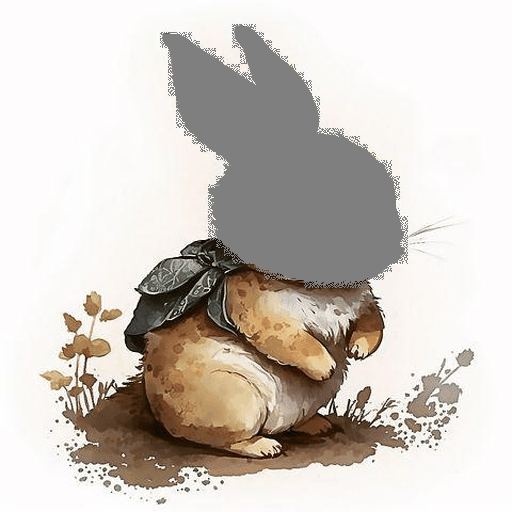}
	\includegraphics[width=0.1\linewidth, trim=0mm 0mm 0mm 0mm, clip]{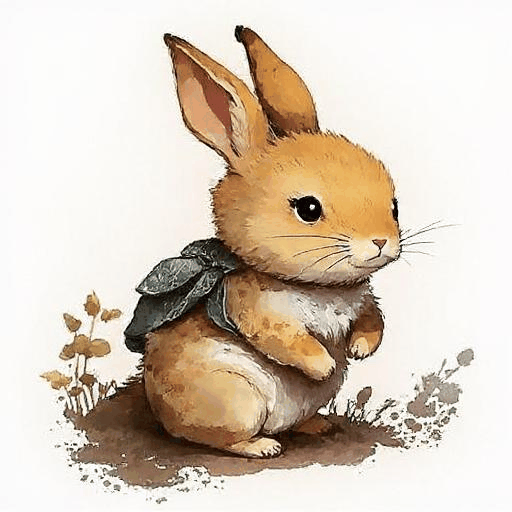}
	\includegraphics[width=0.1\linewidth, trim=0mm 0mm 0mm 0mm, clip]{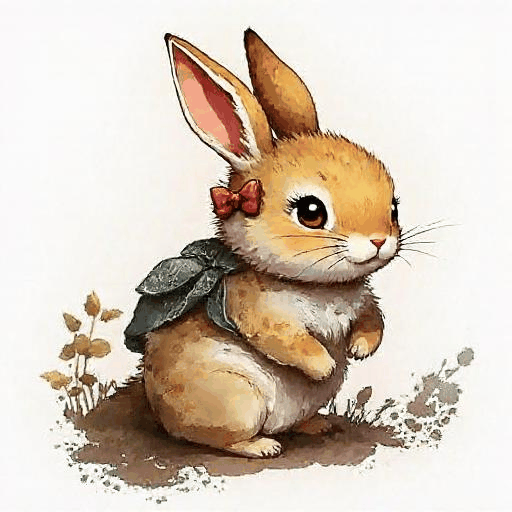}
	\includegraphics[width=0.1\linewidth, trim=0mm 0mm 0mm 0mm, clip]{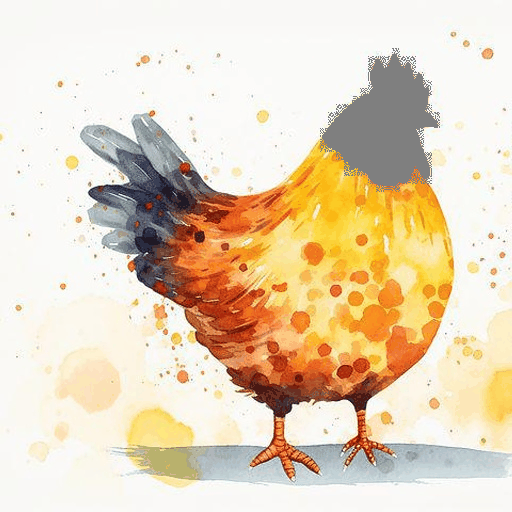}
	\includegraphics[width=0.1\linewidth, trim=0mm 0mm 0mm 0mm, clip]{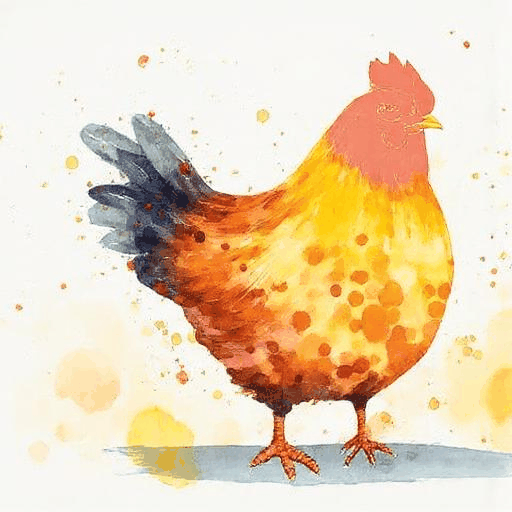}
	\includegraphics[width=0.1\linewidth, trim=0mm 0mm 0mm 0mm, clip]{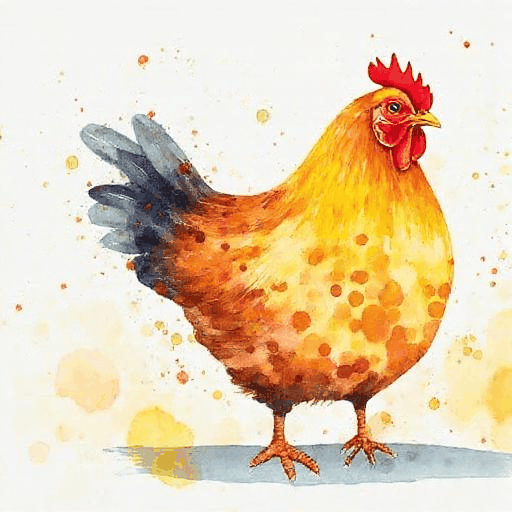}
\caption{Examples from models trained using \textbf{FLUX.1 Fill}.}
\vspace{-2mm}
\end{subfigure}
\caption{\textbf{Qualitative results of ablations.} In each sub-figure, the three images (from left to right) display: the \textit{masked image}, followed by inpainting results from \textit{baseline models} and \textit{baseline models + preference alignment using Ensemble}. We omit text prompts for brevity. Zoom in to seed details.}
\vspace{-8mm}
\label{fig:baseline}
\end{figure*}
\begin{table*}[t]
\centering
\caption{Ablation studies on a new dataset of \textit{I Dream My Painting} \citep{idream}.}
\vspace{-4mm}
\resizebox{1.\linewidth}{!}{
\begin{threeparttable}  
\renewcommand{\arraystretch}{1.2}
\setlength{\tabcolsep}{5pt}
{
     \begin{tabular}{l l C{1.8cm}C{1.8cm}C{1.8cm}C{1.8cm}C{1.8cm}C{1.8cm}C{1.8cm}C{1.8cm}C{1.8cm}C{1.8cm}}
          \toprule
          {inpainting model}  
          & {{CLIPScore}}   & {{Aesthetic}}  & {{ImageR}} & {{PickScore}}  & {{HPSv2}}   & {{VQAScore}} & {{UnifiedR}} & {{Perception}} & {{HPSv3}} & {\cellcolor{sem!15}{GPT-4}}     \\ 
                                     
          \midrule
                {BrushNet} &            24.849 &            5.923 &           -0.246 &            20.550 &            19.749 &            8.503 &            2.317 &            27.317 &           -0.551 &     \cellcolor{sem!15}72.669 \\
  \textbf{BruPA (ours) } &            \textbf{25.460} &            \textbf{6.111} &            \textbf{2.152} &            \textbf{20.735} &            \textbf{21.086} &            \textbf{8.653} &            \textbf{2.463} &            \textbf{28.294} &            \textbf{1.265} &     \cellcolor{sem!15}\textbf{73.739} \\
          \hline
               FLUX.1 Fill &            24.194 &            6.017 &            0.544 &            20.855 &            20.203 &            8.667 &            2.476 &            26.627 &            0.547 &   \cellcolor{sem!15}{76.391} \\
    \textbf{FluPA (ours)} &   \textbf{25.500} &   \textbf{6.448} &   \textbf{5.961} &   \textbf{21.407} &   \textbf{23.770} &   \textbf{9.031} &   \textbf{2.784} &   \textbf{28.868} &   \textbf{5.023} &   \cellcolor{sem!15}\textbf{79.255} \\
          \bottomrule
          
     \end{tabular}
     \begin{tablenotes}
     \item \textbf{Bold} values denote the best results. 
     \end{tablenotes}
}
\end{threeparttable}
}
\label{tab:comparison_on_I_DREAM_MY_PAINT}
\vspace{-2mm}
\end{table*}

The results in \autoref{sec:effective} have shown that \textit{HPSv2}, \textit{PickScore}, and \textit{Ensemble} are promising reward models. Building on this finding, we conduct an investigation into the scalability of preference data using these reward models. 
Specifically, we explore along two dimensions: (1) \textbf{Candidate scaling.} As the number of candidate samples generated from different random seeds increases, their diversity expands. This augmentation in diversity would enhance the accuracy of the construction of preferred and dispreferred samples by enhancing their differences. (2) \textbf{Sample scaling.} A larger dataset enables the model to capture nuanced patterns more comprehensively, leading to deeper learning of preferences. 
% Since the training batch-sizes for BrushNet and FLUX.1 Fill are 64 and 8 respectively, \textit{we use training steps to indicate the consumed samples} to make the scaling experiment consistent. 
Based on the insights in \autoref{sec:effective}, we select Aesthetic, ImageR, HPSv2, PickScore, and GPT-4 as the evaluation models. To enable the model to achieve optimal performance, we conduct a search over two typical hyper-parameters---$\beta$ and learning rate (see details in the Appendix), before the scaling experiments. For each experiment, we tune one scaling dimension and fix the other dimension. The results are reported in \autoref{fig:scaling}. We have the following results and discoveries.

\textbf{Consistent scaling trends across models and benchmarks.} First, we observe that data scaling demonstrates robust trends regardless of the model used---BrushNet and FLUX.1 Fill, in the first and second rows of each sub-figure, respectively. Second, we find that similar scaling trends emerge when evaluating on different benchmarks, as evidenced by the comparable patterns of the two lines within each sub-figure. These findings indicate that the observed scaling behavior is robust and generalizable. However, we also observe some inconsistent phenomena: when using PickScore as the reward model, ImageR/HPSv2 exhibit opposite trends on BrushNet and FLUX.1 Fill. This issue is caused by the characteristics of both reward models and baseline models, as  analyzed in~\autoref{hacking}.

\textbf{Reward hacking from HPSv2 undermines training.}
When evaluated by Aesthetic, ImageR, HPSv2, and PickScore, using HPSv2 as the reward model shows benefits from both candidate scaling and sample scaling. However, its GPT-4 results deteriorate significantly in the later stages of scaling. This observation aligns with our finding in \autoref{sec:effective}, where HPSv2 achieves good results under public model evaluations but sometimes loses to others when assessed by GPT-4. We hypothesize that this degradation stems from some shared common biases among these reward models.

\textbf{Ensemble offers robust data scaling by resisting hacking.}
Although PickScore demonstrates good scaling behavior, its performance remains sub-optimal. In contrast, the Ensemble approach achieves the best results across benchmarks, model structures, evaluation models, and scaling dimensions. This is likely because Ensemble averages the preference choices of different reward models, which eliminates the biases of the employed reward models and improves its resistance to the hacking.
\section{How Reward Hacking Happens?}
\label{hacking}
\begin{table*}[t]
\centering
\caption{Comparisons of state-of-the-art image inpainting models on BrushBench and EditBench.}
\vspace{-4mm}
\resizebox{1.\linewidth}{!}{
\begin{threeparttable}  
\renewcommand{\arraystretch}{1.5}
\setlength{\tabcolsep}{0pt}
{
     \begin{tabular}{l l C{1.2cm}C{1.2cm}C{1.2cm}C{1.2cm}C{1.2cm}C{1.2cm}C{1.2cm}C{1.2cm}C{1.2cm}C{1.2cm}C{1.2cm}C{1.2cm}C{1.2cm}C{1.2cm}C{1.2cm}C{1.2cm}C{1.2cm}C{1.2cm}C{1.2cm}C{1.2cm}}
          \toprule
          \multirow{2}{*}{inpainting model}  
          % & \multirow{2}{*}{venue}  
          & \multicolumn{2}{c}{{CLIPScore}}   & \multicolumn{2}{c}{{Aesthetic}}  & \multicolumn{2}{c}{{ImageR}} & \multicolumn{2}{c}{{PickScore}}  & \multicolumn{2}{c}{{HPSv2}}   & \multicolumn{2}{c}{{VQAScore}} & \multicolumn{2}{c}{{UnifiedR}} & \multicolumn{2}{c}{{Perception}} & \multicolumn{2}{c}{{HPSv3}} & \multicolumn{2}{c}{\cellcolor{sem!15}{GPT-4}}     \\ 
                                             \cmidrule(lr){2-3} \cmidrule(lr){4-5} \cmidrule(lr){6-7} \cmidrule(lr){8-9} \cmidrule(lr){10-11} \cmidrule(lr){12-13} \cmidrule(lr){14-15} \cmidrule(lr){16-17} \cmidrule(lr){18-19} \cmidrule(lr){20-21}
                                       & Brush. & Edit. & Brush. & Edit. & Brush. & Edit. & Brush. & Edit. & Brush. & Edit. & Brush. & Edit. & Brush. & Edit. & Brush. & Edit. & Brush. & Edit. & \cellcolor{sem!15}Brush. & \cellcolor{sem!15}Edit. \\
          \midrule
          SDI
          &                    26.304 &                  26.526 &              6.368 &                    5.377 &              12.026 &               -1.100 &                 22.105 &                20.791 &                   27.079 &               23.203 &                8.981 &             6.923 &                3.268 &              2.069 &                26.190 &                25.382 &                  5.320 &                  0.849 &                    \cellcolor{sem!15}79.004 &                    \cellcolor{sem!15}60.751 \\
          CNI
          &                    26.341 &                  26.972 &              6.305 &                    5.382 &              11.421 &               -1.044 &                 21.953 &                20.874 &                   26.633 &               23.076 &                8.890 &             6.906 &                3.218 &              2.125 &                26.150 &                25.894 &                  4.546 &                  0.894 &                    \cellcolor{sem!15}74.173 &                    \cellcolor{sem!15}63.921 \\
          BLD
          &                    26.337 &      27.666 &              6.262 &                    5.372 &              11.161 &                0.563 &                 21.901 &                20.980 &                   26.723 &               23.839 &                8.852 &           {7.467} &                3.202 &            {2.228} &                26.128 &    \underline{27.093} &                  4.559 &                  1.114 &                    \cellcolor{sem!15}71.794 &                    \cellcolor{sem!15}62.690 \\
          PowerPoint
          &                  {26.265} &                {27.291} &              6.312 &                    5.448 &              11.771 &              {0.720} &                 22.089 &                20.912 &                   27.065 &               23.347 &                8.931 &           {7.238} &                3.271 &            {2.219} &                26.123 &                26.264 &                  5.112 &                  1.068 &                    \cellcolor{sem!15}78.241 &                  \cellcolor{sem!15}63.092   \\
          PrefPaint
          &                    26.268 &                  25.569 &              6.377 &                    5.296 &              11.798 &               -3.023 &                 22.125 &                20.666 &                   26.855 &               22.241 &                8.925 &             6.226 &                3.271 &              1.951 &                26.116 &                24.264 &                  5.208 &                 -0.336 &                    \cellcolor{sem!15}80.327 &                  \cellcolor{sem!15}60.815   \\
          StrDiffusion
          &                    23.872 &                  21.398 &              5.330 &                    4.405 &              -0.063 &              -16.282 &                 20.342 &                19.147 &                   21.431 &               16.616 &                7.281 &             3.662 &                2.417 &              1.236 &                23.381 &                20.102 &                 -2.243 &                 -7.131 &                    \cellcolor{sem!15}34.255 &                    \cellcolor{sem!15}25.200 \\
          HD-Painter
          &           26.367 &                  26.934 &            {6.480} &        \underline{5.640} &            {12.913} &                0.046 &               {22.314} &              {21.016} &                 {27.931} &             {23.951} &                9.019 &             6.682 &              {3.349} &              2.136 &              {26.214} &                25.721 &                {6.224} &                {1.983} &        \cellcolor{sem!15}\underline{85.016} &      \cellcolor{sem!15}\underline{69.087}   \\
          ASUKA
          &                    24.387 &                  20.842 &              6.294 &                    5.078 &               5.208 &              -14.110 &                 21.601 &                19.603 &                   25.285 &               18.702 &                7.681 &             3.631 &                2.862 &              1.282 &                23.959 &                19.017 &                  3.556 &                 -3.506 &                    \cellcolor{sem!15}75.140 &                  \cellcolor{sem!15}58.686   \\
          \hline
          {BrushNet}                            
          &                    26.415 &                  27.337 &              6.425 &                    5.392 &              12.717 &               -1.296 &                 22.133 &                20.616 &                   27.509 &               23.076 &                9.060 &             6.770 &                3.303 &              2.100 &    \underline{26.290} &              {26.410} &                  5.749 &                  0.403 &                    \cellcolor{sem!15}79.391 &                  \cellcolor{sem!15}57.046   \\
          \textbf{BruPA (ours) }                            
          % &                    - 
          &        \textbf{26.547} &                \underline{27.694} &  \underline{6.516} &                  {5.577} &  \underline{13.315} &      \textbf{10.463} &                 22.279 &                20.844 &       \underline{28.037} &               23.933 &    \underline{9.093} &             7.043 &    \underline{3.371} &              2.193 &       \textbf{26.390} &              {26.881} &      \underline{6.276} &                  1.398 &                    \cellcolor{sem!15}83.054 &                  \cellcolor{sem!15}61.186   \\
          FLUX.1 Fill               
          &                    26.244 &                  27.103 &              6.429 &                    5.458 &              12.760 &                4.910 &     \underline{22.327} &    \underline{21.211} &                   27.476 &   \underline{24.076} &                9.081 & \underline{8.021} &                3.360 &  \underline{2.485} &                25.945 &                26.834 &                  6.055 &      \underline{2.470} &                  \cellcolor{sem!15}{83.935} &      \cellcolor{sem!15}{         {66.979}}  \\
          \textbf{FluPA (ours)}                                   
          &                  \underline{26.436} &         \textbf{27.813} &     \textbf{6.546} &           \textbf{5.681} &     \textbf{13.859} &    \underline{7.707} &        \textbf{22.577} &       \textbf{21.559} &          \textbf{28.735} &      \textbf{25.972} &       \textbf{9.152} &    \textbf{8.434} &       \textbf{3.457} &     \textbf{2.649} &                26.096 &       \textbf{27.617} &         \textbf{7.000} &         \textbf{4.230} &           \cellcolor{sem!15}\textbf{87.609} &         \cellcolor{sem!15}\textbf{72.307}   \\
          \bottomrule
          
     \end{tabular}
     \begin{tablenotes}
     \item \textbf{Bold} values denote the best results. \underline{Underlined} values denote the second-best results. All methods are evaluated using official implementations with blending~\citep{BrushNet}. 
     \end{tablenotes}
}
\end{threeparttable}
}
\label{tab:comparison_with_sota}
\vspace{-6mm}
\end{table*}

We identify potential biases in reward models that may lead to reward hacking, as discussed in \autoref{sec:effective} and \autoref{sec:scalable}. In this section, we delve deeper into exploring these intriguing biases---examining their nature and how they make reward hacking happen. To investigate it, we sample inpainting examples in~\autoref{fig:bias} and~\autoref{fig:baseline}. We report the following findings and insights.

\textbf{Reward models exhibit biases in brightness, composition, and color scheme.}
As evidenced by the results from HPSv2 and PickScore---the second and third images in each sub-figure of~\autoref{fig:bias} respectively, we observe notable biases in their preferences. HPSv2 tends to favor images with bright lighting, complex composition with rich details, and vivid colors. In contrast, PickScore shows a preference for dim lighting, simple composition with few details, and muted colors.

\textbf{Biases in reward models affect different baseline models in distinct ways.}
Although each reward model has its own inherent biases, we find that their influence varies across baseline models. For instance, BrushNet trained using HPSv2 produces inpainting outputs characterized by excessively bright lighting, overly intricate details, and unnaturally vivid colors---they seem deviate from human aesthetic preferences. In contrast, FLUX.1 Fill trained using HPSv2 generates visually pleasing results. PickScore shows a similar disparity in performance. 
It stems from the characteristics of baseline models as shown in~\autoref{fig:baseline}: BrushNet generates vibrant images, making PickScore particularly suitable for it; while FLUX.1 Fill produces plain images, aligning with HPSv2's property.

\textbf{Ensemble shows generality and generalization by mitigating biases.}
Ensemble, a simple and straightforward method implemented through reward ensembling, exhibits strong versatility across models by producing balanced and aesthetically pleasing inpainting results, as shown in~\autoref{fig:bias} and~\autoref{fig:baseline}. It likely stems from Ensemble’s ability to mitigate biases inherent in reward models.

\section{Ablation Studies and Comparisons with State-of-the-Art}
\begin{figure*}[t]
\centering
\includegraphics[width=\linewidth, trim=0mm 0mm 0mm 0mm, clip]{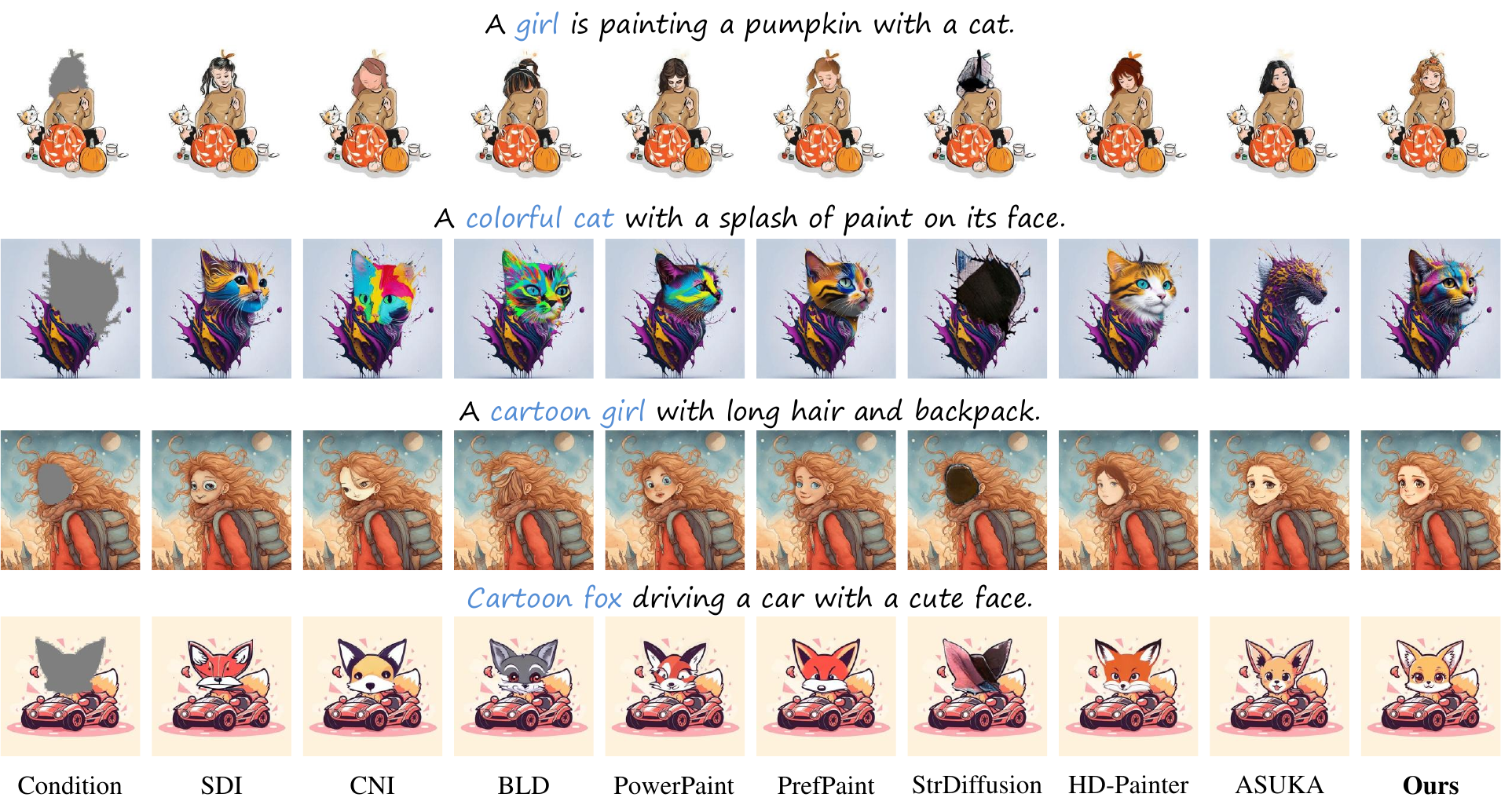}
\vspace{-6mm}
\caption{Qualitative comparisons with state-of-the-art image inpainting models.}
\label{fig:sota}
\vspace{-4mm}
\end{figure*}

\begin{figure*}[t]
\centering
\includegraphics[width=\linewidth, trim=7mm 7mm 7mm 7mm, clip]{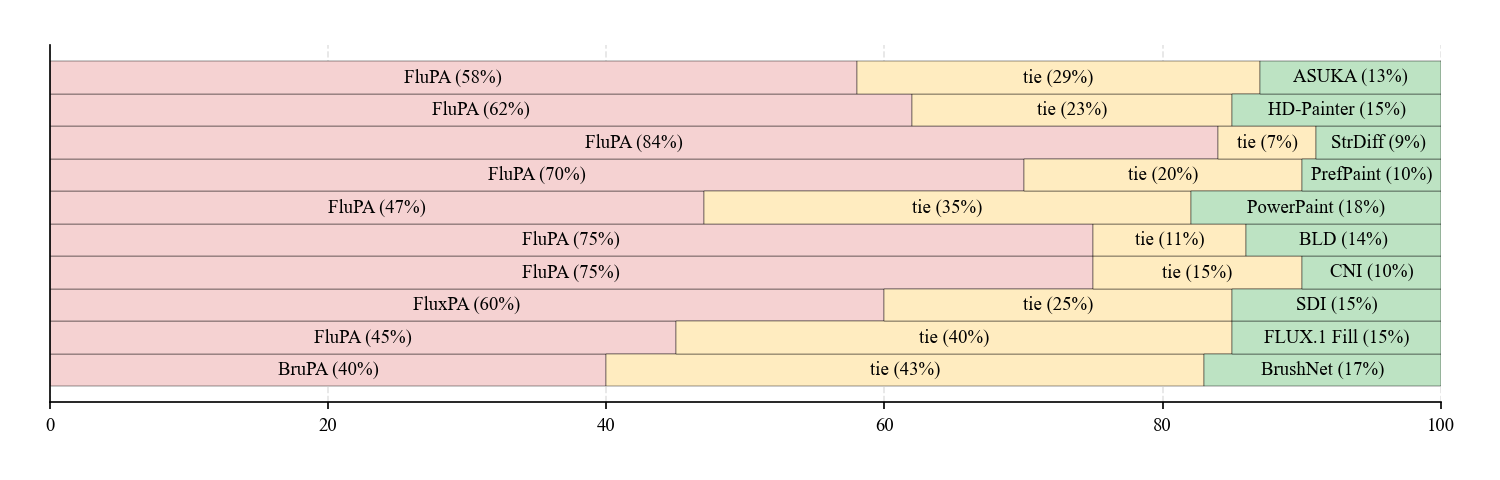}
\vspace{-6mm}
\caption{\textbf{User studies.} 
We compare each pair of models by randomly sampling 100 pairs from their inpainting results. We invite 30 volunteers to participate in a blind assessment to determine which one is better (``A win'', ``B win'', or ``tie'') based on their preferences. We report the \textbf{winning rates}.}
\label{fig:user}
\vspace{-6mm}
\end{figure*}

We name our methods \textbf{BruPA} and \textbf{FluPA}---BrushNet and FLUX.1 Fill with Ensemble-based preference alignment. We compare them with state-of-the-art image inpainting models. Specifically, the following methods are compared (they are introduced in~\autoref{sec:related-work}): SDI~\citep{sd}, CNI~\citep{CNI}, BLD~\citep{BLD}, PowerPaint~\citep{powerpaint}, BrushNet~\citep{BrushNet}, PrefPaint~\citep{Prefpaint}, StrDiffusion~\citep{StructureMatters}, FLUX.1 Fill~\citep{flux1filldev}, HD-Painter~\citep{HD-Painter}, and ASUKA~\citep{ASUKA}, where \textit{BrushNet and FLUX.1 Fill are also the baseline models for ablation studies}.

\textbf{Ablation studies.}
We report quantitative and qualitative ablation studies, i.e., before and after preference alignment training, in~\autoref{tab:comparison_with_sota} and~\autoref{fig:baseline}, respectively. After preference alignment using Ensemble, our method significantly surpasses the baseline models by achieving much better results and yielding visually appealing results. We further conduct ablation studies on a new dataset, reported in~\autoref{tab:comparison_on_I_DREAM_MY_PAINT}. It also confirms that our improvement is generalizable across data distributions.

\textbf{Comparisons with state-of-the-art.}
\autoref{tab:comparison_with_sota} reports the results. 
Our BruPA and FluPA set new state-of-the-arts, attaining the best results on all evaluations and the second-best results in nearly half of the cases. Notably, even on coarser metrics—CLIPScore, VQAScore, and Perception (analyzed in \autoref{sec:effective})—our methods still outperform competitors. Besides, BruPA and FluPA significantly outperform BrushNet and FLUX.1 Fill, i.e, the baselines before applying preference alignment. The qualitative results are reported in~\autoref{fig:sota}, and our model generates images with better aesthetics.

\textbf{User studies.} As shown in~\autoref{fig:user}, our models align with human preferences better.

\section{Conclusion and Limitation Discussions}
We conduct extensive studies on image inpainting with preference alignment and obtain key insights into the effectiveness, scalability, and challenges in achieving alignment. We find that a simple ensemble method mitigate biases and achieve non-trivial results. Yet, our work is confined to images and DPO training; future extensions can generalize these findings to video, 3D data, and RLHF.

\clearpage

\textbf{Reproducibility statement.}
We conduct experiments using the official implementations of BrushNet~\citep{BrushNet} and FLUX.1 Fill~\citep{flux1filldev}. The DPO loss is implemented based on the official code from~\cite{Diffusion-DPO}. Additional implementation details are also provided in~\autoref{sec:effective}, ~\autoref{sec:scalable}, and the Appendix. Our code will be open-sourced to ensure reproducibility.

\bibliography{iclr2026_conference}
\bibliographystyle{iclr2026_conference}

\clearpage
\appendix
\section{Details to Make GPT-4 as an Evaluator}
Given GPT-4's~\citep{GPT-4} strong multi-modal understanding capabilities, we use it to evaluate image inpainting results. Specifically, we provide GPT-4 with (1) a system prompt, (2) an input image, (3) the mask to inpaint, (4) an inpainting prompt, (5) and the inpainting result. 

We randomly choose 500 inpainting pairs and invite volunteers to determine which one is better. GPT-4 achieves 86\% accuracy; HPSv2: 82\%, PickScore: 80\%, ImageReward: 77\%, Aesthetic: 80\%.

The system prompt designed by us is given below:

\begin{tcolorbox}[colback=gray!10,colframe=gray!80,title={GPT-4 System Prompt for Image Inpainting Evaluation}]
You are a human expert in analysis of image inpainting.
Please evaluate the image inpainting result based on the following three criteria:
\begin{itemize}
    \item Aesthetic Quality (0–40 points):
    \begin{itemize}
        \item Visual appeal in color harmony, composition, style coherence
        \item Texture realism and naturalness
    \end{itemize}
    \item Structural Coherence (0–30 points)
    \begin{itemize}
        \item Preservation of geometric structures and content continuity
        \item Seamlessness at mask boundaries
    \end{itemize}
    \item Semantic Alignment (0–30 points)
    \begin{itemize}
        \item Faithfulness to the Text Prompt instructions
        \item Contextual consistency of added or restored content
    \end{itemize}
\end{itemize}

For each criterion, provide:
\begin{itemize}
    \item A sub‑score.
    \item A 1–2‑sentence justification.
\end{itemize}
Then compute the total score (0–100).

\end{tcolorbox}

\section{Searches of Hyper-Parameters}
\begin{figure*}[h]
  \begin{subfigure}[t]{\linewidth}
    \centering
    % \vspace{2mm}
    \includegraphics[width=0.195\linewidth, trim=5mm 5mm 5mm 5mm, clip]{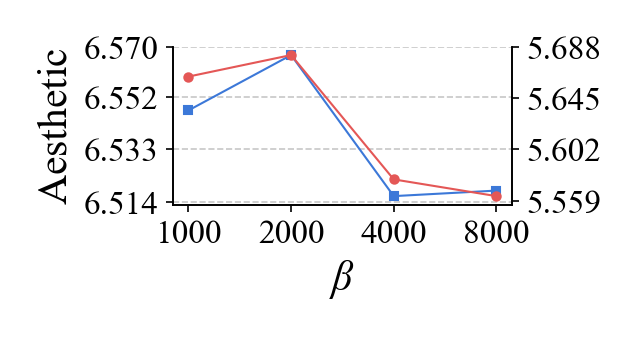}
    \includegraphics[width=0.195\linewidth, trim=5mm 5mm 5mm 5mm, clip]{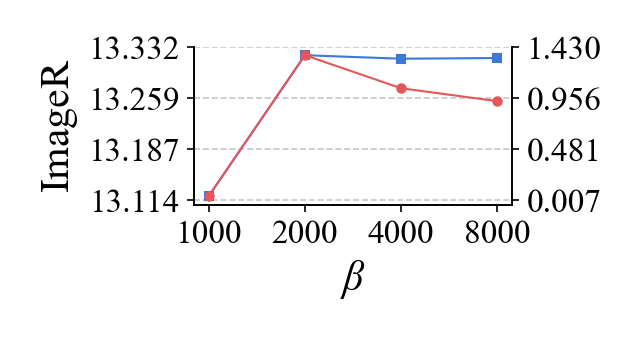}
    \includegraphics[width=0.195\linewidth, trim=5mm 5mm 5mm 5mm, clip]{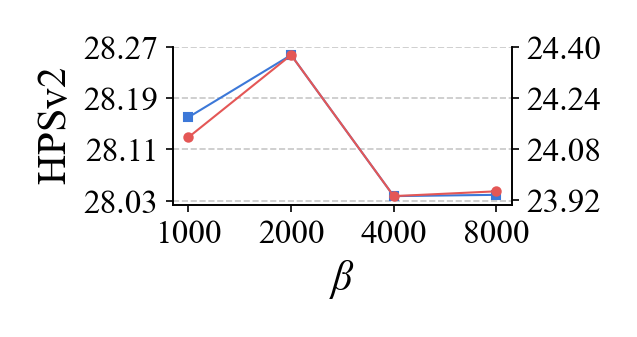}
    \includegraphics[width=0.195\linewidth, trim=5mm 5mm 5mm 5mm, clip]{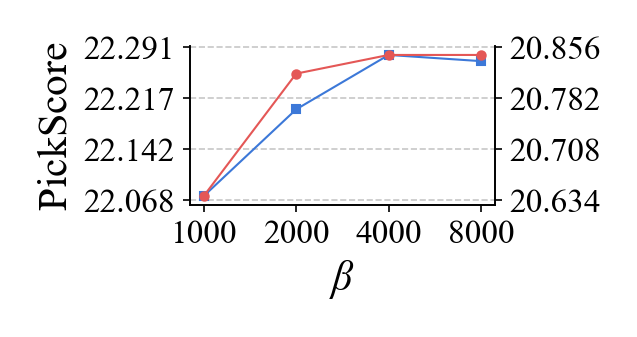}
    \includegraphics[width=0.195\linewidth, trim=5mm 5mm 5mm 5mm, clip]{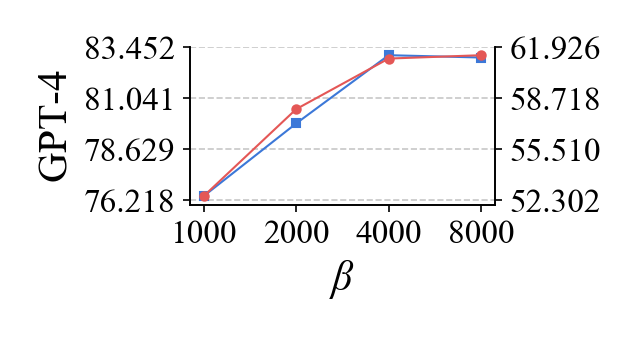}
    \includegraphics[width=0.195\linewidth, trim=5mm 5mm 5mm 5mm, clip]{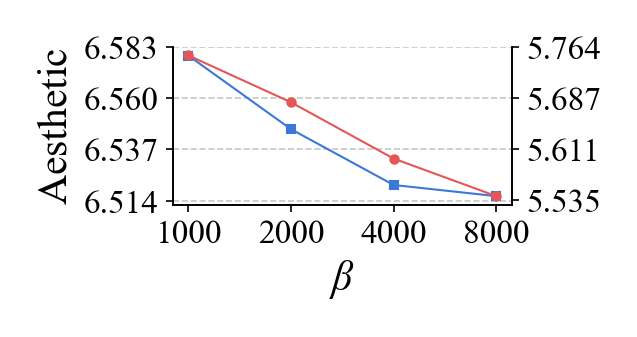}
    \includegraphics[width=0.195\linewidth, trim=5mm 5mm 5mm 5mm, clip]{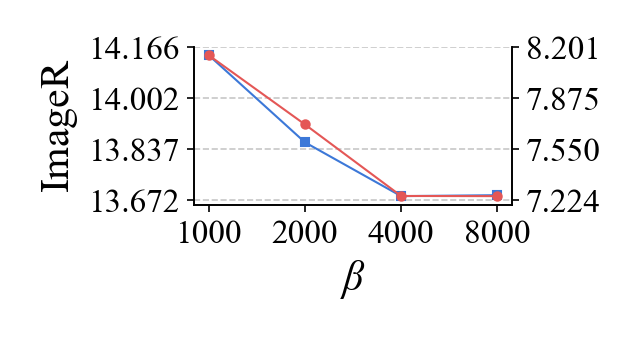}
    \includegraphics[width=0.195\linewidth, trim=5mm 5mm 5mm 5mm, clip]{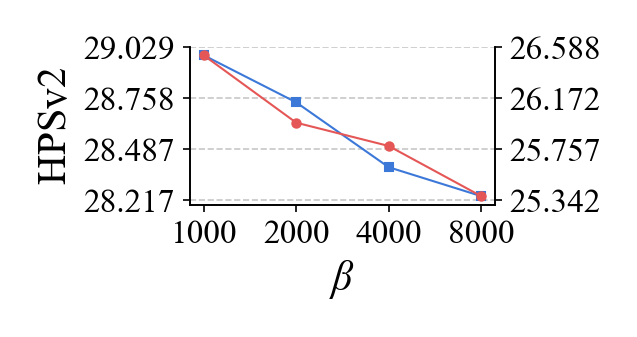}
    \includegraphics[width=0.195\linewidth, trim=5mm 5mm 5mm 5mm, clip]{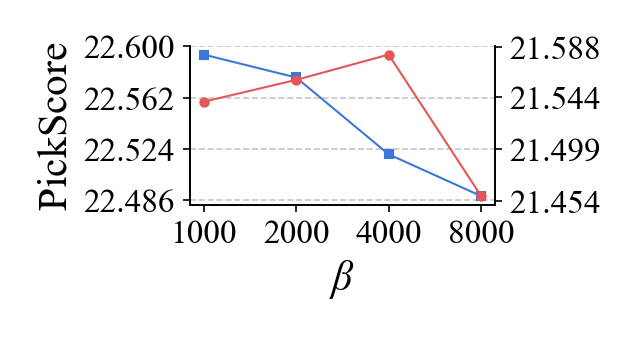}
    \includegraphics[width=0.195\linewidth, trim=5mm 5mm 5mm 5mm, clip]{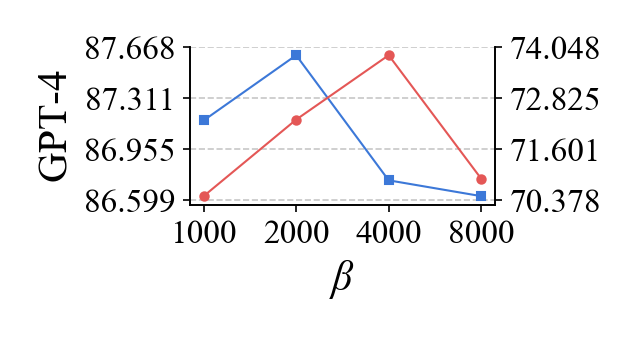}
    \vspace{-6mm}
    \caption{Sensitivity analysis on $\beta$.}
\end{subfigure}
\begin{subfigure}[t]{\linewidth}
    \centering
    % \vspace{2mm}
    \includegraphics[width=0.195\linewidth, trim=5mm 5mm 5mm 5mm, clip]{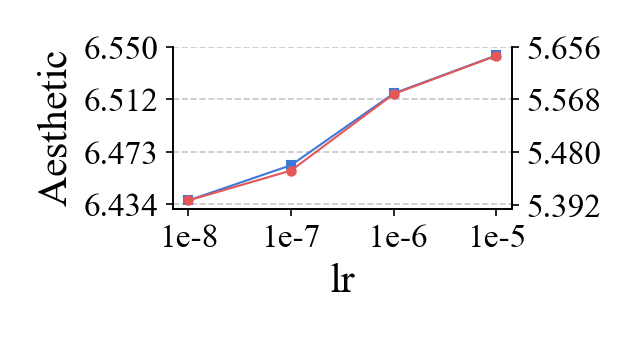}
    \includegraphics[width=0.195\linewidth, trim=5mm 5mm 5mm 5mm, clip]{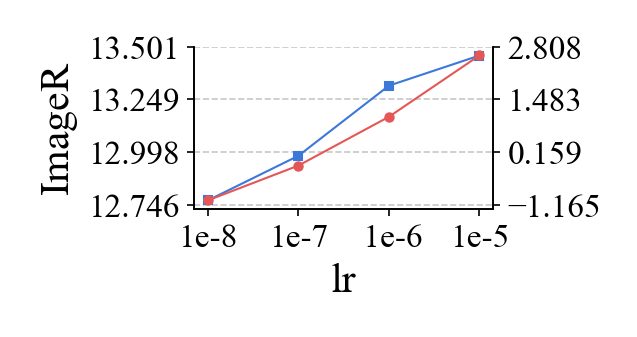}
    \includegraphics[width=0.195\linewidth, trim=5mm 5mm 5mm 5mm, clip]{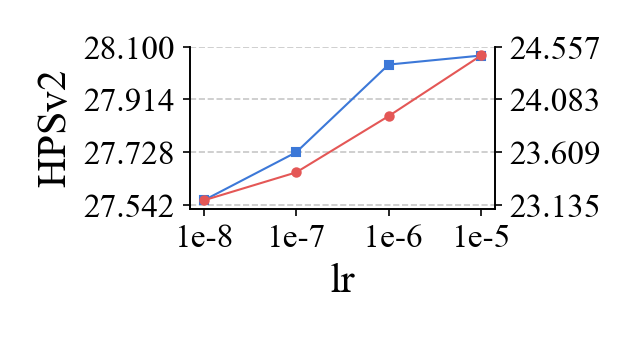}
    \includegraphics[width=0.195\linewidth, trim=5mm 5mm 5mm 5mm, clip]{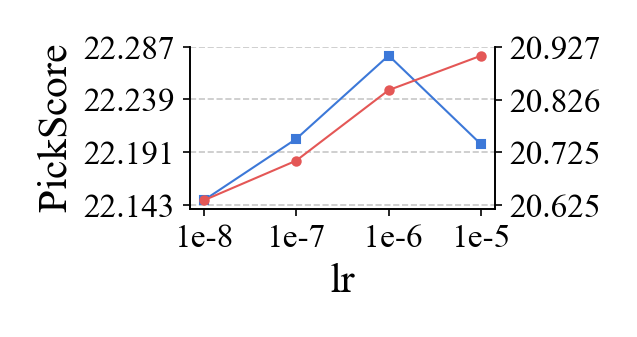}
    \includegraphics[width=0.195\linewidth, trim=5mm 5mm 5mm 5mm, clip]{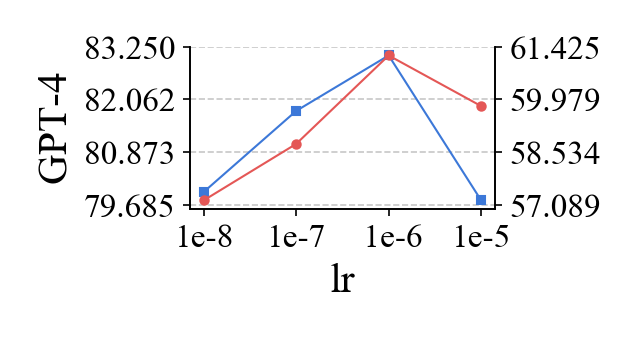}
    \includegraphics[width=0.195\linewidth, trim=5mm 5mm 5mm 5mm, clip]{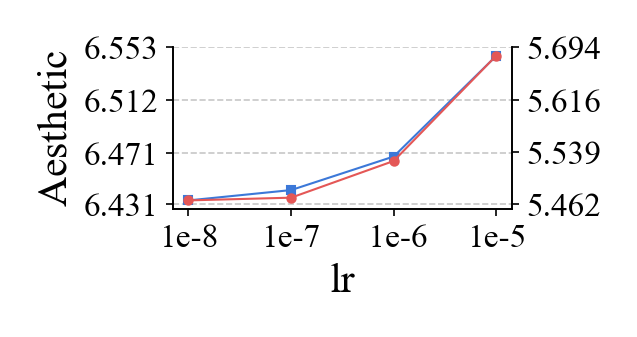}
    \includegraphics[width=0.195\linewidth, trim=5mm 5mm 5mm 5mm, clip]{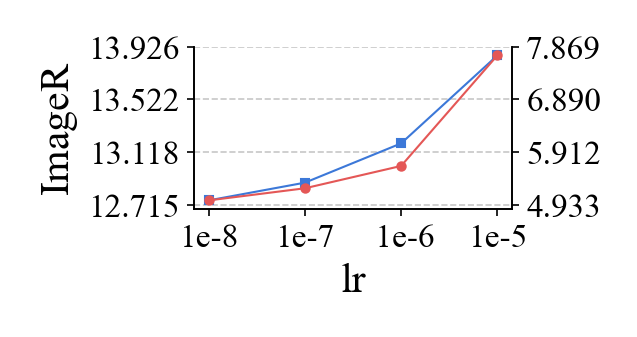}
    \includegraphics[width=0.195\linewidth, trim=5mm 5mm 5mm 5mm, clip]{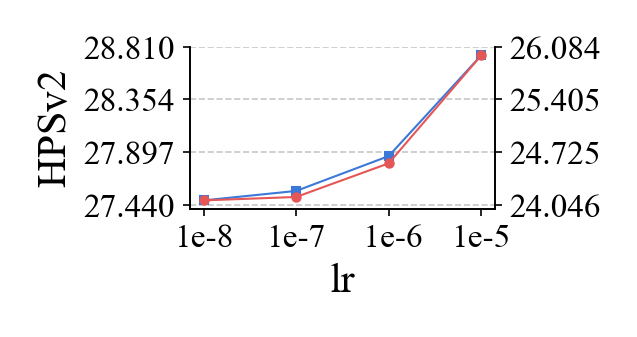}
    \includegraphics[width=0.195\linewidth, trim=5mm 5mm 5mm 5mm, clip]{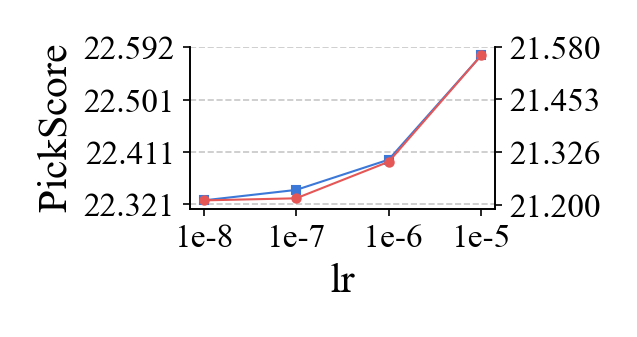}
    \includegraphics[width=0.195\linewidth, trim=5mm 5mm 5mm 5mm, clip]{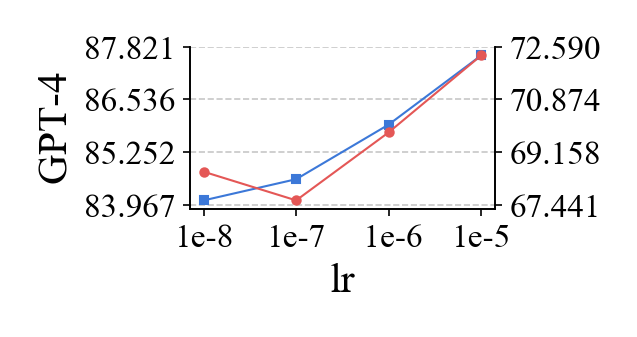}
    \vspace{-6mm}
    \caption{Sensitivity analysis on lr.}
\end{subfigure}
\vspace{-3mm}
\caption{\textbf{Searches} of $\beta$, i.e., sub-figure (a), and learning rate (lr), i.e., sub-figure (b). The first and second row of each sub-figure is based on \textbf{BrushNet} and \textbf{FLUX.1 Fill}, respectively.}
\label{fig:lr-beta}
\vspace{-4mm}
\end{figure*}

We conduct hyper-parameter searches for Ensemble, as shown in~\autoref{fig:lr-beta}. For Ensemble, we finally adopt a learning rate of 1e-6, and set $\beta=4000$ for BrushNet; while using a learning rate of 1e-5, and set $\beta=2000$ for FLUX.1 Fill. For HPSv2, we use a learning rate of 1e-5 with $\beta=4000$ for BrushNet; and a learning rate of 1e-6 with $\beta=8000$ on FLUX.1 Fill. For PickScore, we set the learning rate to 1e-7 and use $\beta=2000$ for both BrushNet and FLUX.1 Fill.

\section{More Results on Reward Model Bias Studies}
\begin{figure}[!htbp]
\begin{subfigure}[t]{.49\linewidth}
\centering
	\includegraphics[width=0.24\linewidth, trim=0mm 0mm 0mm 0mm, clip]{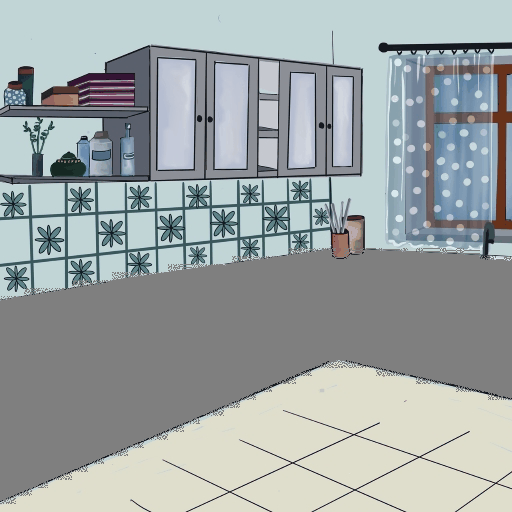}
	\includegraphics[width=0.24\linewidth, trim=0mm 0mm 0mm 0mm, clip]{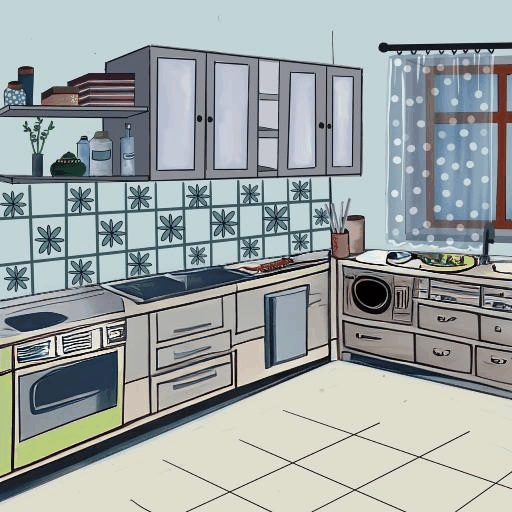}
	\includegraphics[width=0.24\linewidth, trim=0mm 0mm 0mm 0mm, clip]{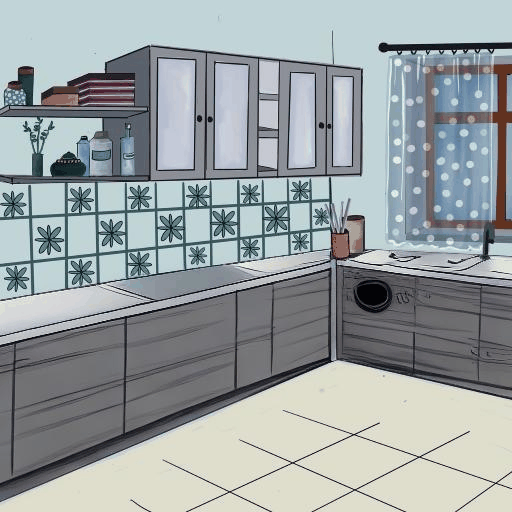}
	\includegraphics[width=0.24\linewidth, trim=0mm 0mm 0mm 0mm, clip]{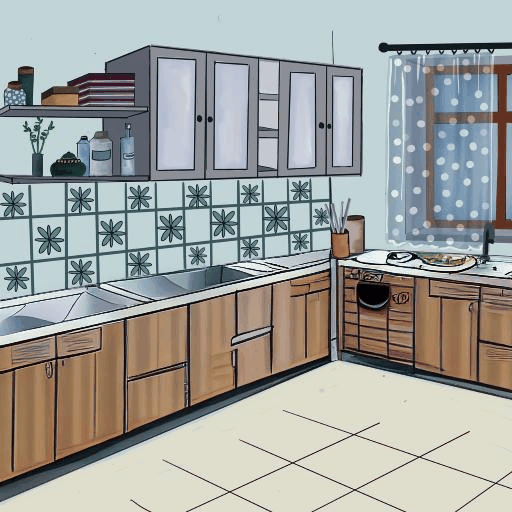}
\caption{A cartoon drawing of a kitchen.}
\end{subfigure}
\begin{subfigure}[t]{.49\linewidth}
\centering
	\includegraphics[width=0.24\linewidth, trim=0mm 0mm 0mm 0mm, clip]{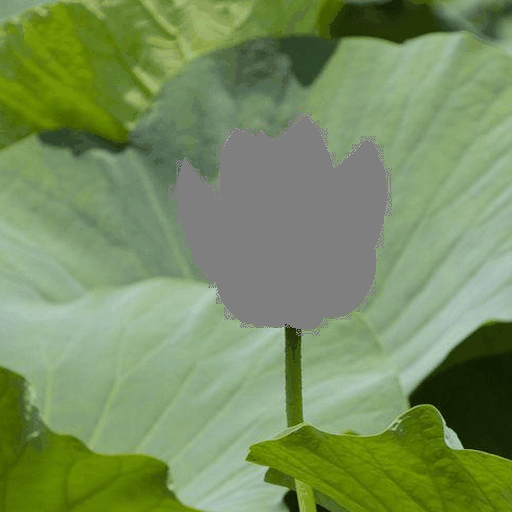}
	\includegraphics[width=0.24\linewidth, trim=0mm 0mm 0mm 0mm, clip]{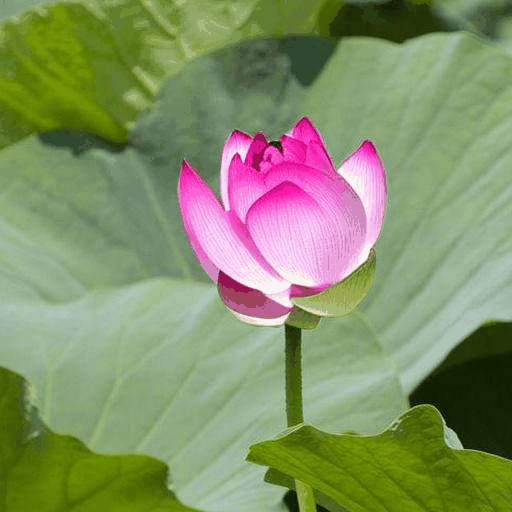}
	\includegraphics[width=0.24\linewidth, trim=0mm 0mm 0mm 0mm, clip]{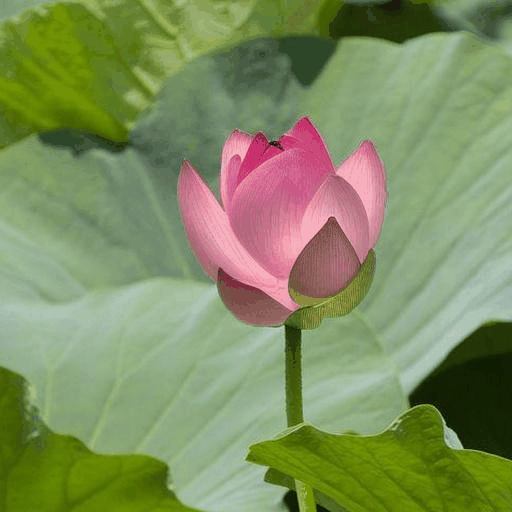}
	\includegraphics[width=0.24\linewidth, trim=0mm 0mm 0mm 0mm, clip]{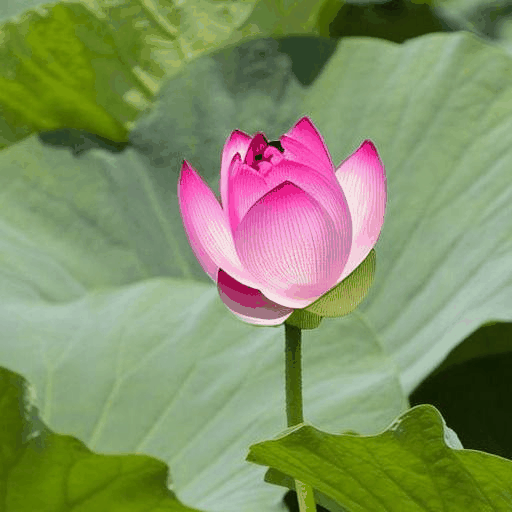}
\caption{A pink lotus flower blooming with green leaves.}
\end{subfigure} 
\begin{subfigure}[t]{.49\linewidth}
\centering
	\includegraphics[width=0.24\linewidth, trim=0mm 0mm 0mm 0mm, clip]{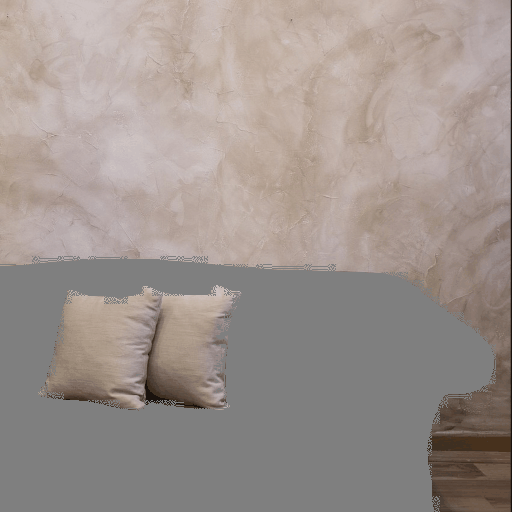}
	\includegraphics[width=0.24\linewidth, trim=0mm 0mm 0mm 0mm, clip]{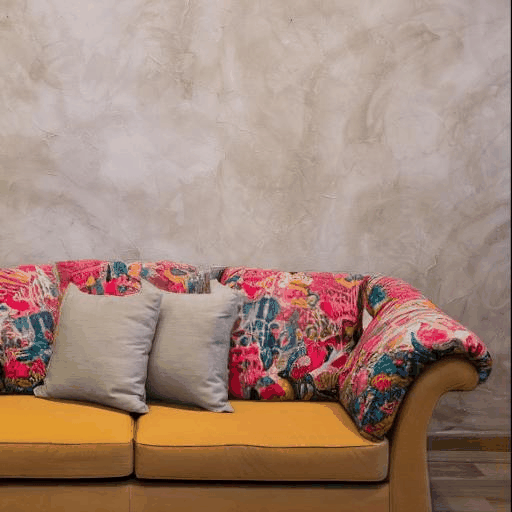}
	\includegraphics[width=0.24\linewidth, trim=0mm 0mm 0mm 0mm, clip]{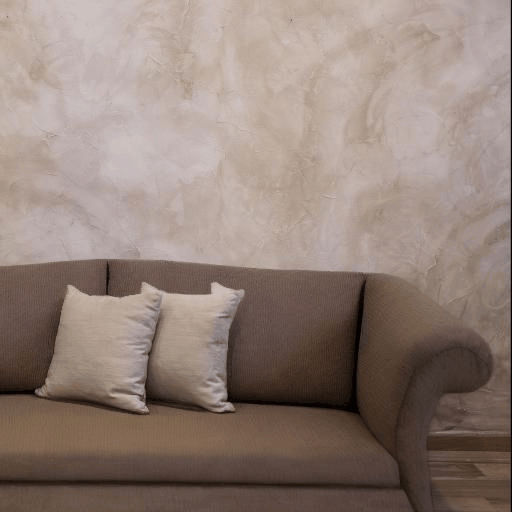}
	\includegraphics[width=0.24\linewidth, trim=0mm 0mm 0mm 0mm, clip]{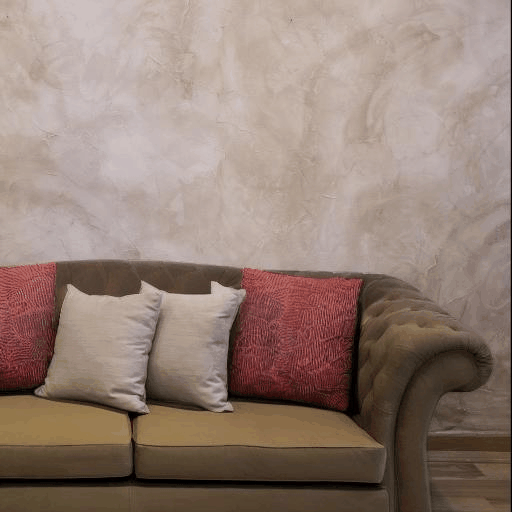}
\caption{A couch with pillows and a wall behind it.}
\end{subfigure} 
\begin{subfigure}[t]{.49\linewidth}
\centering
	\includegraphics[width=0.24\linewidth, trim=0mm 0mm 0mm 0mm, clip]{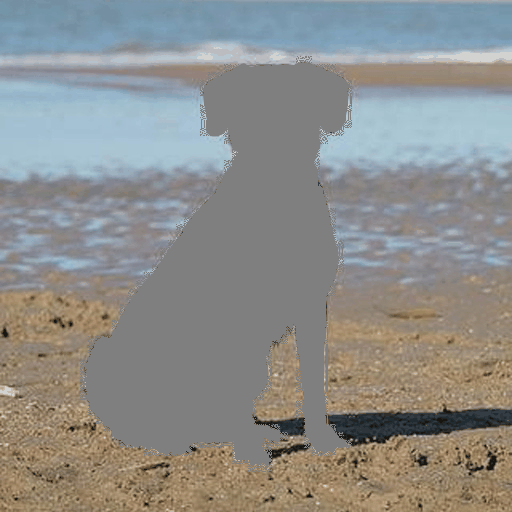}
	\includegraphics[width=0.24\linewidth, trim=0mm 0mm 0mm 0mm, clip]{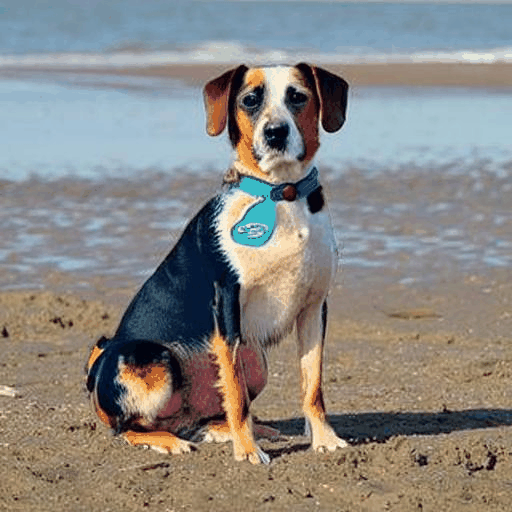}
	\includegraphics[width=0.24\linewidth, trim=0mm 0mm 0mm 0mm, clip]{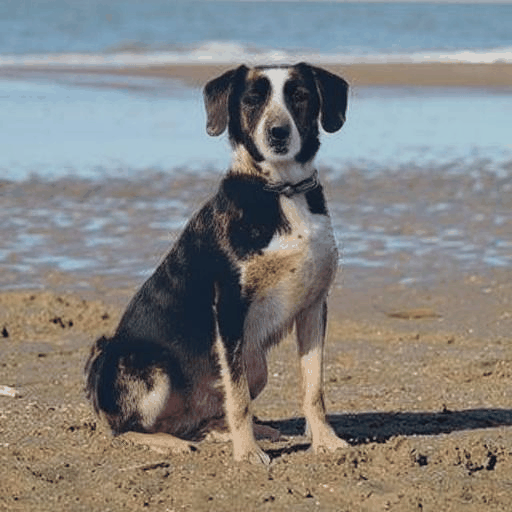}
	\includegraphics[width=0.24\linewidth, trim=0mm 0mm 0mm 0mm, clip]{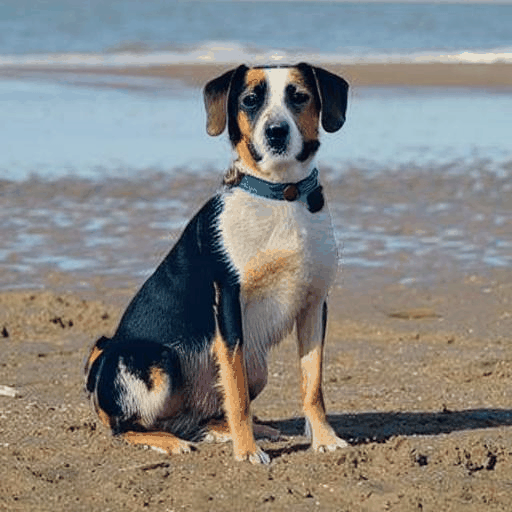}
\caption{A dog sitting on the beach.}
\end{subfigure} 
\begin{subfigure}[t]{.49\linewidth}
\centering
	\includegraphics[width=0.24\linewidth, trim=0mm 0mm 0mm 0mm, clip]{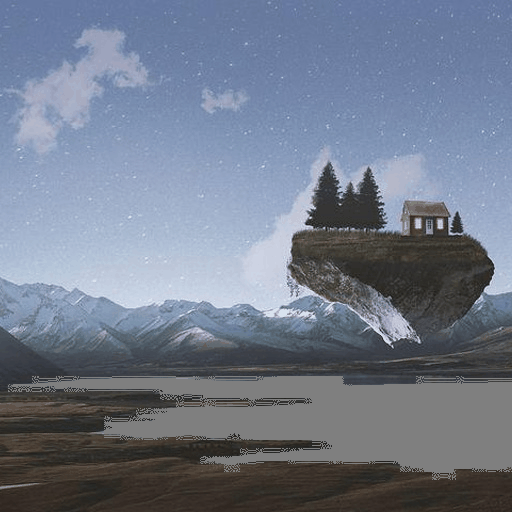}
	\includegraphics[width=0.24\linewidth, trim=0mm 0mm 0mm 0mm, clip]{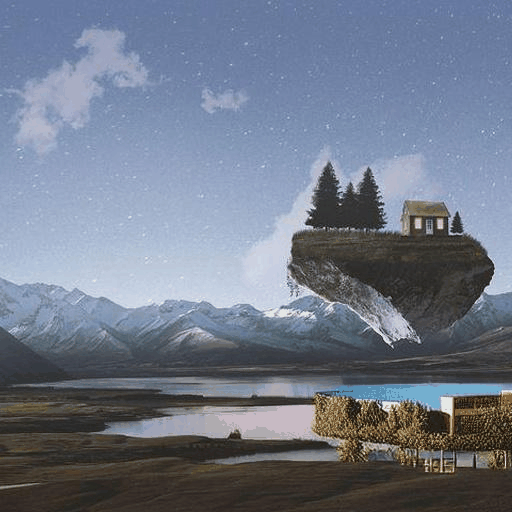}
	\includegraphics[width=0.24\linewidth, trim=0mm 0mm 0mm 0mm, clip]{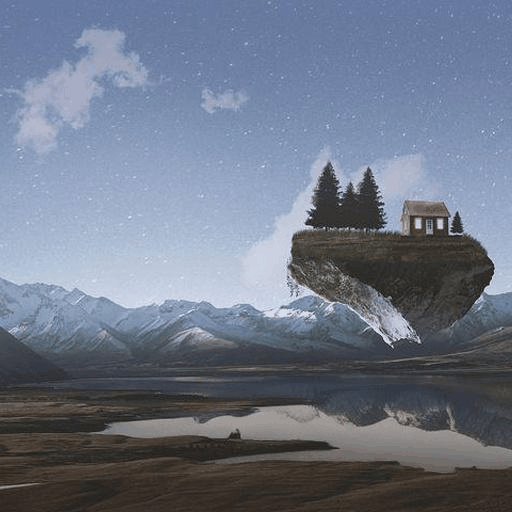}
	\includegraphics[width=0.24\linewidth, trim=0mm 0mm 0mm 0mm, clip]{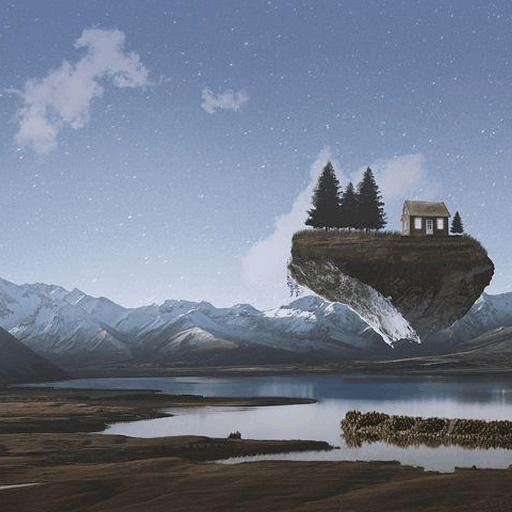}
\caption{A house floating in the air over a lake.}
\end{subfigure}
\begin{subfigure}[t]{.49\linewidth}
\centering
	\includegraphics[width=0.24\linewidth, trim=0mm 0mm 0mm 0mm, clip]{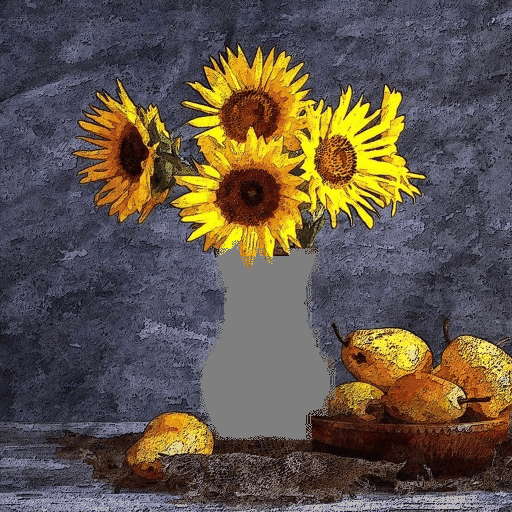}
	\includegraphics[width=0.24\linewidth, trim=0mm 0mm 0mm 0mm, clip]{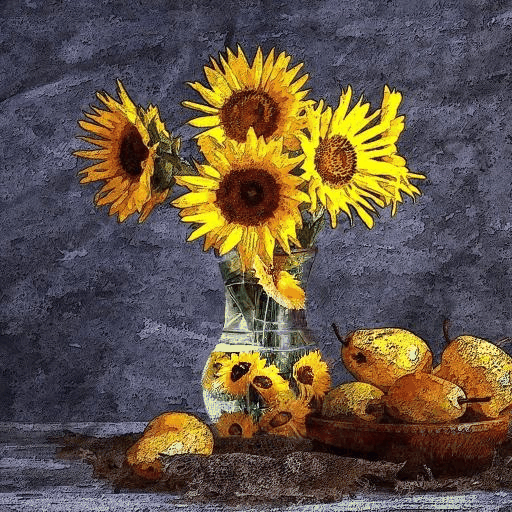}
	\includegraphics[width=0.24\linewidth, trim=0mm 0mm 0mm 0mm, clip]{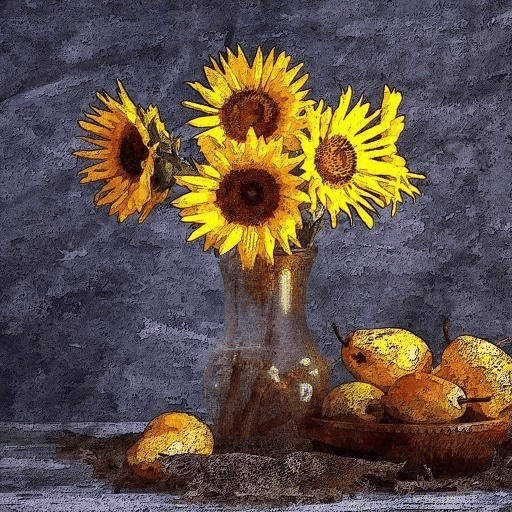}
	\includegraphics[width=0.24\linewidth, trim=0mm 0mm 0mm 0mm, clip]{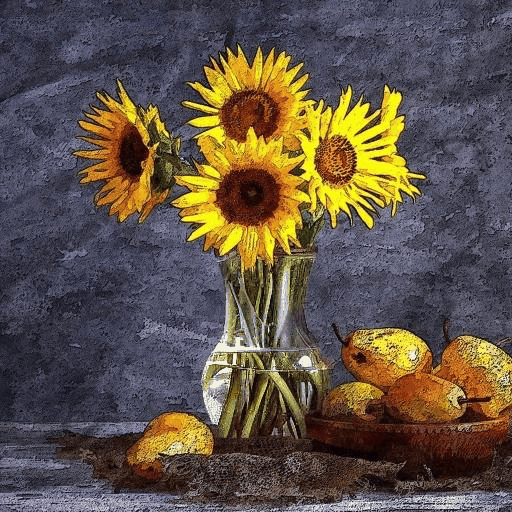}
\caption{Sunflowers in a vase with pears on a table.}
\end{subfigure}
\begin{subfigure}[t]{.49\linewidth}
\centering
	\includegraphics[width=0.24\linewidth, trim=0mm 0mm 0mm 0mm, clip]{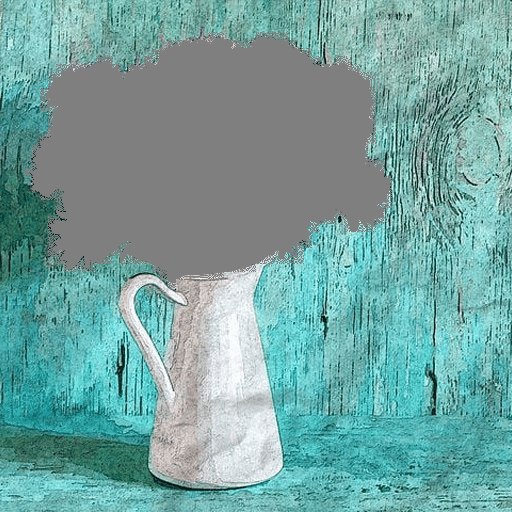}
	\includegraphics[width=0.24\linewidth, trim=0mm 0mm 0mm 0mm, clip]{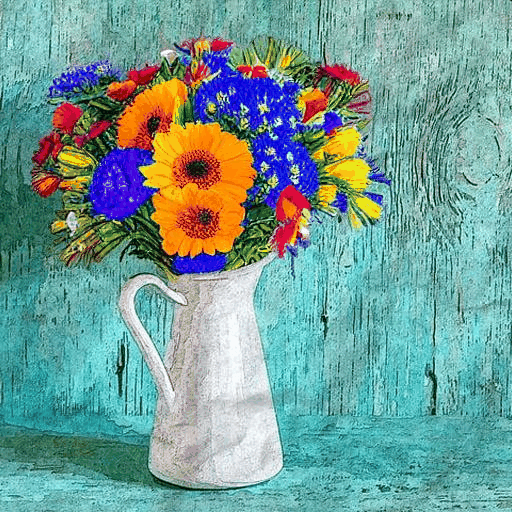}
	\includegraphics[width=0.24\linewidth, trim=0mm 0mm 0mm 0mm, clip]{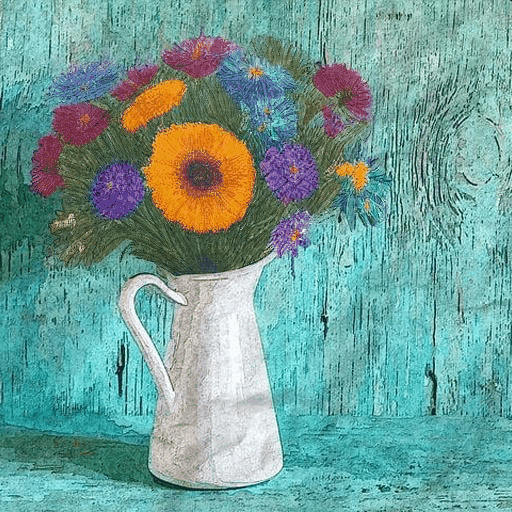}
	\includegraphics[width=0.24\linewidth, trim=0mm 0mm 0mm 0mm, clip]{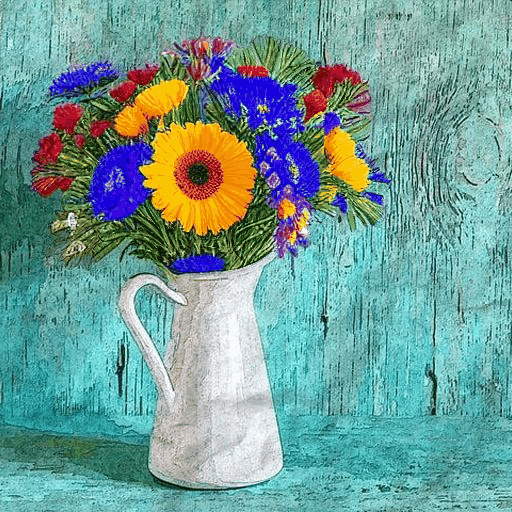}
\caption{A vase filled with colorful flowers on a table.}
\end{subfigure}
\begin{subfigure}[t]{.49\linewidth}
\centering
	\includegraphics[width=0.24\linewidth, trim=0mm 0mm 0mm 0mm, clip]{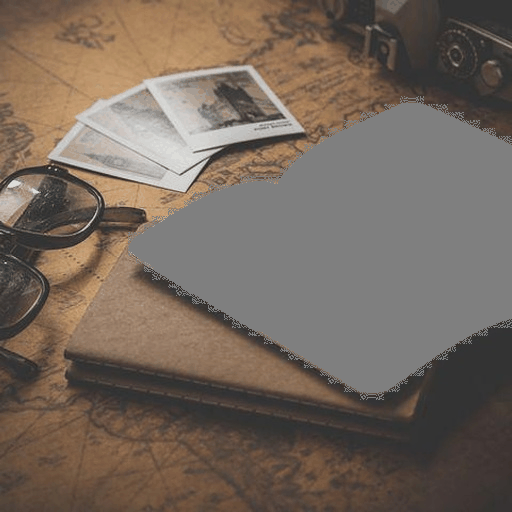}
	\includegraphics[width=0.24\linewidth, trim=0mm 0mm 0mm 0mm, clip]{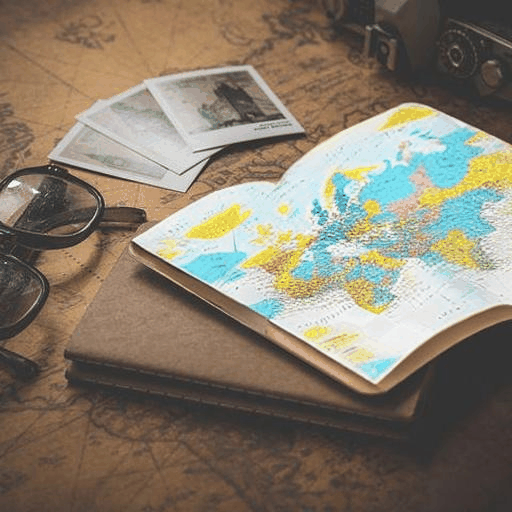}
	\includegraphics[width=0.24\linewidth, trim=0mm 0mm 0mm 0mm, clip]{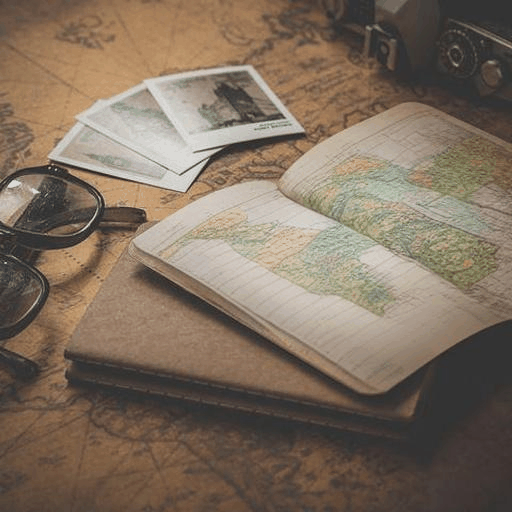}
	\includegraphics[width=0.24\linewidth, trim=0mm 0mm 0mm 0mm, clip]{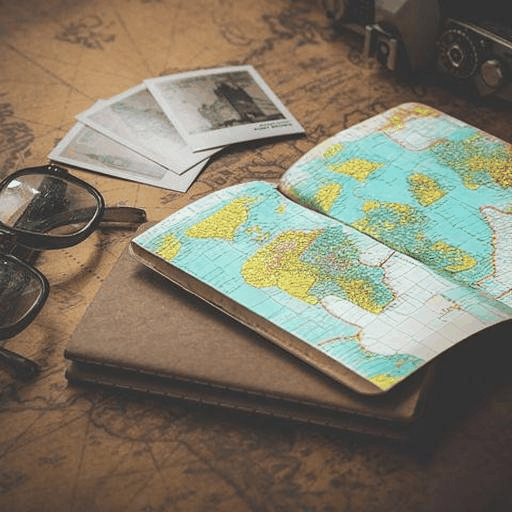}
\caption{A notebook, glasses and a camera on a map.}
\end{subfigure} 
\begin{subfigure}[t]{.49\linewidth}
\centering
	\includegraphics[width=0.24\linewidth, trim=0mm 0mm 0mm 0mm, clip]{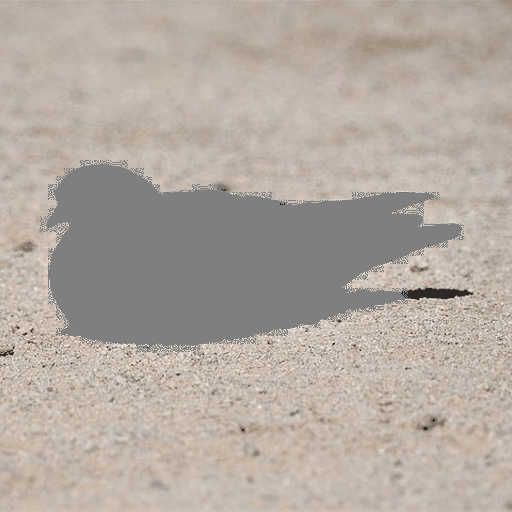}
	\includegraphics[width=0.24\linewidth, trim=0mm 0mm 0mm 0mm, clip]{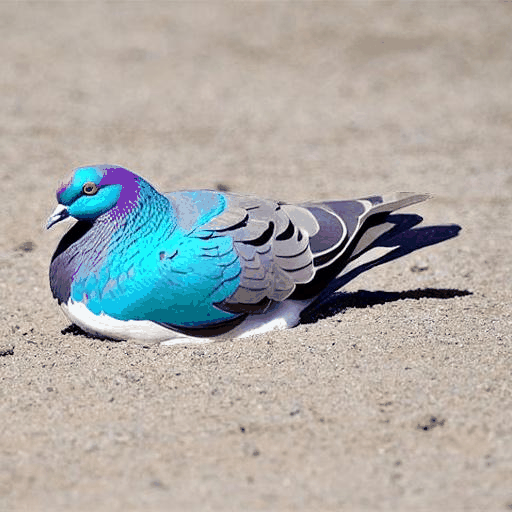}
	\includegraphics[width=0.24\linewidth, trim=0mm 0mm 0mm 0mm, clip]{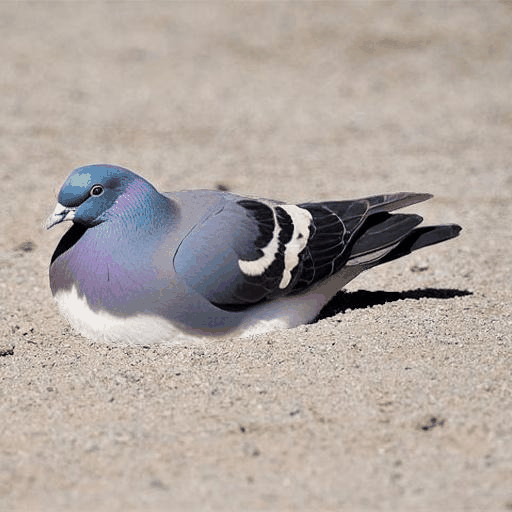}
	\includegraphics[width=0.24\linewidth, trim=0mm 0mm 0mm 0mm, clip]{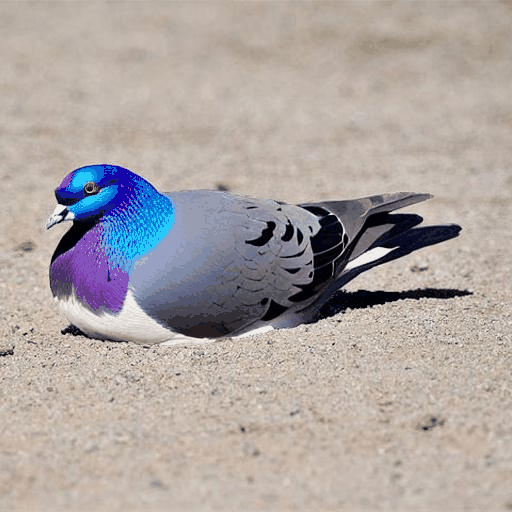}
\caption{A pigeon is sitting on the ground.}
\end{subfigure} 
\begin{subfigure}[t]{.49\linewidth}
\centering
	\includegraphics[width=0.24\linewidth, trim=0mm 0mm 0mm 0mm, clip]{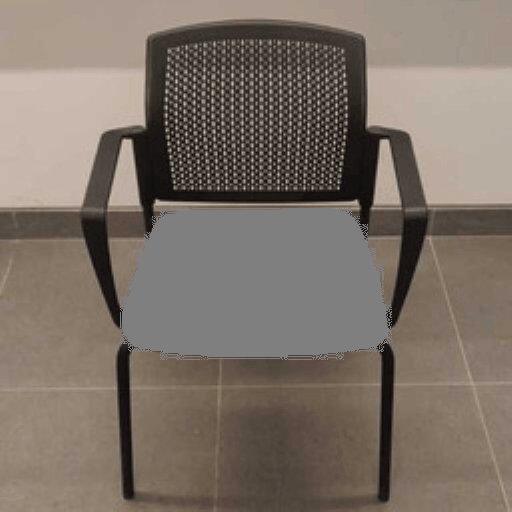}
	\includegraphics[width=0.24\linewidth, trim=0mm 0mm 0mm 0mm, clip]{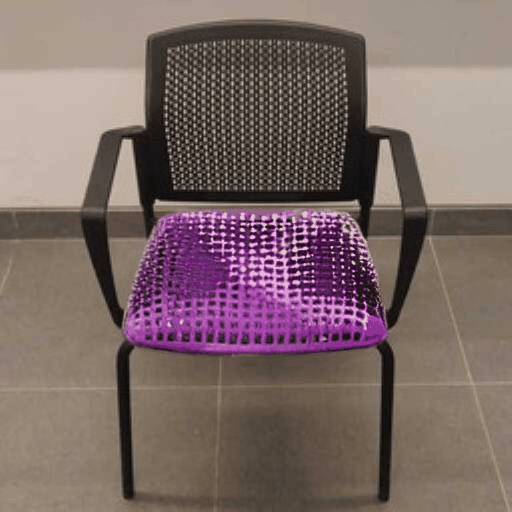}
	\includegraphics[width=0.24\linewidth, trim=0mm 0mm 0mm 0mm, clip]{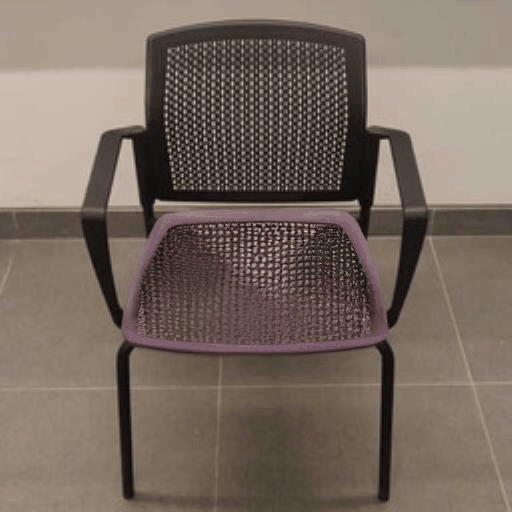}
	\includegraphics[width=0.24\linewidth, trim=0mm 0mm 0mm 0mm, clip]{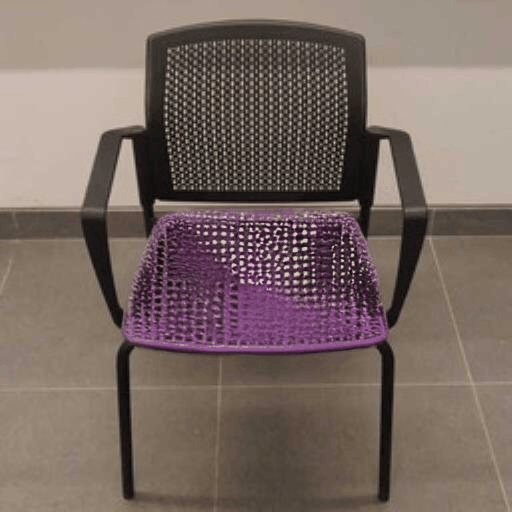}
\caption{A purple chair with a black seat and back.}
\end{subfigure} 
\begin{subfigure}[t]{.49\linewidth}
\centering
	\includegraphics[width=0.24\linewidth, trim=0mm 0mm 0mm 0mm, clip]{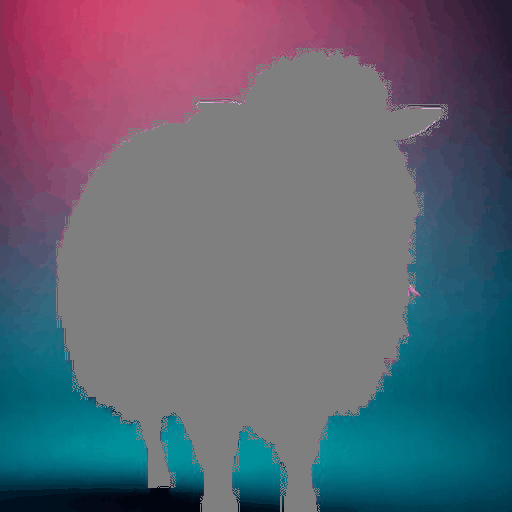}
	\includegraphics[width=0.24\linewidth, trim=0mm 0mm 0mm 0mm, clip]{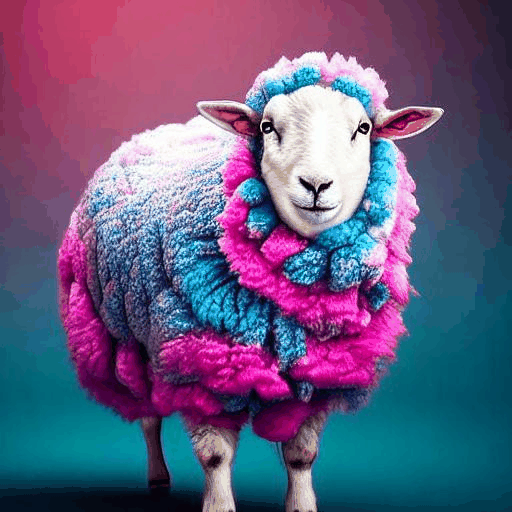}
	\includegraphics[width=0.24\linewidth, trim=0mm 0mm 0mm 0mm, clip]{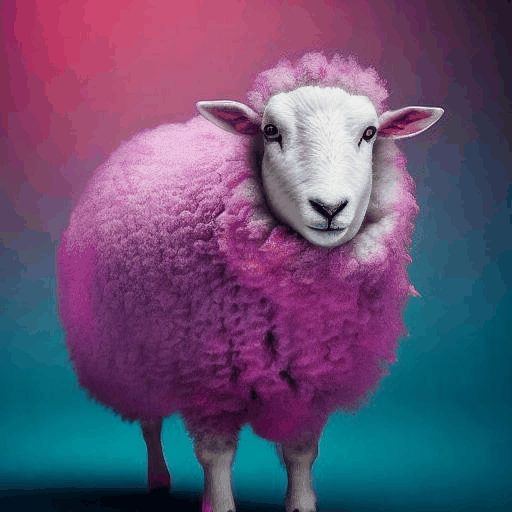}
	\includegraphics[width=0.24\linewidth, trim=0mm 0mm 0mm 0mm, clip]{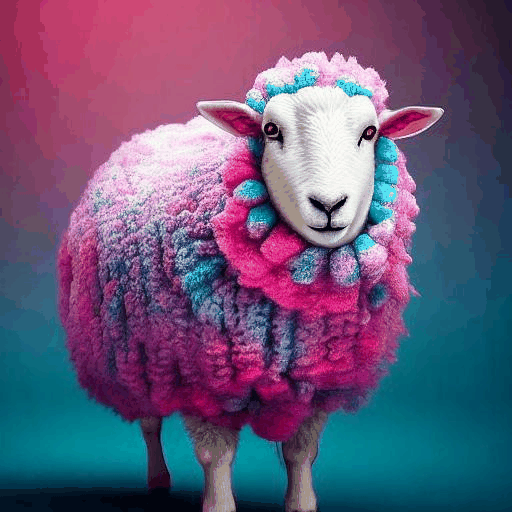}
\caption{A sheep with pink fur is standing.}
\end{subfigure} 
\begin{subfigure}[t]{.49\linewidth}
\centering
	\includegraphics[width=0.24\linewidth, trim=0mm 0mm 0mm 0mm, clip]{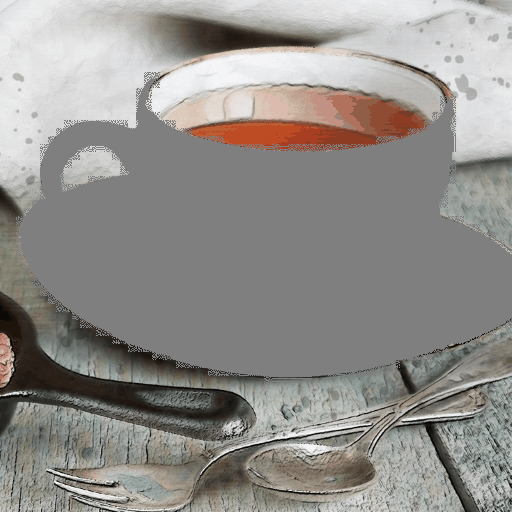}
	\includegraphics[width=0.24\linewidth, trim=0mm 0mm 0mm 0mm, clip]{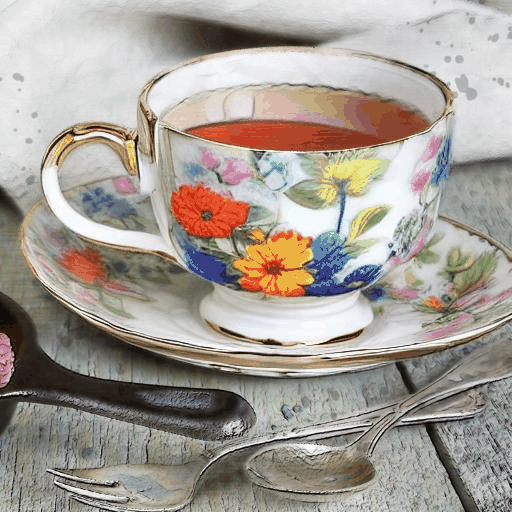}
	\includegraphics[width=0.24\linewidth, trim=0mm 0mm 0mm 0mm, clip]{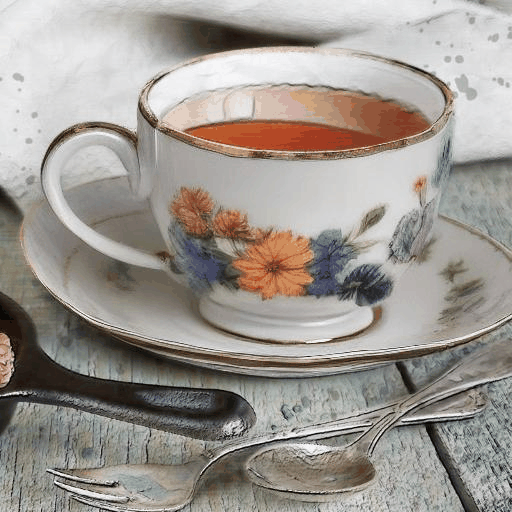}
	\includegraphics[width=0.24\linewidth, trim=0mm 0mm 0mm 0mm, clip]{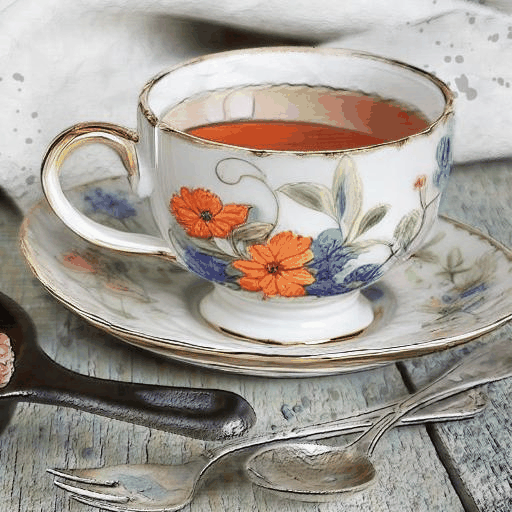}
\caption{A teacup and saucer with spoons.}
\end{subfigure} 
\begin{subfigure}[t]{.49\linewidth}
\centering
	\includegraphics[width=0.24\linewidth, trim=0mm 0mm 0mm 0mm, clip]{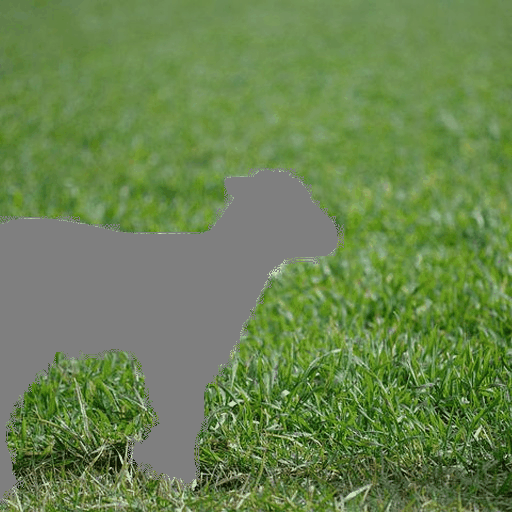}
	\includegraphics[width=0.24\linewidth, trim=0mm 0mm 0mm 0mm, clip]{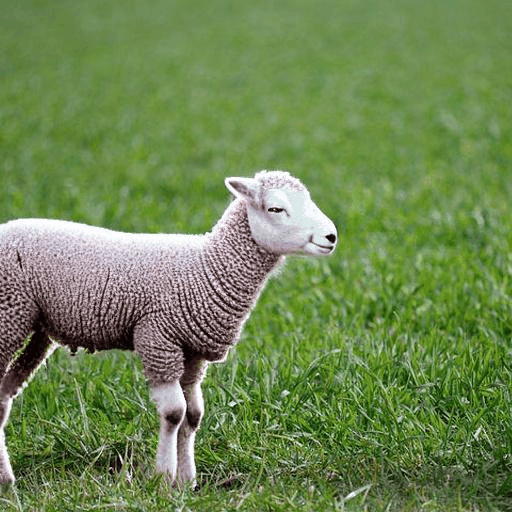}
	\includegraphics[width=0.24\linewidth, trim=0mm 0mm 0mm 0mm, clip]{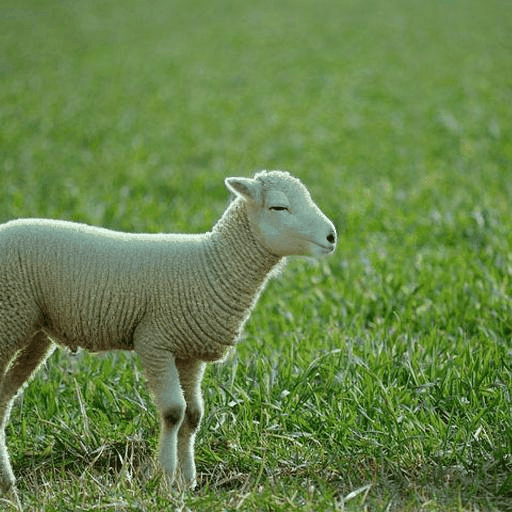}
	\includegraphics[width=0.24\linewidth, trim=0mm 0mm 0mm 0mm, clip]{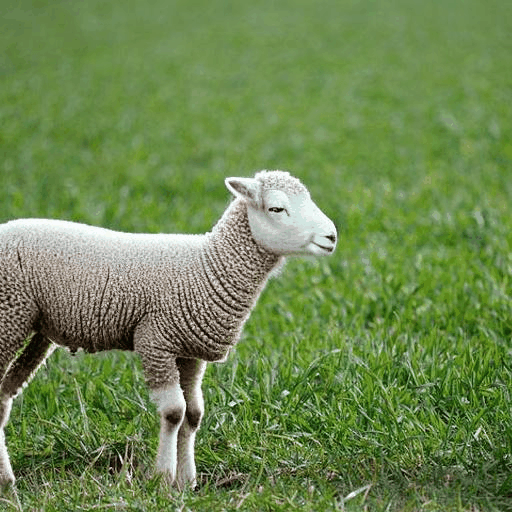}
\caption{A lamb standing in a field of green grass.}
\end{subfigure}
\begin{subfigure}[t]{.49\linewidth}
\centering
	\includegraphics[width=0.24\linewidth, trim=0mm 0mm 0mm 0mm, clip]{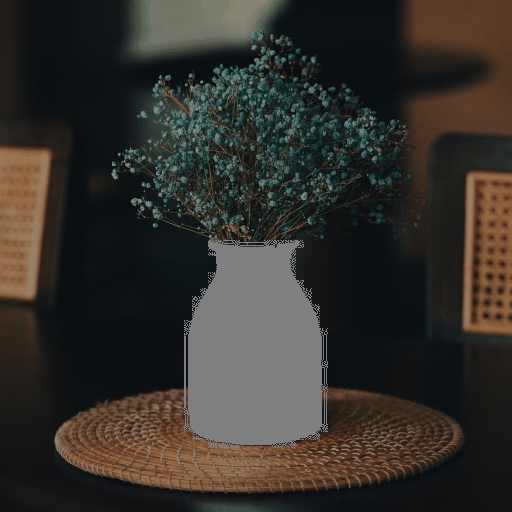}
	\includegraphics[width=0.24\linewidth, trim=0mm 0mm 0mm 0mm, clip]{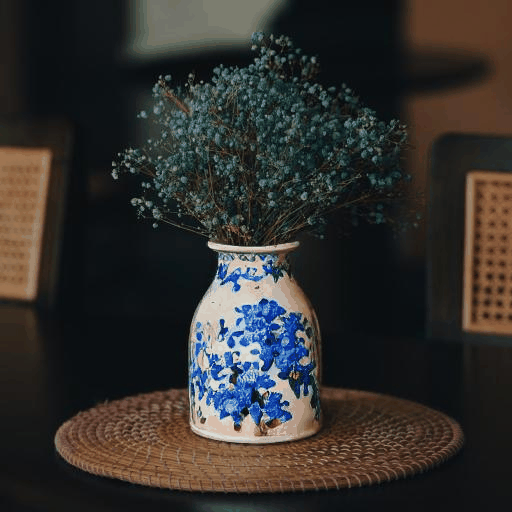}
	\includegraphics[width=0.24\linewidth, trim=0mm 0mm 0mm 0mm, clip]{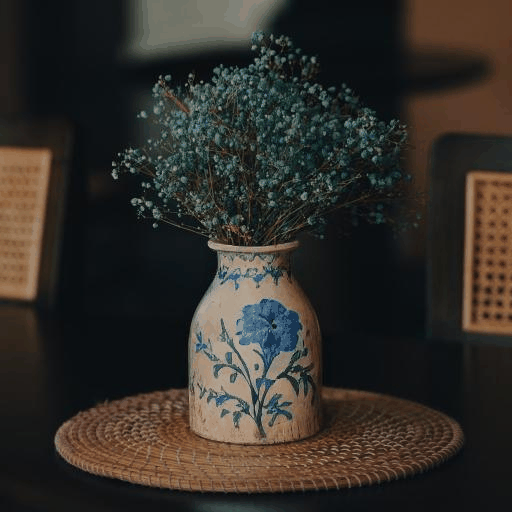}
	\includegraphics[width=0.24\linewidth, trim=0mm 0mm 0mm 0mm, clip]{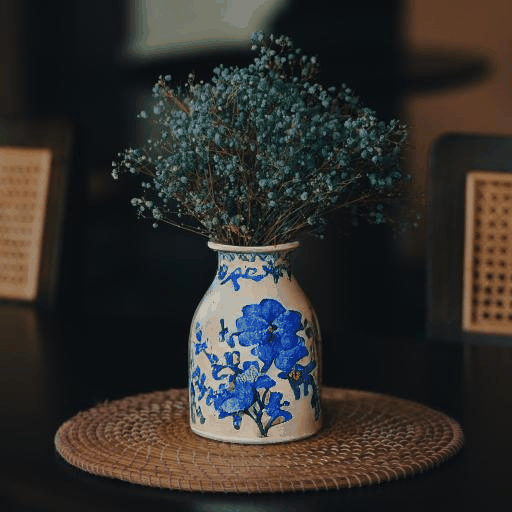}
\caption{A vase with blue flowers sitting on a table.}
\end{subfigure} 
\begin{subfigure}[t]{.49\linewidth}
\centering
	\includegraphics[width=0.24\linewidth, trim=0mm 0mm 0mm 0mm, clip]{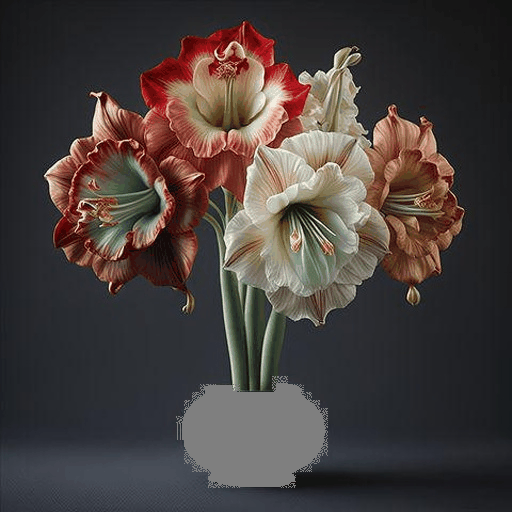}
	\includegraphics[width=0.24\linewidth, trim=0mm 0mm 0mm 0mm, clip]{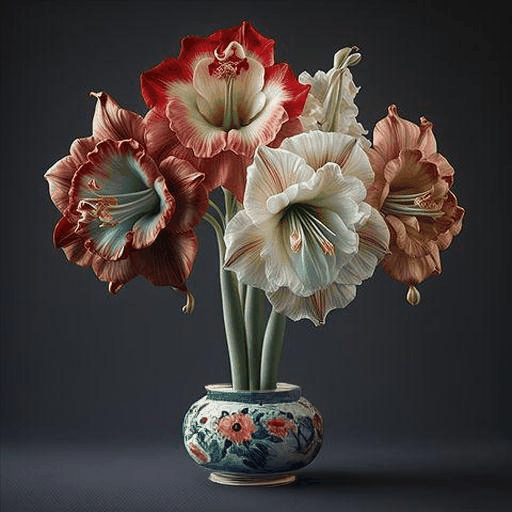}
	\includegraphics[width=0.24\linewidth, trim=0mm 0mm 0mm 0mm, clip]{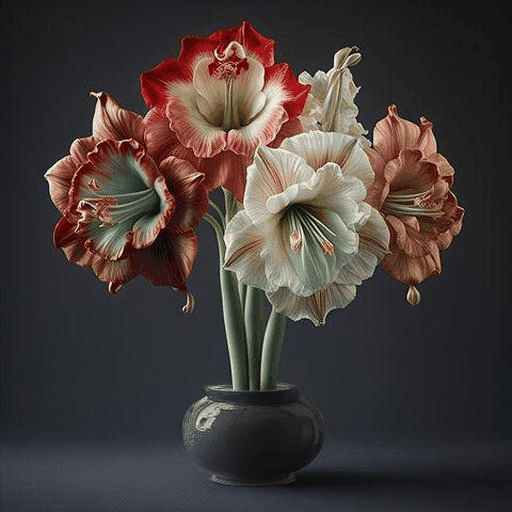}
	\includegraphics[width=0.24\linewidth, trim=0mm 0mm 0mm 0mm, clip]{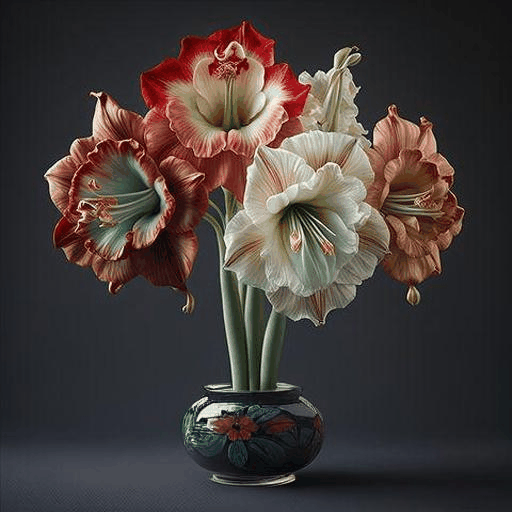}
\caption{A vase with some flowers in it.}
\end{subfigure} 
\hfill
\begin{subfigure}[t]{.49\linewidth}
\centering
	\includegraphics[width=0.24\linewidth, trim=0mm 0mm 0mm 0mm, clip]{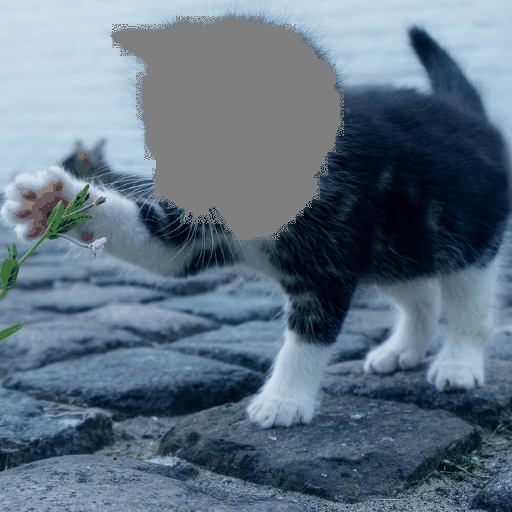}
	\includegraphics[width=0.24\linewidth, trim=0mm 0mm 0mm 0mm, clip]{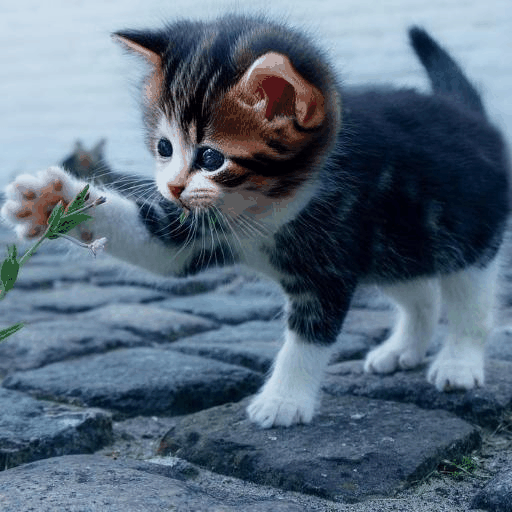}
	\includegraphics[width=0.24\linewidth, trim=0mm 0mm 0mm 0mm, clip]{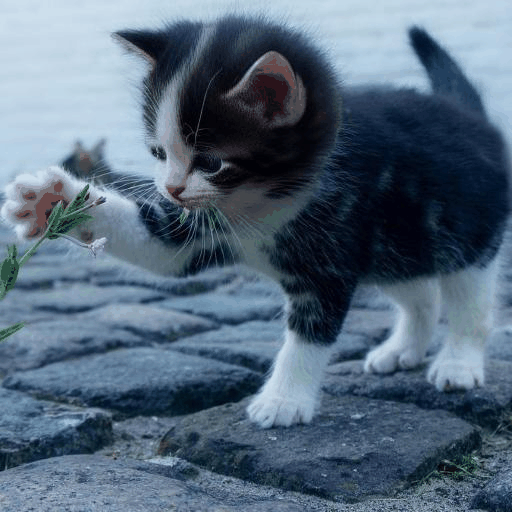}
	\includegraphics[width=0.24\linewidth, trim=0mm 0mm 0mm 0mm, clip]{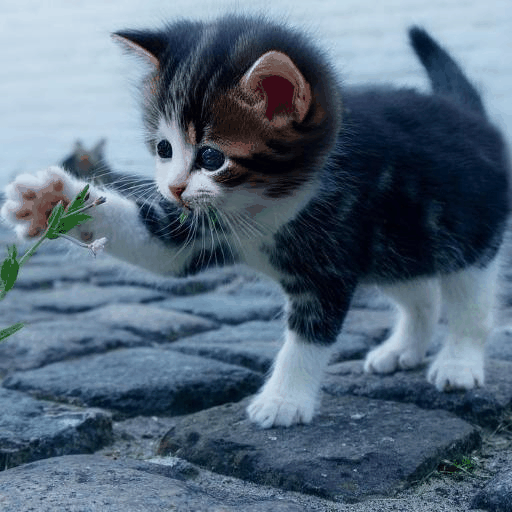}
\caption{A kitten is playing with a flower.}
\end{subfigure}
\caption{\textbf{More results on reward model bias studies using BrushNet.} In each sub-figure, the four images (from left to right) display: the \textit{masked image}, followed by inpainting results from models trained using \textit{HPSv2}, \textit{PickScore}, and \textit{Ensemble}. For optimal detail, view figures zoomed in.}
\label{fig:brushnet_bias_appendix}
\end{figure}

\begin{figure}[!htbp]
\begin{subfigure}[t]{.49\linewidth}
\centering
	\includegraphics[width=0.24\linewidth, trim=0mm 0mm 0mm 0mm, clip]{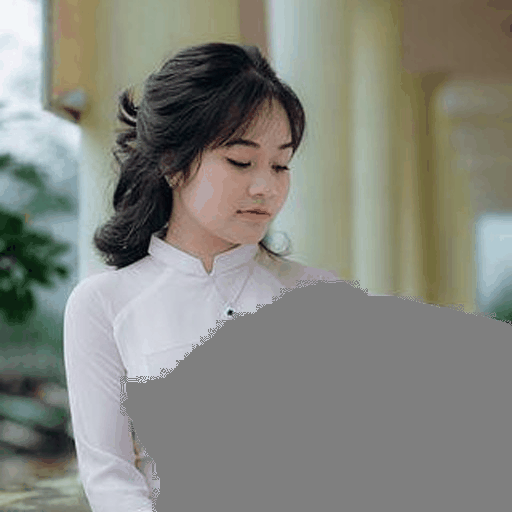}
	\includegraphics[width=0.24\linewidth, trim=0mm 0mm 0mm 0mm, clip]{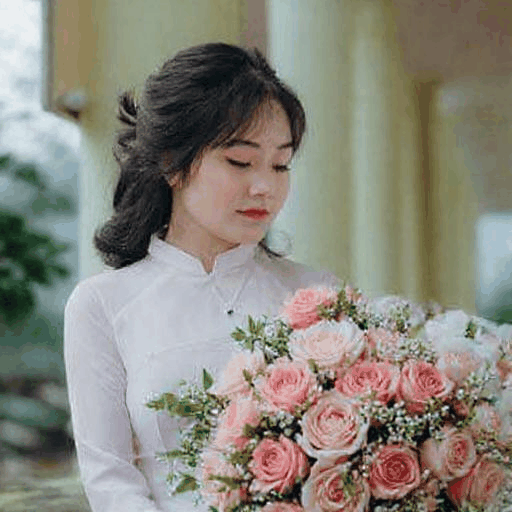}
	\includegraphics[width=0.24\linewidth, trim=0mm 0mm 0mm 0mm, clip]{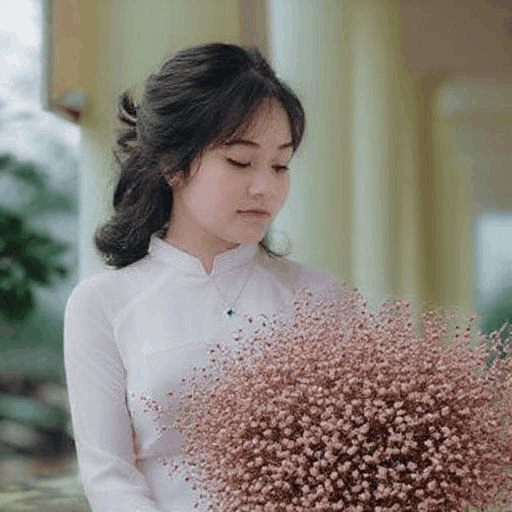}
	\includegraphics[width=0.24\linewidth, trim=0mm 0mm 0mm 0mm, clip]{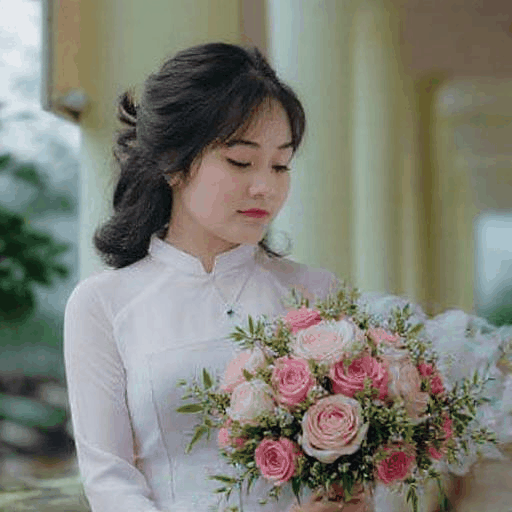}
\caption{A woman in white holding a bouquet of flowers.}
\end{subfigure} 
\begin{subfigure}[t]{.49\linewidth}
\centering
	\includegraphics[width=0.24\linewidth, trim=0mm 0mm 0mm 0mm, clip]{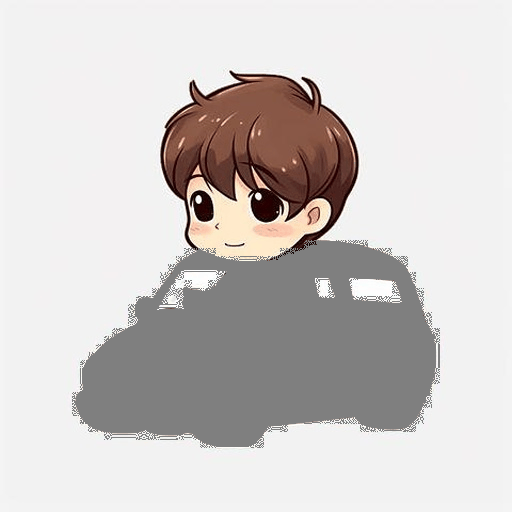}
	\includegraphics[width=0.24\linewidth, trim=0mm 0mm 0mm 0mm, clip]{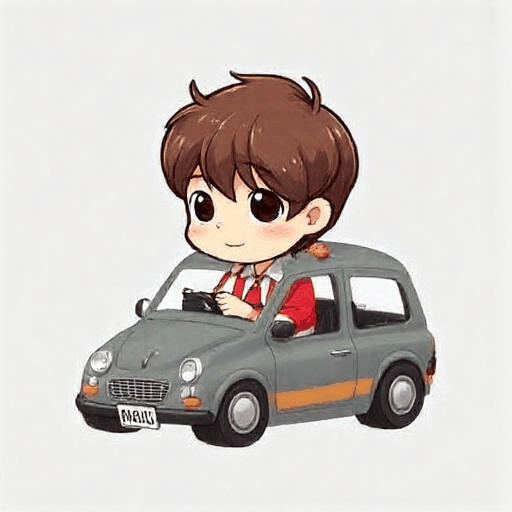}
	\includegraphics[width=0.24\linewidth, trim=0mm 0mm 0mm 0mm, clip]{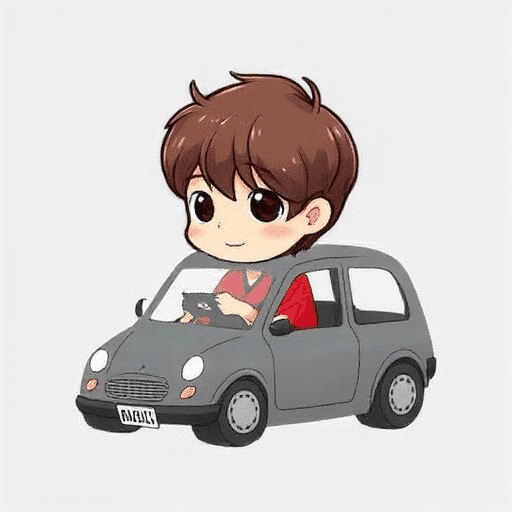}
	\includegraphics[width=0.24\linewidth, trim=0mm 0mm 0mm 0mm, clip]{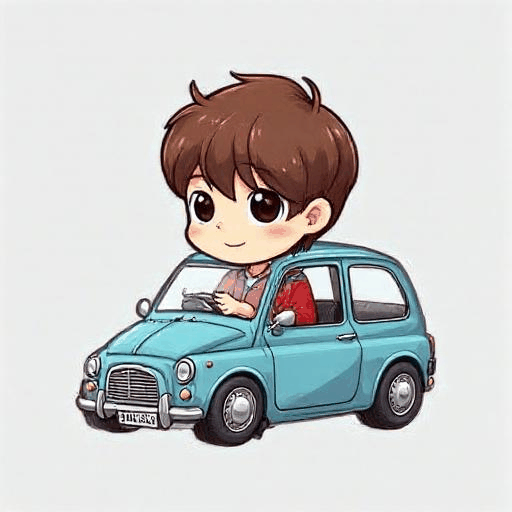}
\caption{A cartoon boy driving a car.}
\end{subfigure} 
\begin{subfigure}[t]{.49\linewidth}
\centering
	\includegraphics[width=0.24\linewidth, trim=0mm 0mm 0mm 0mm, clip]{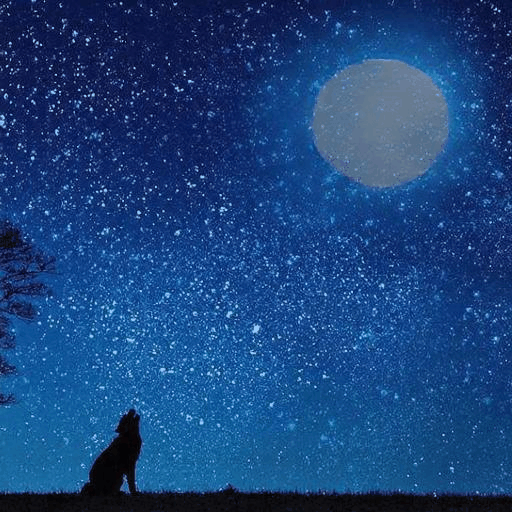}
	\includegraphics[width=0.24\linewidth, trim=0mm 0mm 0mm 0mm, clip]{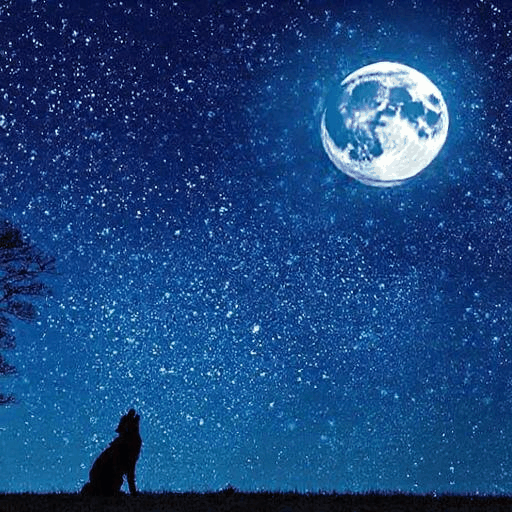}
	\includegraphics[width=0.24\linewidth, trim=0mm 0mm 0mm 0mm, clip]{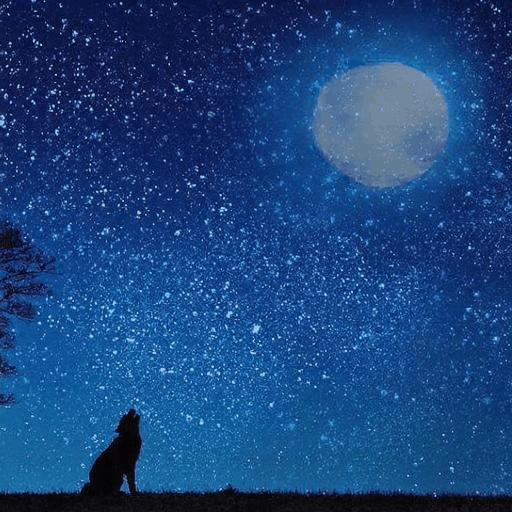}
	\includegraphics[width=0.24\linewidth, trim=0mm 0mm 0mm 0mm, clip]{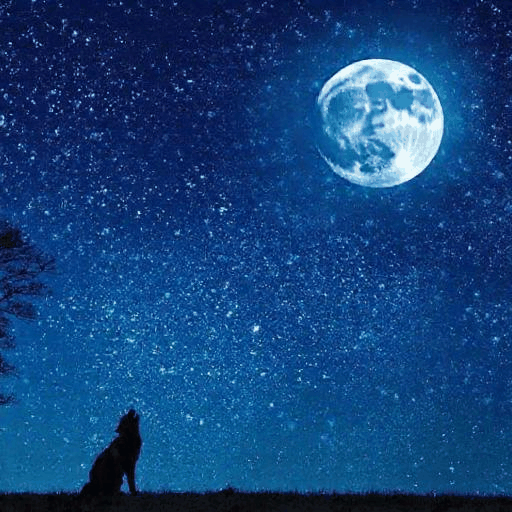}
\caption{Wolf howling at the moon.}
\end{subfigure}
\begin{subfigure}[t]{.49\linewidth}
\centering
	\includegraphics[width=0.24\linewidth, trim=0mm 0mm 0mm 0mm, clip]{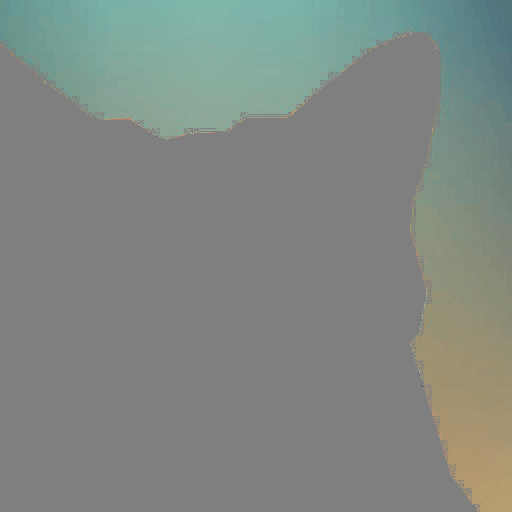}
	\includegraphics[width=0.24\linewidth, trim=0mm 0mm 0mm 0mm, clip]{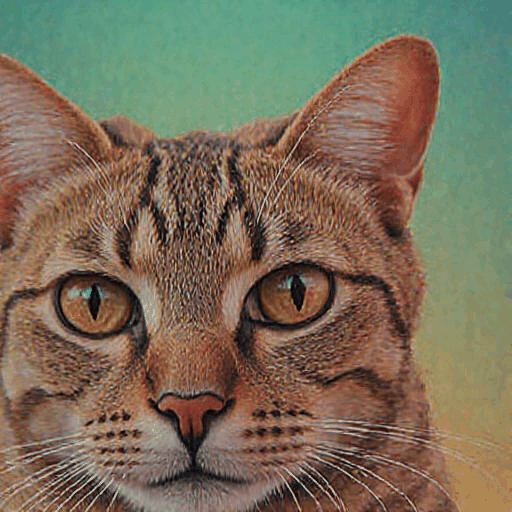}
	\includegraphics[width=0.24\linewidth, trim=0mm 0mm 0mm 0mm, clip]{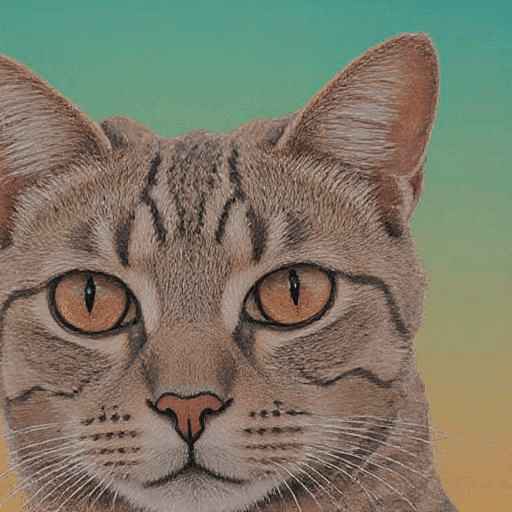}
	\includegraphics[width=0.24\linewidth, trim=0mm 0mm 0mm 0mm, clip]{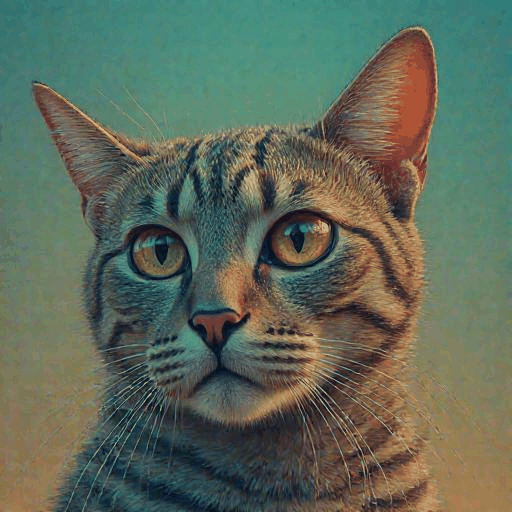}
\caption{A cat is shown in low polygonal style.}
\end{subfigure} 
\begin{subfigure}[t]{.49\linewidth}
\centering
	\includegraphics[width=0.24\linewidth, trim=0mm 0mm 0mm 0mm, clip]{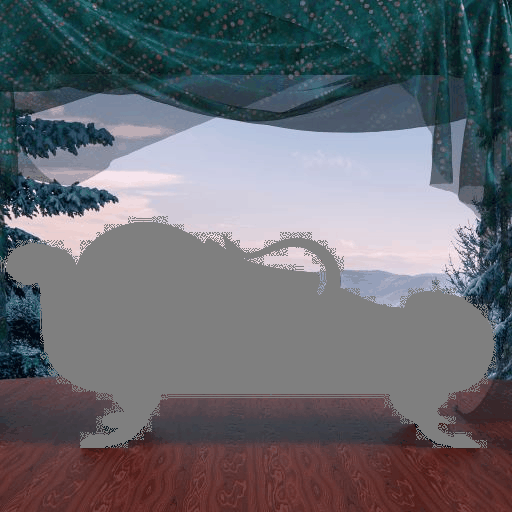}
	\includegraphics[width=0.24\linewidth, trim=0mm 0mm 0mm 0mm, clip]{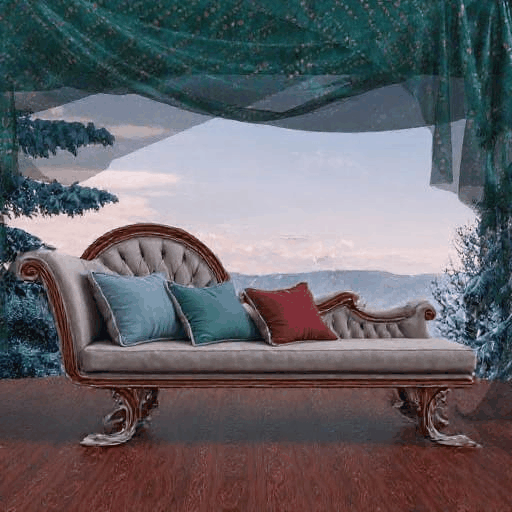}
	\includegraphics[width=0.24\linewidth, trim=0mm 0mm 0mm 0mm, clip]{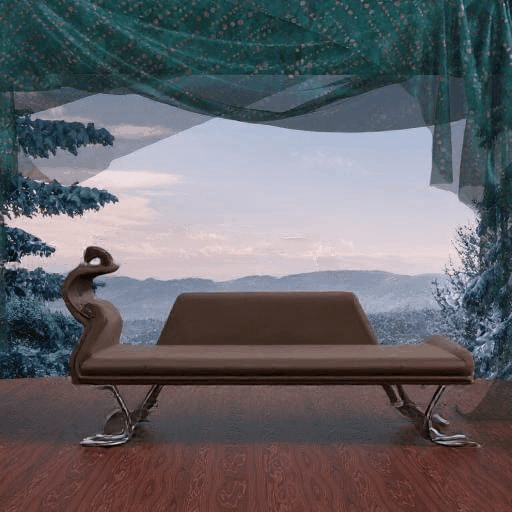}
	\includegraphics[width=0.24\linewidth, trim=0mm 0mm 0mm 0mm, clip]{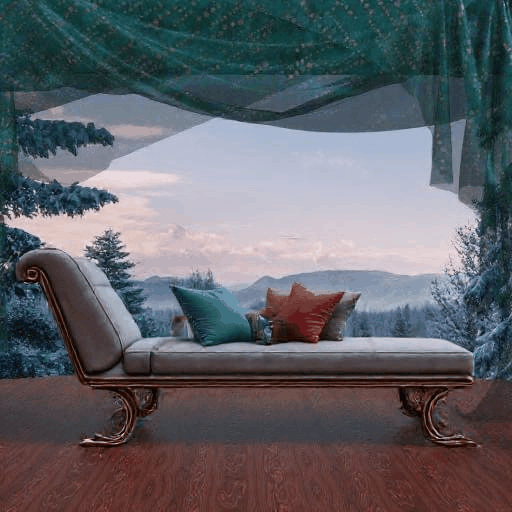}
\caption{A couch with a winged chair at a window.}
\end{subfigure} 
\begin{subfigure}[t]{.49\linewidth}
\centering
	\includegraphics[width=0.24\linewidth, trim=0mm 0mm 0mm 0mm, clip]{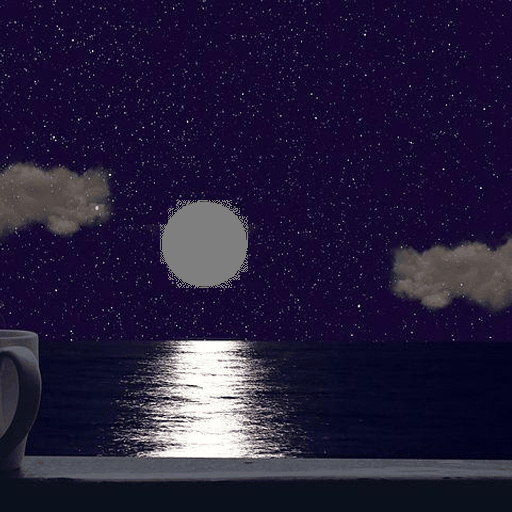}
	\includegraphics[width=0.24\linewidth, trim=0mm 0mm 0mm 0mm, clip]{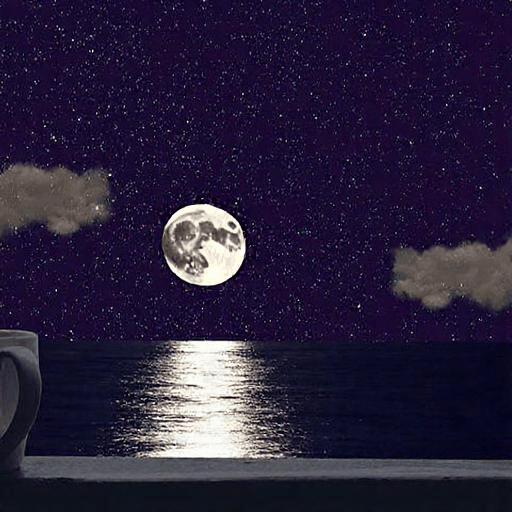}
	\includegraphics[width=0.24\linewidth, trim=0mm 0mm 0mm 0mm, clip]{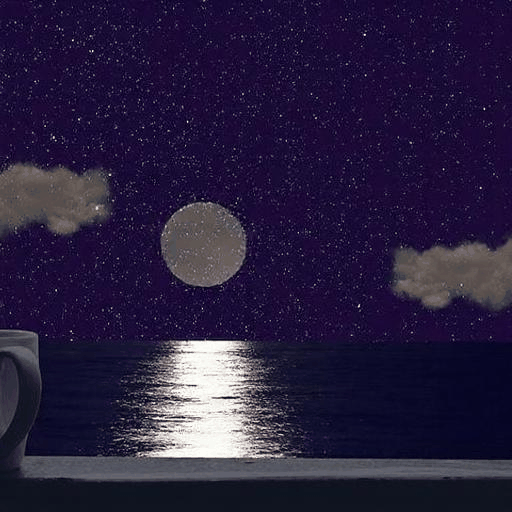}
	\includegraphics[width=0.24\linewidth, trim=0mm 0mm 0mm 0mm, clip]{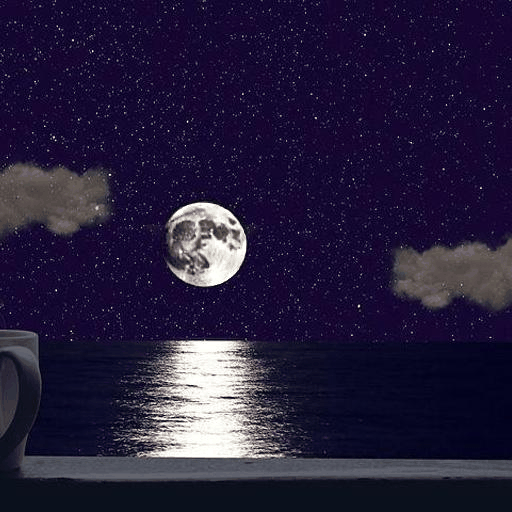}
\caption{A bright moon on the sea.}
\end{subfigure} 
\begin{subfigure}[t]{.49\linewidth}
\centering
	\includegraphics[width=0.24\linewidth, trim=0mm 0mm 0mm 0mm, clip]{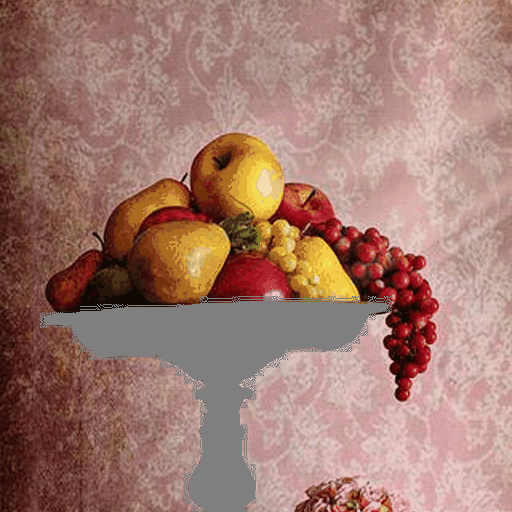}
	\includegraphics[width=0.24\linewidth, trim=0mm 0mm 0mm 0mm, clip]{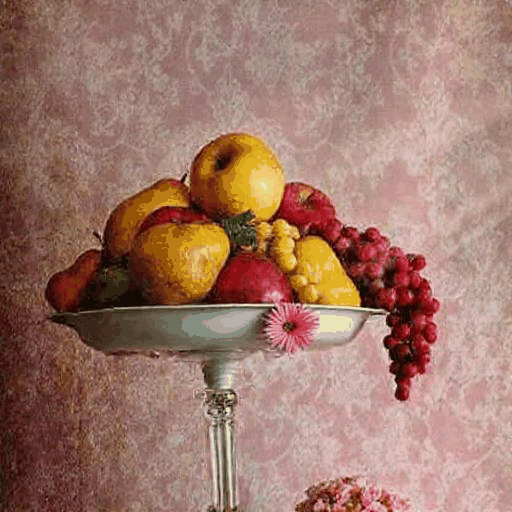}
	\includegraphics[width=0.24\linewidth, trim=0mm 0mm 0mm 0mm, clip]{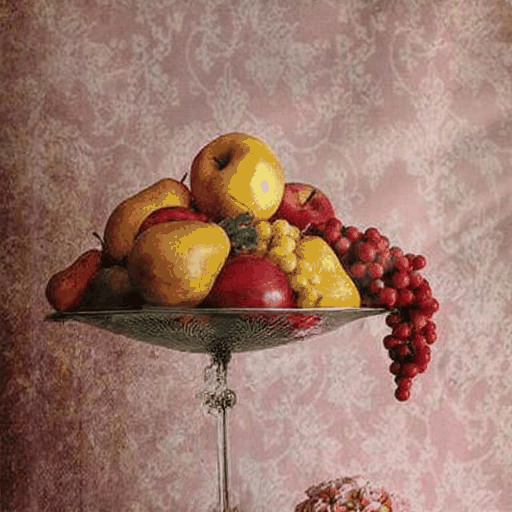}
	\includegraphics[width=0.24\linewidth, trim=0mm 0mm 0mm 0mm, clip]{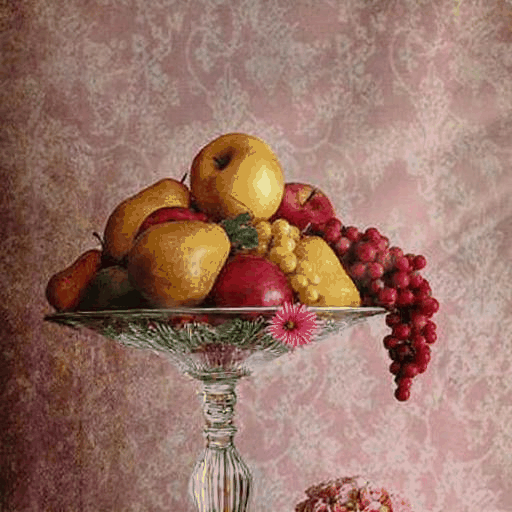}
\caption{A fruit bowl with a pink flower on top.}
\end{subfigure} 
\begin{subfigure}[t]{.49\linewidth}
\centering
	\includegraphics[width=0.24\linewidth, trim=0mm 0mm 0mm 0mm, clip]{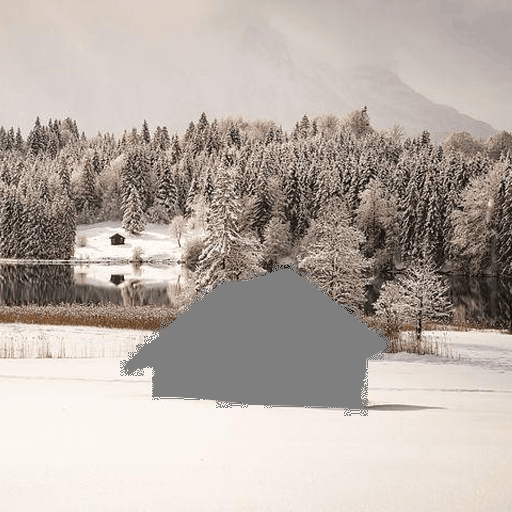}
	\includegraphics[width=0.24\linewidth, trim=0mm 0mm 0mm 0mm, clip]{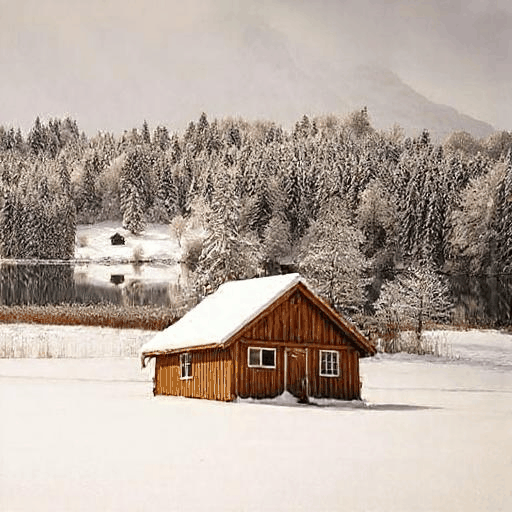}
	\includegraphics[width=0.24\linewidth, trim=0mm 0mm 0mm 0mm, clip]{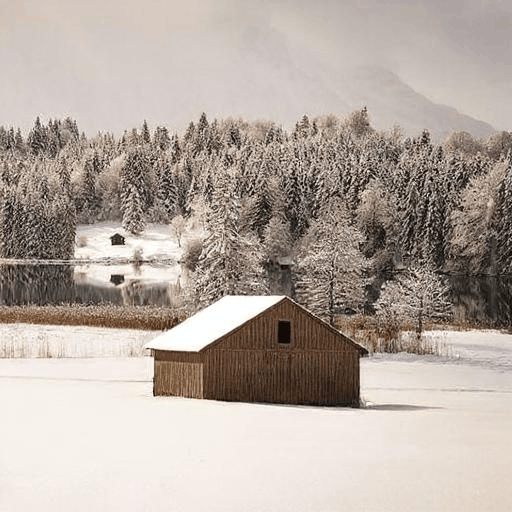}
	\includegraphics[width=0.24\linewidth, trim=0mm 0mm 0mm 0mm, clip]{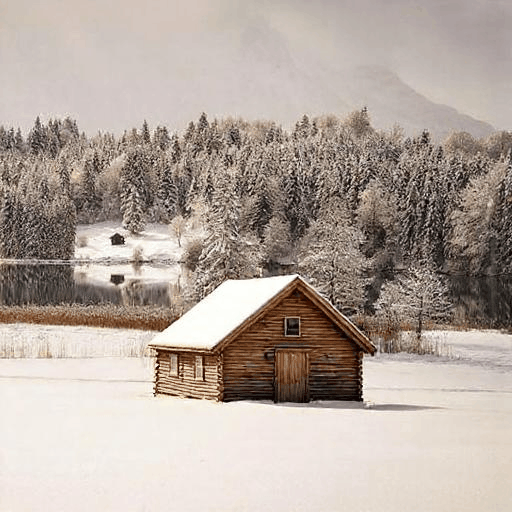}
\caption{A small cabin in the snow near a lake.}
\end{subfigure} 
\begin{subfigure}[t]{.49\linewidth}
\centering
	\includegraphics[width=0.24\linewidth, trim=0mm 0mm 0mm 0mm, clip]{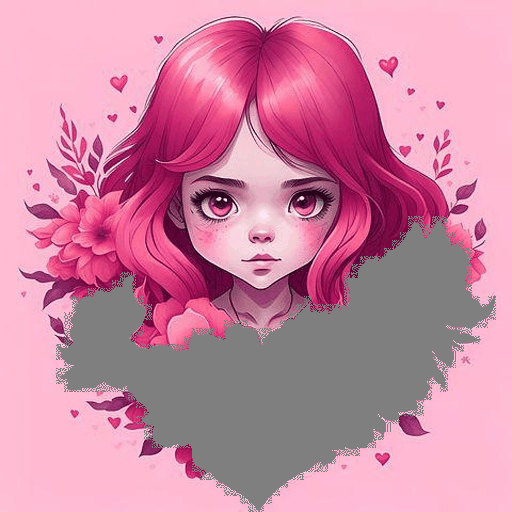}
	\includegraphics[width=0.24\linewidth, trim=0mm 0mm 0mm 0mm, clip]{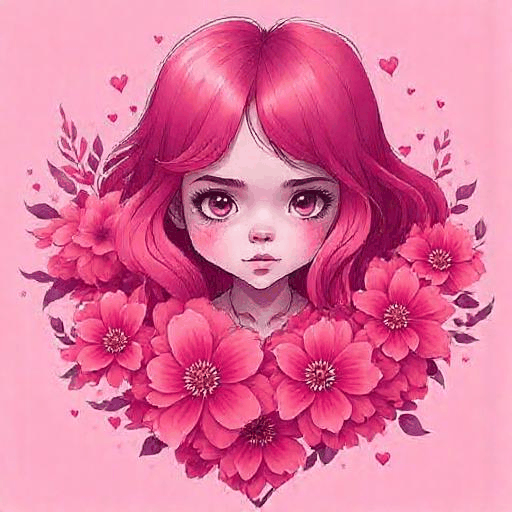}
	\includegraphics[width=0.24\linewidth, trim=0mm 0mm 0mm 0mm, clip]{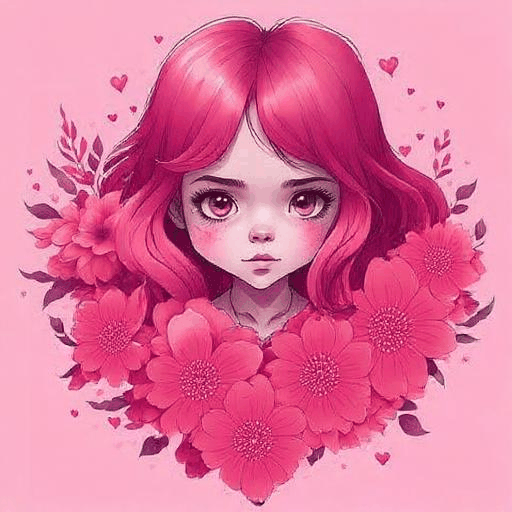}
	\includegraphics[width=0.24\linewidth, trim=0mm 0mm 0mm 0mm, clip]{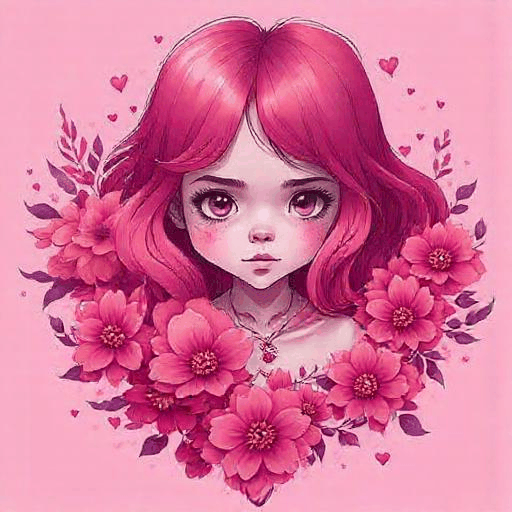}
\caption{A girl with pink hair and flowers on her face.}
\end{subfigure}
\begin{subfigure}[t]{.49\linewidth}
\centering
	\includegraphics[width=0.24\linewidth, trim=0mm 0mm 0mm 0mm, clip]{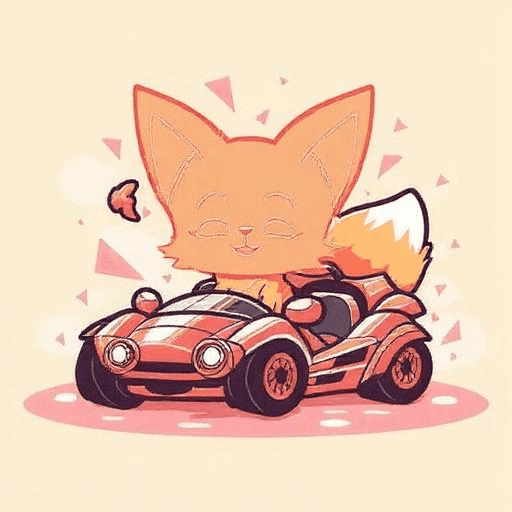}
	\includegraphics[width=0.24\linewidth, trim=0mm 0mm 0mm 0mm, clip]{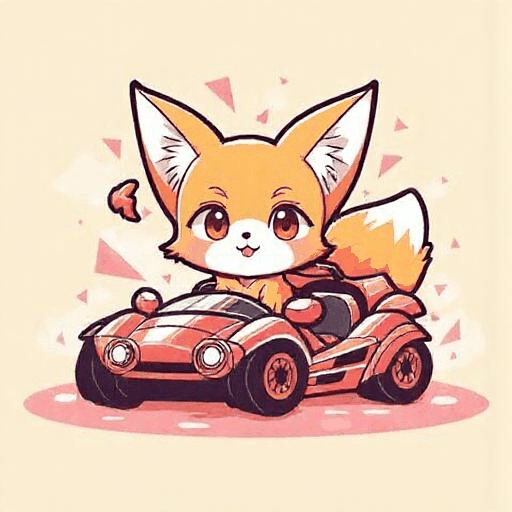}
	\includegraphics[width=0.24\linewidth, trim=0mm 0mm 0mm 0mm, clip]{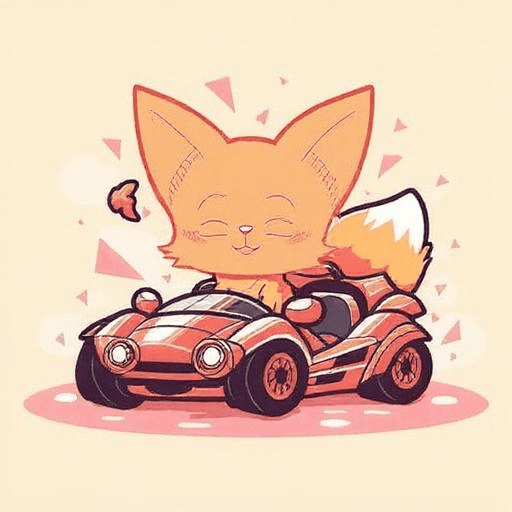}
	\includegraphics[width=0.24\linewidth, trim=0mm 0mm 0mm 0mm, clip]{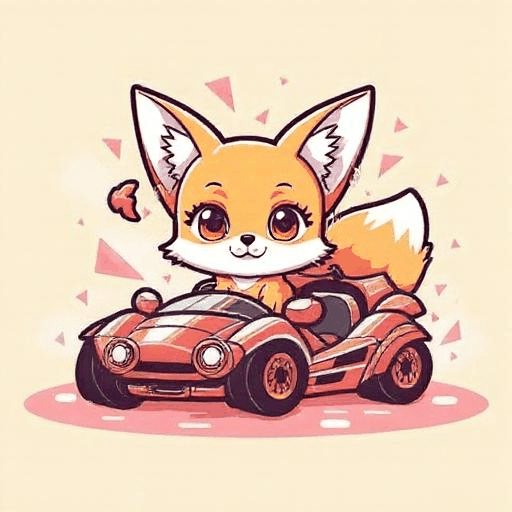}
\caption{Cartoon fox driving a car with a cute face.}
\end{subfigure} 
\begin{subfigure}[t]{.49\linewidth}
\centering
	\includegraphics[width=0.24\linewidth, trim=0mm 0mm 0mm 0mm, clip]{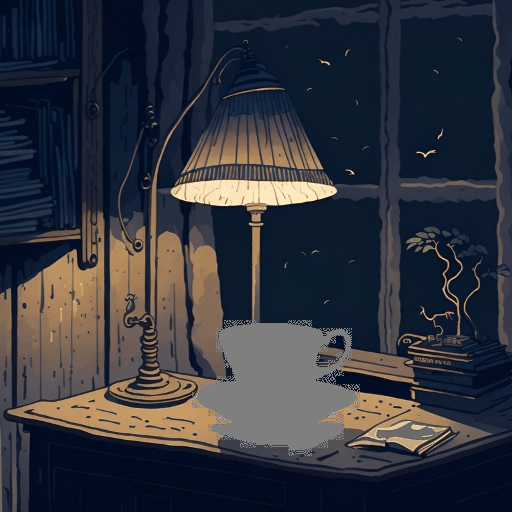}
	\includegraphics[width=0.24\linewidth, trim=0mm 0mm 0mm 0mm, clip]{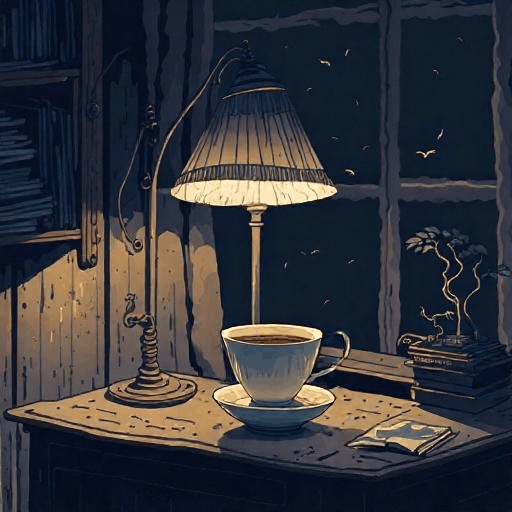}
	\includegraphics[width=0.24\linewidth, trim=0mm 0mm 0mm 0mm, clip]{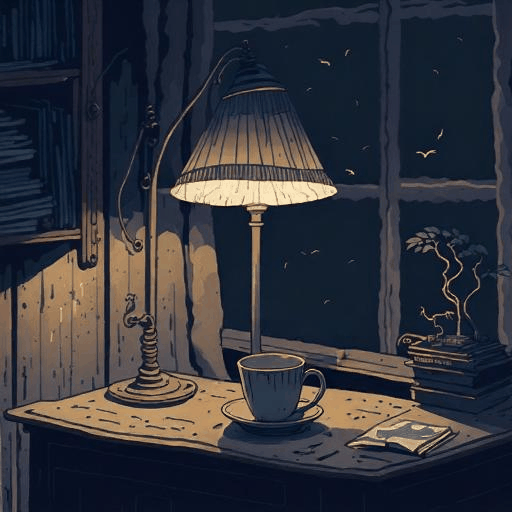}
	\includegraphics[width=0.24\linewidth, trim=0mm 0mm 0mm 0mm, clip]{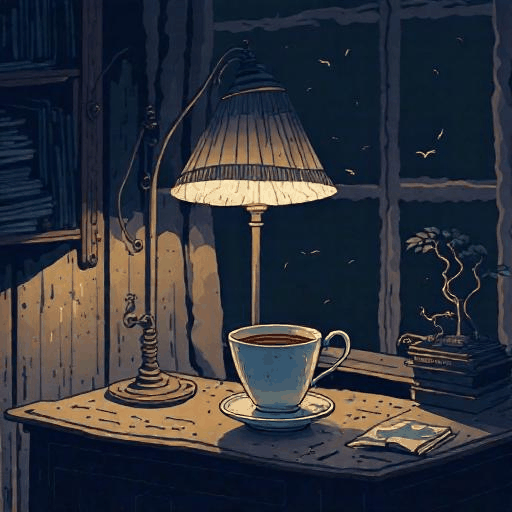}
\caption{A lamp and a cup of tea on a table.}
\end{subfigure} 
\begin{subfigure}[t]{.49\linewidth}
\centering
	\includegraphics[width=0.24\linewidth, trim=0mm 0mm 0mm 0mm, clip]{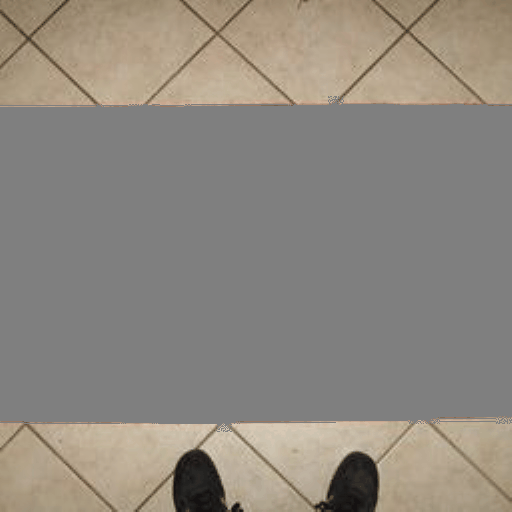}
	\includegraphics[width=0.24\linewidth, trim=0mm 0mm 0mm 0mm, clip]{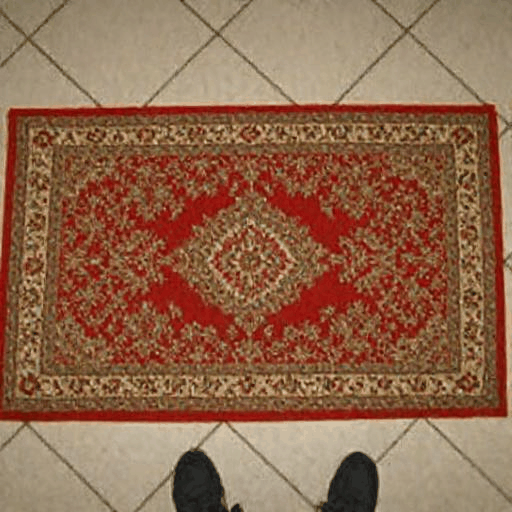}
	\includegraphics[width=0.24\linewidth, trim=0mm 0mm 0mm 0mm, clip]{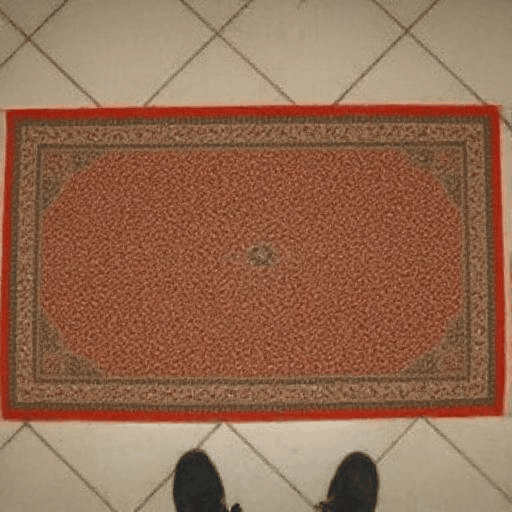}
	\includegraphics[width=0.24\linewidth, trim=0mm 0mm 0mm 0mm, clip]{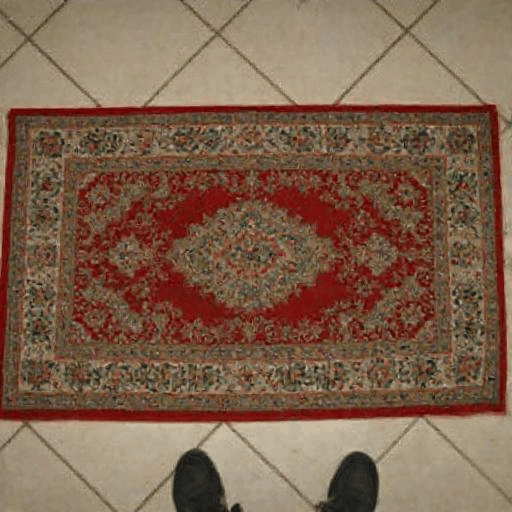}
\caption{A person standing on a tile floor with a rug.}
\end{subfigure} 
\begin{subfigure}[t]{.49\linewidth}
\centering
	\includegraphics[width=0.24\linewidth, trim=0mm 0mm 0mm 0mm, clip]{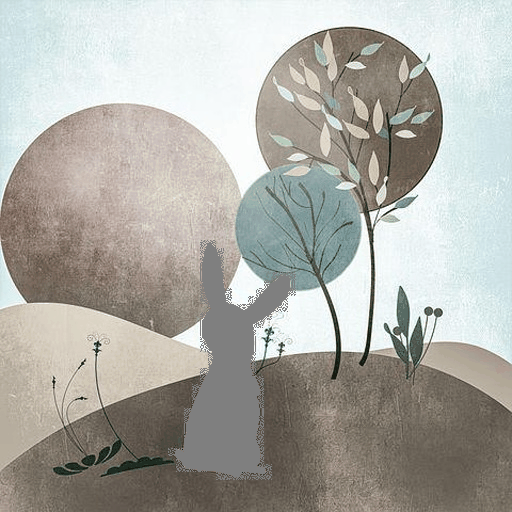}
	\includegraphics[width=0.24\linewidth, trim=0mm 0mm 0mm 0mm, clip]{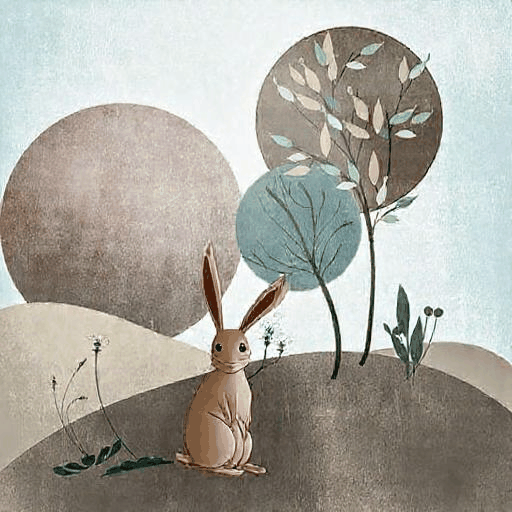}
	\includegraphics[width=0.24\linewidth, trim=0mm 0mm 0mm 0mm, clip]{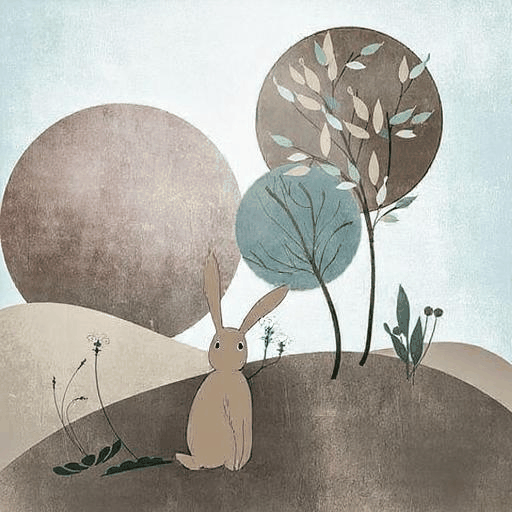}
	\includegraphics[width=0.24\linewidth, trim=0mm 0mm 0mm 0mm, clip]{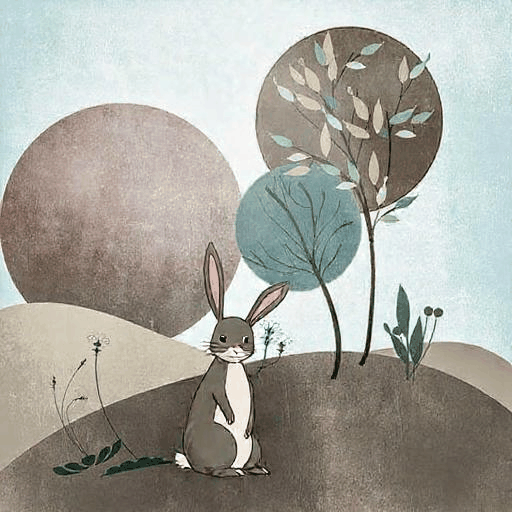}
\caption{A rabbit sitting on a hill with trees.}
\end{subfigure} 
\begin{subfigure}[t]{.49\linewidth}
\centering
	\includegraphics[width=0.24\linewidth, trim=0mm 0mm 0mm 0mm, clip]{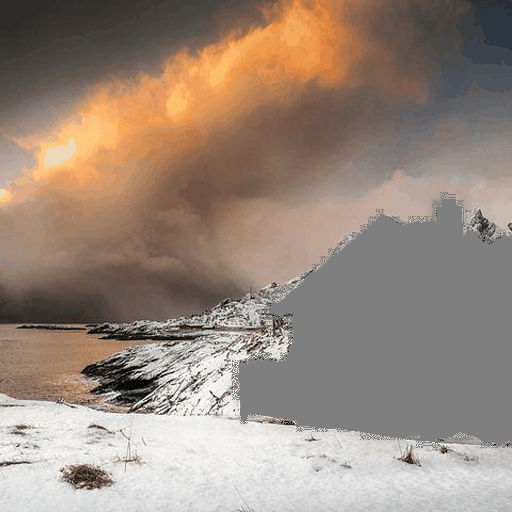}
	\includegraphics[width=0.24\linewidth, trim=0mm 0mm 0mm 0mm, clip]{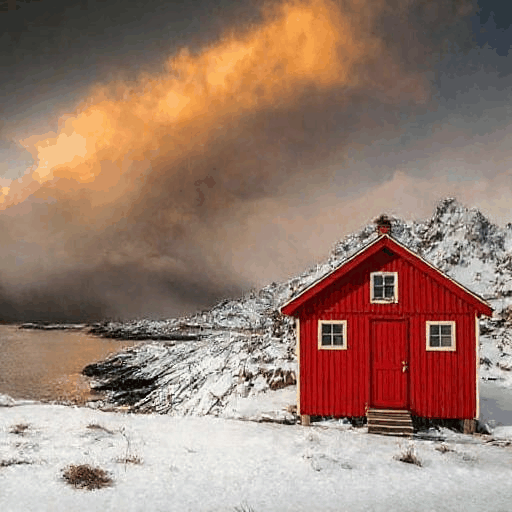}
	\includegraphics[width=0.24\linewidth, trim=0mm 0mm 0mm 0mm, clip]{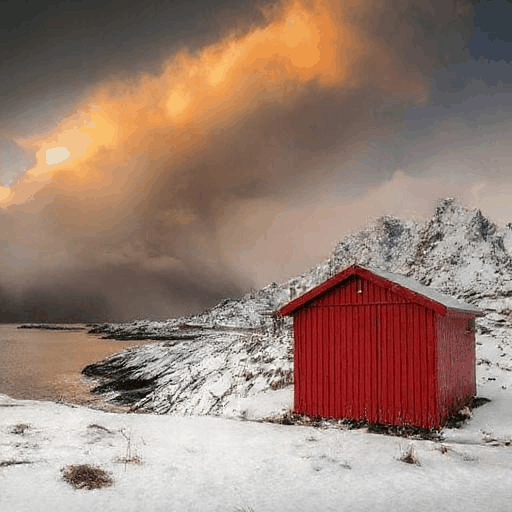}
	\includegraphics[width=0.24\linewidth, trim=0mm 0mm 0mm 0mm, clip]{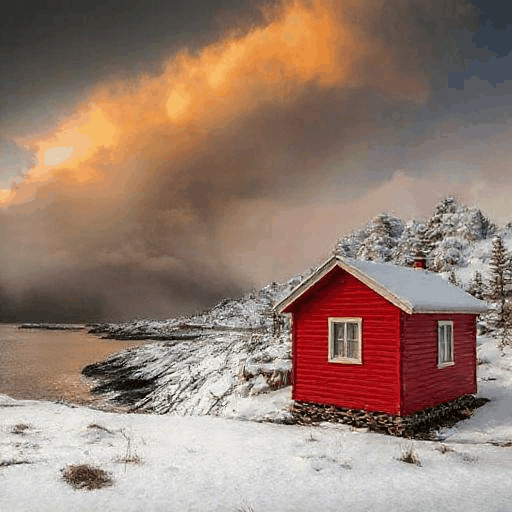}
\caption{A red cabin sits on the shore of a lake.}
\end{subfigure} 
\begin{subfigure}[t]{.49\linewidth}
\centering
	\includegraphics[width=0.24\linewidth, trim=0mm 0mm 0mm 0mm, clip]{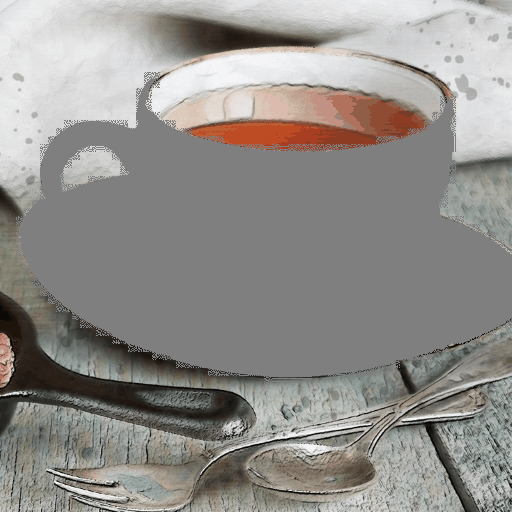}
	\includegraphics[width=0.24\linewidth, trim=0mm 0mm 0mm 0mm, clip]{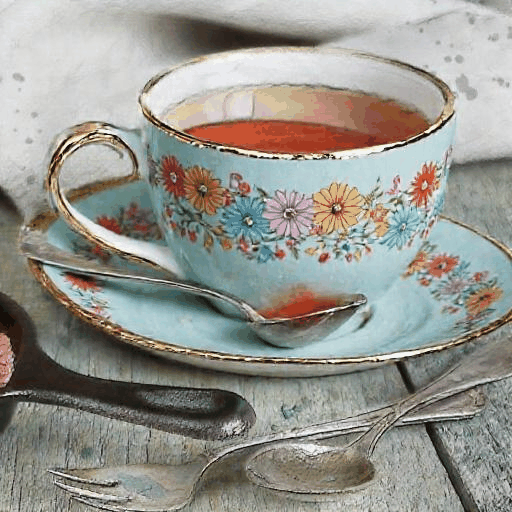}
	\includegraphics[width=0.24\linewidth, trim=0mm 0mm 0mm 0mm, clip]{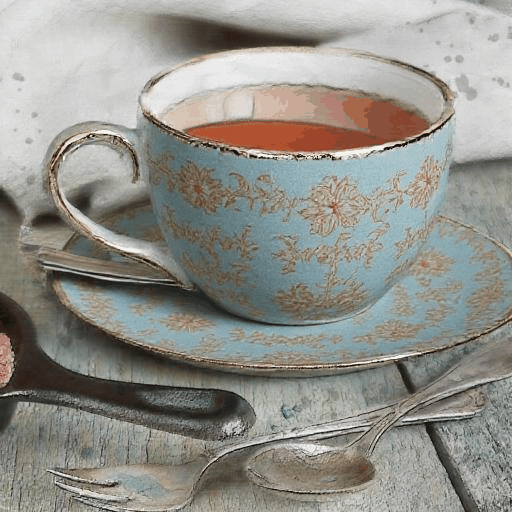}
	\includegraphics[width=0.24\linewidth, trim=0mm 0mm 0mm 0mm, clip]{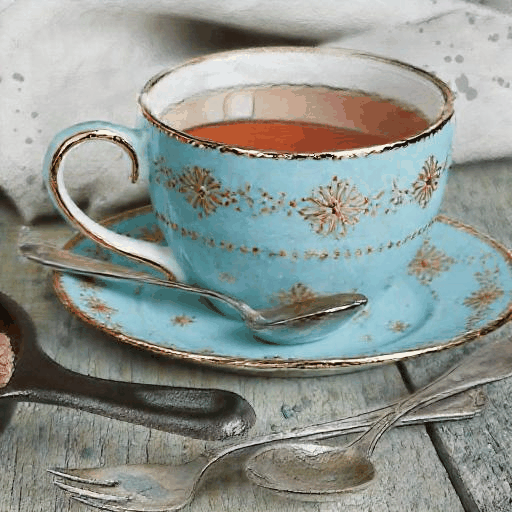}
\caption{A teacup and saucer with spoons.}
\end{subfigure} 
\hfill
\begin{subfigure}[t]{.49\linewidth}
\centering
	\includegraphics[width=0.24\linewidth, trim=0mm 0mm 0mm 0mm, clip]{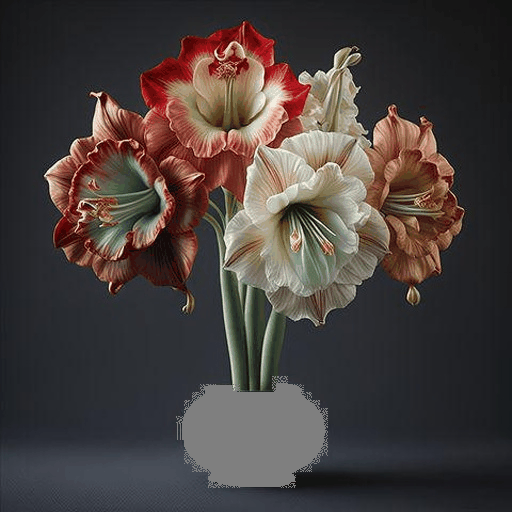}
	\includegraphics[width=0.24\linewidth, trim=0mm 0mm 0mm 0mm, clip]{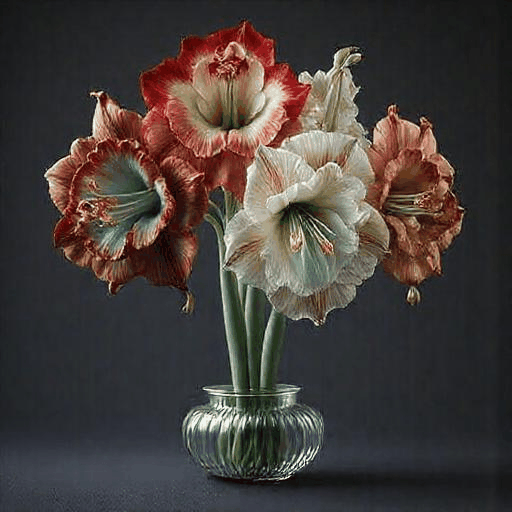}
	\includegraphics[width=0.24\linewidth, trim=0mm 0mm 0mm 0mm, clip]{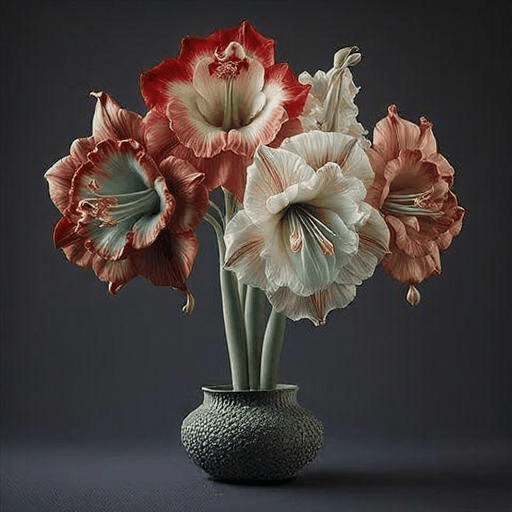}
	\includegraphics[width=0.24\linewidth, trim=0mm 0mm 0mm 0mm, clip]{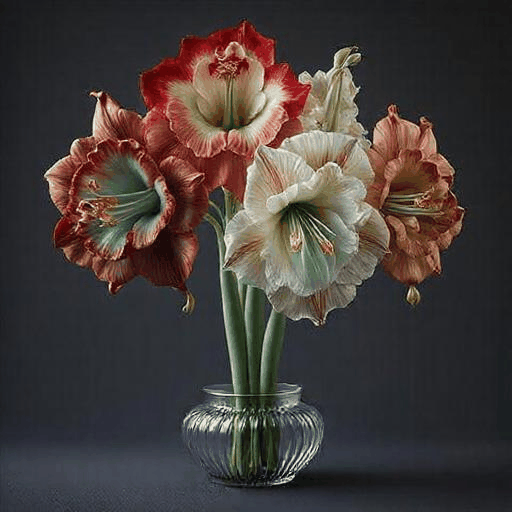}
\caption{A vase with flowers in it on a dark background.}
\end{subfigure}
\caption{\textbf{More results on reward model bias studies using FLUX.1 Fill.} In each sub-figure, the four images (from left to right) display: the \textit{masked image}, followed by inpainting results from models trained using \textit{HPSv2}, \textit{PickScore}, and \textit{Ensemble}. For optimal detail, view figures zoomed in.}
\label{fig:flux_bias_appendix}
\end{figure}
% }

\end{document}